\RequirePackage{fix-cm}
\documentclass[twocolumn]{svjour3}
\smartqed
\usepackage{graphicx}
\usepackage{graphicx}
\usepackage{amsmath,amsfonts,amssymb}
\usepackage{color}
\usepackage{xspace}
\usepackage{xcolor}
\usepackage{array}
\usepackage{multirow}
\usepackage{tabularx}
\usepackage{subcaption}
\usepackage{natbib}
\setcitestyle{authoryear,open={(},close={)}}
\usepackage{url}
\usepackage{tabularx}
\usepackage{mdframed}
\usepackage{hhline}
\RequirePackage{snapshot}
\usepackage{nicefrac}
\usepackage{afterpage}
\usepackage{tikz}
\usetikzlibrary{calc}

\newcommand{\figref}[1]{\Fig~\ref{#1}}
\newcommand{\secref}[1]{Section~\ref{#1}}
\newcommand{\eqnref}[1]{Eq.~\eqref{#1}}
\newcommand{\tabref}[1]{Table~\ref{#1}}

\DeclareMathOperator*{\ntimes}{\!\times\!}

\makeatletter
\DeclareRobustCommand\onedot{\futurelet\@let@token\@onedot}
\def\@onedot{\ifx\@let@token.\else.\null\fi\xspace}
\def\eg{e.g\onedot} 
\def\ie{i.e\onedot} 
\def\cf{cf\onedot} 
 
\def\wrt{wrt\onedot}

\def\etal{et~al\onedot} 
\def\Fig{Fig\onedot}   
\makeatother

\newcommand{\boldparagraph}[1]{\vspace{0.2cm}\noindent{\bf #1:} }

\definecolor{darkgreen}{rgb}{0,0.7,0}
\newcommand{\red}[1]{#1}

\newcommand{\uk}{\ensuremath{\bot}}
\newcommand{\AML}{AML\xspace}
\newcommand{\dAML}{dAML\xspace}
\newcommand{\ML}{ML\xspace}
\newcommand{\Sup}{Sup\xspace}

\newcommand{\VAE}{VAE\xspace}
\newcommand{\DVAE}{DVAE\xspace}
\newcommand{\VAEs}{VAEs\xspace}
\newcommand{\DVAEs}{DVAEs\xspace}
\newcommand{\BL}{Na\"ive\xspace}
\newcommand{\Dai}{Dai17\xspace}
\newcommand{\Engelmann}{Eng16\xspace}
\newcommand{\ICP}{ICP\xspace}
\newcommand{\M}{Mean\xspace}
\newcommand{\clean}{SN-clean\xspace}
\newcommand{\noisy}{SN-noisy\xspace}
\newcommand{\Abs}{Ham\xspace}
\newcommand{\Acc}{Acc\xspace}
\newcommand{\Compl}{Comp\xspace}
\newcommand{\IoU}{IoU\xspace}

\newcommand{\GANs}{GANs\xspace}
\newcommand{\Kinect}{Kinect\xspace}

\usepackage{hyphenat}
\definecolor{rred}{rgb}{0.65,0.23,0.25}
\definecolor{rbeige}{rgb}{0.66,0.45,0.23 }
\definecolor{rgreen}{rgb}{0.22,0.54,0.19}
\definecolor{lightgray}{rgb}{0.75,0.75,0.75}

\newcommand{\cropleft}{2}
\newcommand{\croplower}{2}
\newcommand{\cropright}{2}
\newcommand{\cropupper}{1.9}

\newcommand{\figskipabove}{2}
\newcommand{\figskipcaption}{2} 
\newcommand{\figskipbelow}{2} 

\renewcommand{\floatpagefraction}{0.7}
\begin{document}

\renewcommand{\floatpagefraction}{2}

\title{Learning 3D Shape Completion under Weak Supervision}
\author{David Stutz \and Andreas Geiger}

\institute{David Stutz \at
  Max Planck Institute for Informatics\\
  Campus E1 4, 66123 Saarbr\"{u}cken, Germany\\
  Tel.: +49 681 9325 2021\\
  \email{david.stutz@mpi-inf.mpg.de}\\
  \and
  Andreas Geiger \at
  Max Planck Institute for Intelligent Systems and University of T\"{u}bingen\\
  Max-Planck-Ring 4, 72076 T\"{u}bingen, Germany
}
\date{}

\maketitle

\begin{abstract}
We address the problem of 3D shape completion from sparse and noisy point clouds, a fundamental problem in computer vision and robotics. Recent approaches are either data-driven or learning-based: Data-driven approaches rely on a shape model whose parameters are optimized to fit the observations; Learning-based approaches, in contrast, avoid the expensive optimization step by learning to directly predict complete shapes from incomplete observations in a fully-supervised setting. However, full supervision is often not available in practice. In this work, we propose a weakly-supervised learning-based approach to 3D shape completion which neither requires slow optimization nor direct supervision. While we also learn a shape prior on synthetic data, we amortize, \ie, \emph{learn}, maximum likelihood fitting using deep neural networks resulting in efficient shape completion without sacrificing accuracy. On synthetic benchmarks based on ShapeNet \citep{Chang2015ARXIV} and ModelNet \citep{Wu2015CVPR} as well as on real robotics data from KITTI  \citep{Geiger2012CVPR} and \Kinect \citep{Yang2018ARXIVb}, we demonstrate that the proposed amortized maximum likelihood approach is able to compete with the fully supervised baseline of \cite{Dai2017CVPRa} and outperforms the data-driven approach of \cite{Engelmann2016GCPR}, while requiring less supervision and being significantly faster.
\end{abstract}
\section{Introduction}
\label{sec:introduction}

\begin{figure}[t]
    \vspace*{-16px}
    \vspace*{-\figskipabove px}
    \centering
    {\scriptsize
    \begin{subfigure}[t]{0.235\textwidth}
        \begin{subfigure}[t]{0.49\textwidth}
            \vspace{0px}\centering
            \includegraphics[width=2.25cm,trim={3cm 3cm 3cm 3cm},clip]{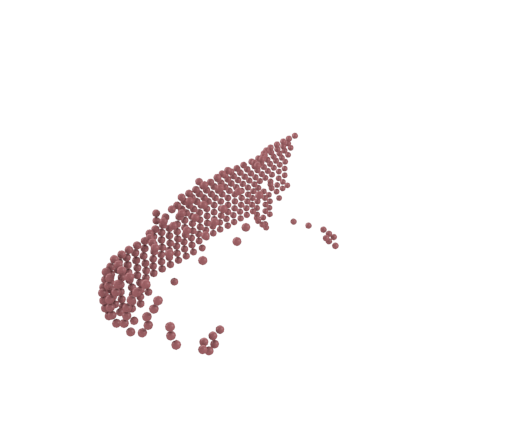}
        \end{subfigure}
        \begin{subfigure}[t]{0.49\textwidth}
            \vspace{0px}\centering
            \includegraphics[width=2.25cm,trim={3cm 3cm 3cm 3cm},clip]{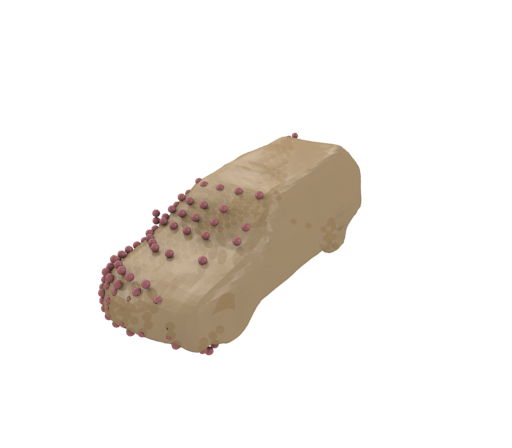}
        \end{subfigure}
        \vspace*{-2px}
        \subcaption{ShapeNet (Synthetic)}
    \end{subfigure}
    \begin{subfigure}[t]{0.235\textwidth}
        \begin{subfigure}[t]{0.49\textwidth}
            \vspace{0px}\centering
            \includegraphics[width=2.25cm,trim={3cm 3cm 3cm 3cm},clip]{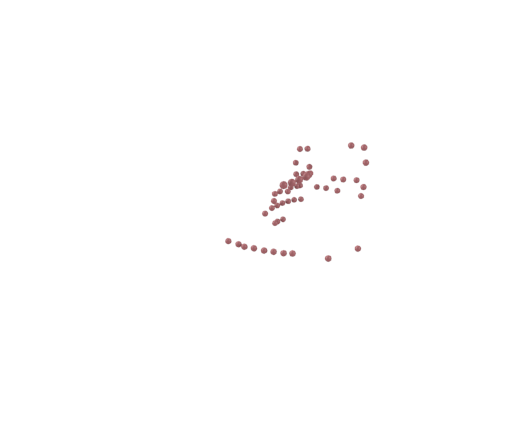}
        \end{subfigure}
        \begin{subfigure}[t]{0.49\textwidth}
            \vspace{0px}\centering
            \includegraphics[width=2.25cm,trim={3cm 3cm 3cm 3cm},clip]{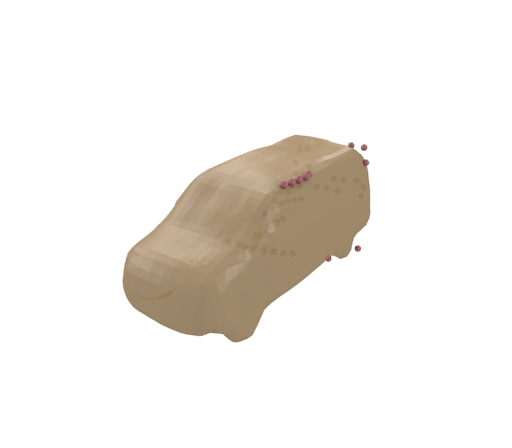}
        \end{subfigure}
        \subcaption{KITTI (Real)}
    \end{subfigure}
    \\[2px]
    \begin{subfigure}[t]{0.235\textwidth}
        \begin{subfigure}[t]{0.49\textwidth}
            \vspace{0px}\centering
            \includegraphics[width=2.25cm,trim={3cm 3cm 3cm 3cm},clip]{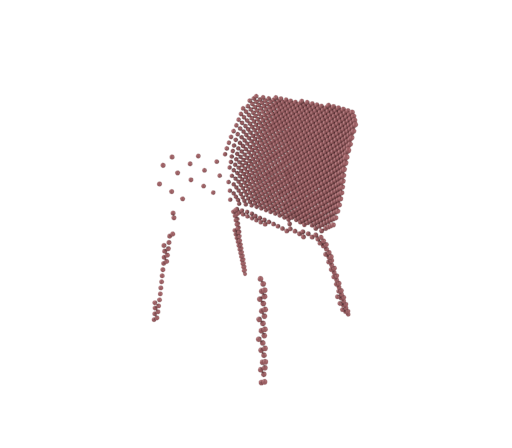}
        \end{subfigure}
        \begin{subfigure}[t]{0.49\textwidth}
            \vspace{0px}\centering
            \includegraphics[width=2.25cm,trim={3cm 3cm 3cm 3cm},clip]{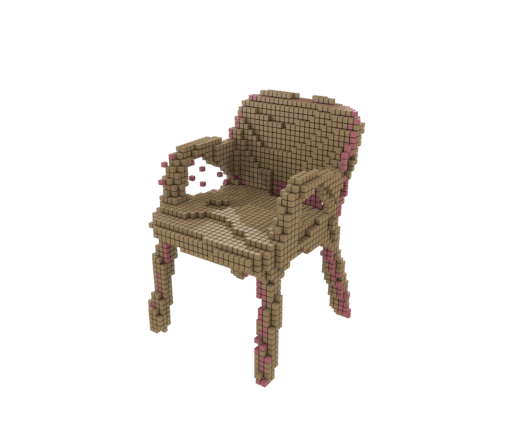}
        \end{subfigure}
        \subcaption{ModelNet (Synthetic)}
    \end{subfigure}
    \begin{subfigure}[t]{0.235\textwidth}
        \begin{subfigure}[t]{0.49\textwidth}
            \vspace{0px}\centering
            \includegraphics[width=2.25cm,trim={3cm 3cm 3cm 3cm},clip]{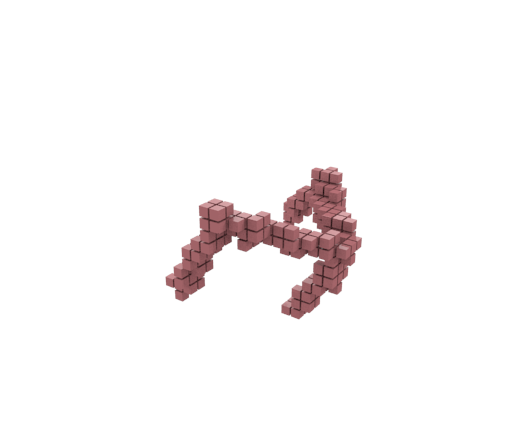}
        \end{subfigure}
        \begin{subfigure}[t]{0.49\textwidth}
            \vspace{0px}\centering
            \includegraphics[width=2.25cm,trim={3cm 3cm 3cm 3cm},clip]{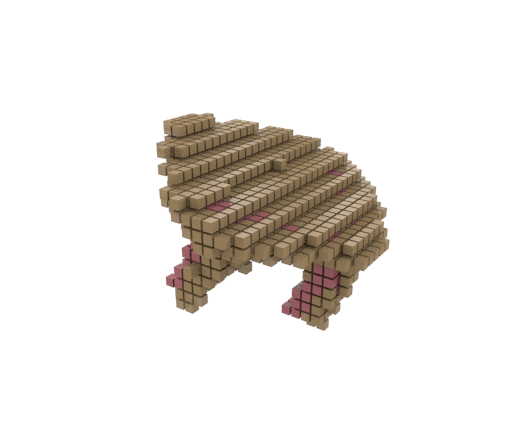}
        \end{subfigure}
        \subcaption{\Kinect (Real)}
    \end{subfigure}
    }
    \vspace*{-\figskipcaption px}
    \caption{{\bf 3D Shape Completion.} Results for cars on ShapeNet \citep{Chang2015ARXIV} and KITTI \citep{Geiger2012CVPR} and for chairs and tables on ModelNet \citep{Wu2015CVPR} and \Kinect \citep{Yang2018ARXIVb}. Learning shape completion on real-world data is challenging due to sparse and noisy observations and missing ground truth. Occupancy grids (bottom) or meshes from signed distance functions (SDFs, top) at various resolutions in {\color{rbeige}beige} and point cloud observations in {\color{rred}red}.}
    \label{fig:introduction}
    \vspace*{-\figskipbelow px}
\end{figure}
\begin{figure*}
	\vspace*{-4px}
    \centering
    \begin{tikzpicture}
        \node[anchor=west] at (-1.25, 3.75) {{\footnotesize{\bf(1) Shape Prior} (Section 3.2)}};
        
        \node[rectangle,draw=black,anchor=west] (prior) at (-1,2.5) {
            \includegraphics[height=1cm,trim={3cm 3cm 3cm 3cm},clip]{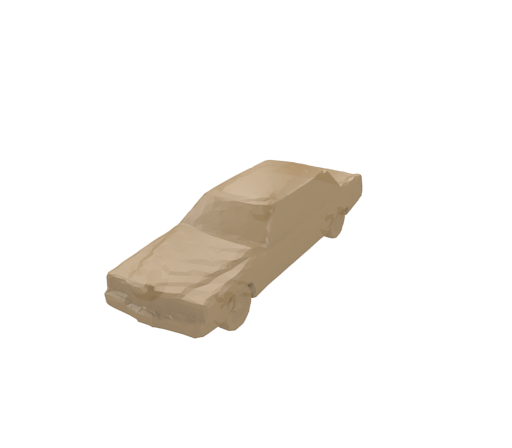}
            \includegraphics[height=1cm,trim={3cm 3cm 3cm 3cm},clip]{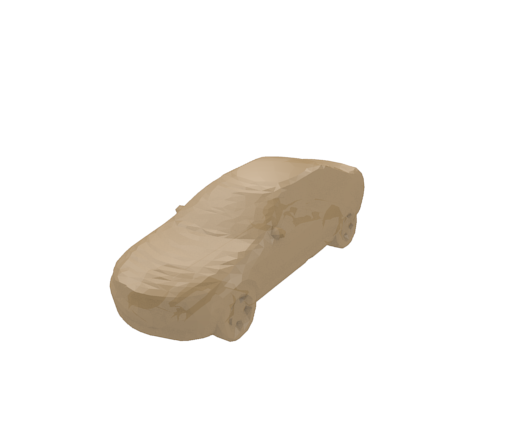}
            \includegraphics[height=1cm,trim={3cm 3cm 3cm 3cm},clip]{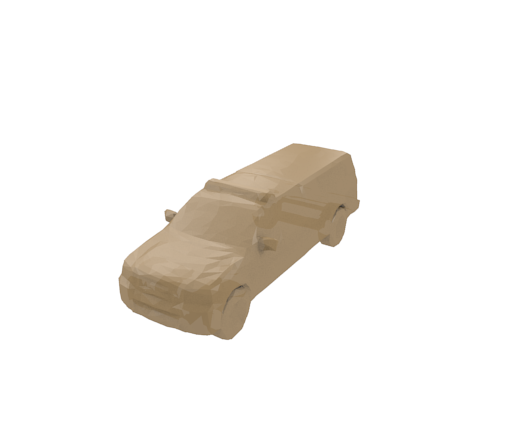}
            \includegraphics[height=1cm,trim={3cm 3cm 3cm 3cm},clip]{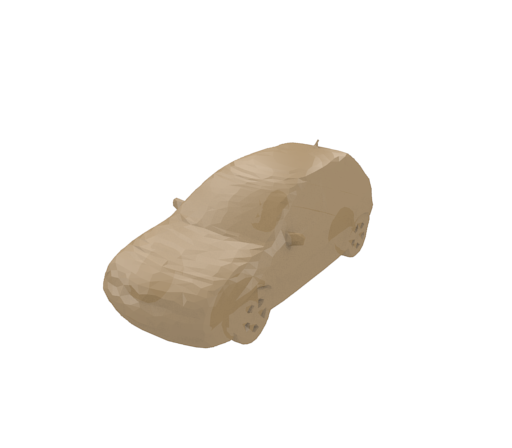}
        };
        \node at (1.6,3.3) {\scriptsize Synthetic Reference Shapes};
        
        \node[] (y) at (0, -1.5) {\scriptsize Shape $y$};
        
        \node[rectangle,draw=black,anchor=west] (input) at (-1, 0) {
            \begin{tabular}{c}
            \includegraphics[height=1cm,trim={3cm 3cm 3cm 3cm},clip]{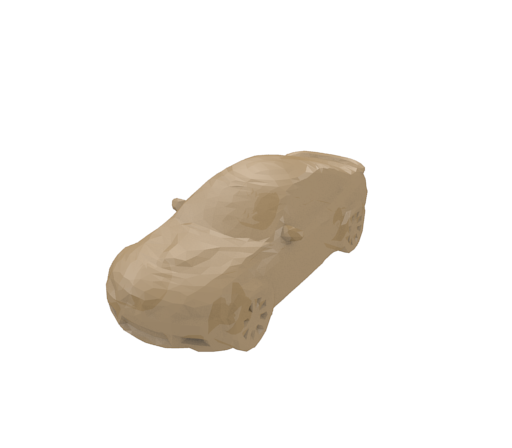}\\
            \includegraphics[height=1cm,trim={3cm 3cm 3cm 3cm},clip]{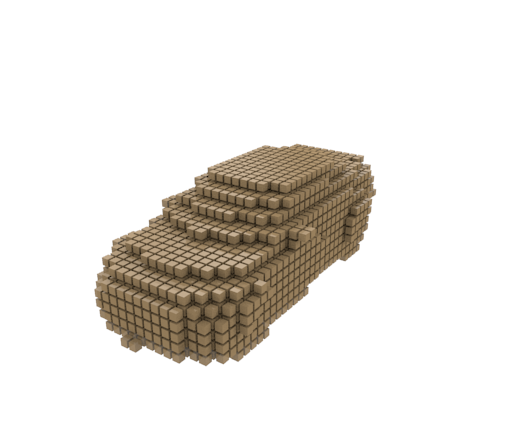}
            \end{tabular}
        };  
        
        \draw ($(input.north east) + (0.2,0)$) -- ($(input.north east) + (1.5,-0.75)$) -- ($(input.north east) + (1.5,-1.75)$) -- ($(input.south east) + (0.2,0)$) -- ($(input.north east) + (0.2,0)$);
        \node at ($(input.east) + (0.825,0)$) {\scriptsize encoder};
        
        \node (z) at (2.75, 0) {\scriptsize $z$};
        
        \node[rectangle,draw=black,anchor=east] (output) at (6.5, 0) {
            \begin{tabular}{c}
            \includegraphics[height=1cm,trim={3cm 3cm 3cm 3cm},clip]{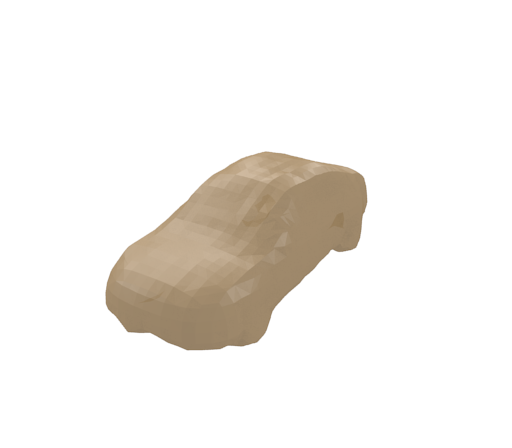}\\
            \includegraphics[height=1cm,trim={3cm 3cm 3cm 3cm},clip]{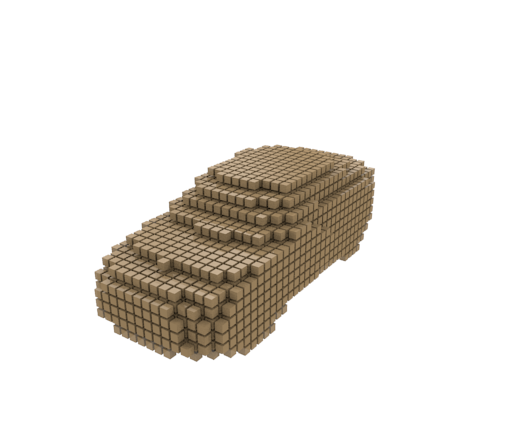}
            \end{tabular}
        };
        
        \draw ($(output.north west) - (0.2,0)$) -- ($(output.north west) - (1.5,0.75)$) -- ($(output.north west) - (1.5,1.75)$) -- ($(output.south west) - (0.2,0)$) -- ($(output.north west) - (0.2,0)$);
        \node at ($(output.west) - (0.825,0)$) {\scriptsize decoder};
        
        \node (ry) at (5.5, -1.5) {\scriptsize Rec. Shape $\tilde{y}$};
        
        \node[] (L) at (2.75, -1.9) {\scriptsize Reconstruction Loss};
        
        \draw[-] (ry) -- ($(ry) - (0,0.4)$);
        \draw[-] ($(ry) - (0,0.4)$) -- (L);
        \draw[-] (y) -- ($(y) - (0,0.4)$);
        \draw[-] ($(y) - (0,0.4)$) -- (L);
        
        \draw[-] (3.5, 0.85) -- (3.5, 1.5);
        \draw[-] (3.5, 1.5) -- (12.5, 1.5);
        \draw[->] (12.5, 1.5) -- (12.5, 0.85);
        \node at (7.25, 1.75) {\scriptsize retain fixed decoder};
        
        \draw[-,dotted] (7.25,-2.15) -- (7.25,1.45);
        \draw[-,dotted] (7.25,2) -- (7.25,2.45);
        \draw[-,dotted] (7.25,3) -- (7.25,4.05);
        
        \node at (7.25, 2.75) {\scriptsize\textbf{no correspondence needed}};
        \begin{scope}[shift={(9,0)}]
        
        \node[anchor=west] at (-1.25, 3.75) {{\footnotesize{\bf(2) Shape Inference} (Section 3.3)}};
        
        \node[rectangle,draw=black,anchor=east] (inference) at (6.5,2.5) {
            \includegraphics[height=1cm,trim={3cm 3cm 3cm 3cm},clip]{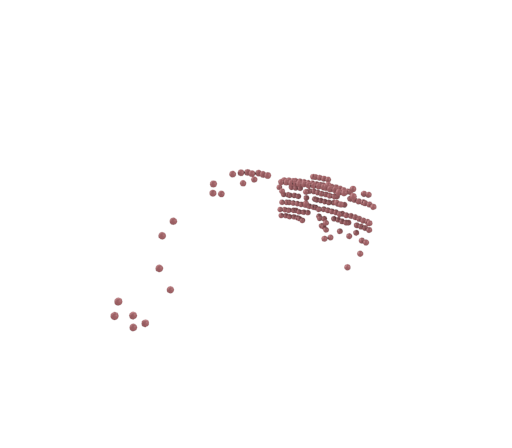}
            \includegraphics[height=1cm,trim={3cm 3cm 3cm 3cm},clip]{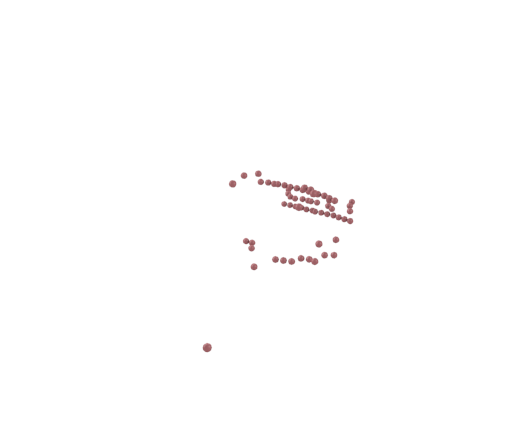}
            \includegraphics[height=1cm,trim={3cm 3cm 3cm 3cm},clip]{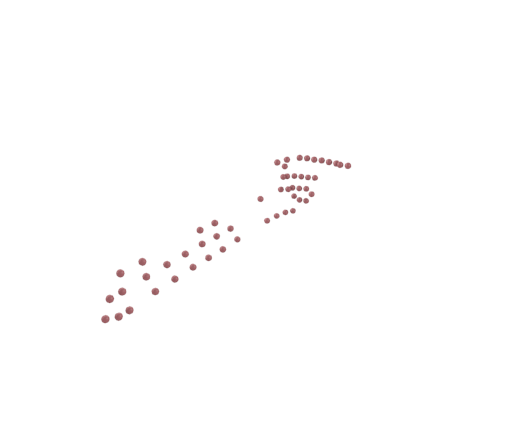}
            \includegraphics[height=1cm,trim={3cm 3cm 3cm 3cm},clip]{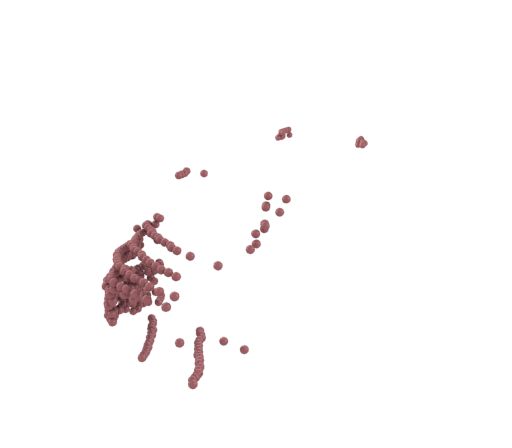}
        };
        \node at (3.8,3.3) {\scriptsize Real Observations w/o Targets};
        
        \draw[-] (prior) -- (inference);
        
        \node[] (y) at (0, -1.5) {\scriptsize Observation $x$};
        
        \node[rectangle,draw=black,anchor=west] (input) at (-1, 0) {
            \begin{tabular}{c}
            \includegraphics[height=1cm,trim={3cm 3cm 3cm 3cm},clip]{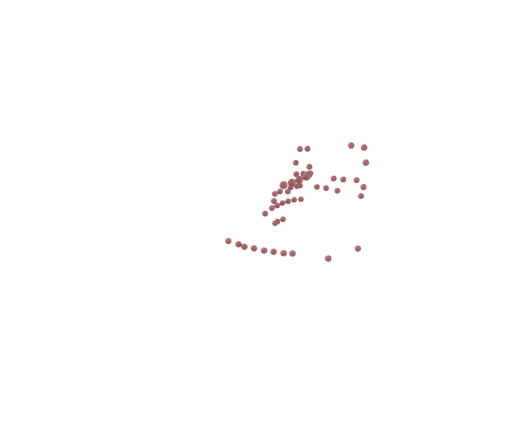}\\
            \includegraphics[height=1cm,trim={3cm 3cm 3cm 3cm},clip]{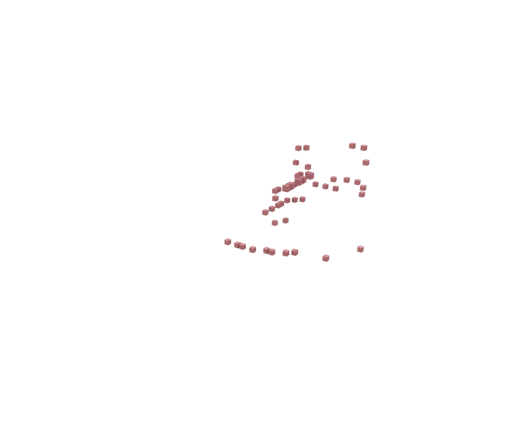}
            \end{tabular}
        };
        
        \draw ($(input.north east) + (0.2,0)$) -- ($(input.north east) + (1.5,-0.75)$) -- ($(input.north east) + (1.5,-1.75)$) -- ($(input.south east) + (0.2,0)$) -- ($(input.north east) + (0.2,0)$);
        \node at ($(input.east) + (0.825,0)$) {\scriptsize\begin{tabular}{c}new\\encoder\end{tabular}};
        
        \node (z) at (2.75, 0) {\scriptsize $z$};
        
        \node[rectangle,draw=black,anchor=east] (output) at (6.5, 0) {
            \begin{tabular}{c}
            \includegraphics[height=1cm,trim={3cm 3cm 3cm 3cm},clip]{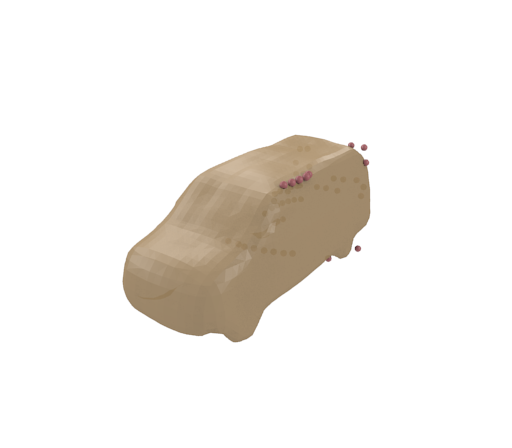}\\
            \includegraphics[height=1cm,trim={3cm 3cm 3cm 3cm},clip]{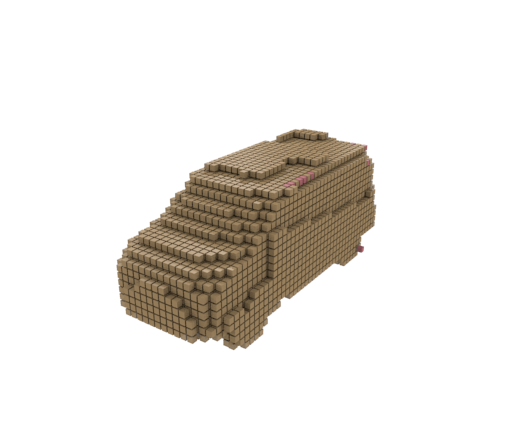}
            \end{tabular}
        };
        
        \draw ($(output.north west) - (0.2,0)$) -- ($(output.north west) - (1.5,0.75)$) -- ($(output.north west) - (1.5,1.75)$) -- ($(output.south west) - (0.2,0)$) -- ($(output.north west) - (0.2,0)$);
        \node at ($(output.west) - (0.825,0)$) {\scriptsize \begin{tabular}{c}fixed\\decoder\end{tabular}};
        
        \node (ry) at (5.5, -1.5) {\scriptsize Prop. Shape $\tilde{y}$};
        
        \node[] (L) at (2.75, -1.9) {\scriptsize Maximum Likelihood Loss};
        
        \draw[-] (ry) -- ($(ry) - (0,0.4)$);
        \draw[-] ($(ry) - (0,0.4)$) -- (L);
        \draw[-] (y) -- ($(y) - (0,0.4)$);
        \draw[-] ($(y) - (0,0.4)$) -- (L);
        \end{scope}
    \end{tikzpicture}
    \vspace*{-12px}
    \caption{{{\bf Amortized Maximum Likelihood (AML) for 3D Shape Completion on KITTI.}
            (1) We train a denoising variational auto-encoder (\DVAE) \citep{Kingma2014ICLR,Im2017AAAI} as shape prior on ShapeNet using occupancy grids and signed distance functions (SDFs) to represent shapes. (2) The fixed generative model, \ie, decoder, then allows to learn shape completion using an unsupervised maximum likelihood (\ML) loss by training a new recognition model, \ie, encoder. The retained generative model constraints the space of possible shapes while the \ML loss aligns the predicted shape with the observations.}}
    \label{fig:method}
    \vspace*{-\figskipbelow px}
    \vspace*{\figskipbelow px}
\end{figure*}
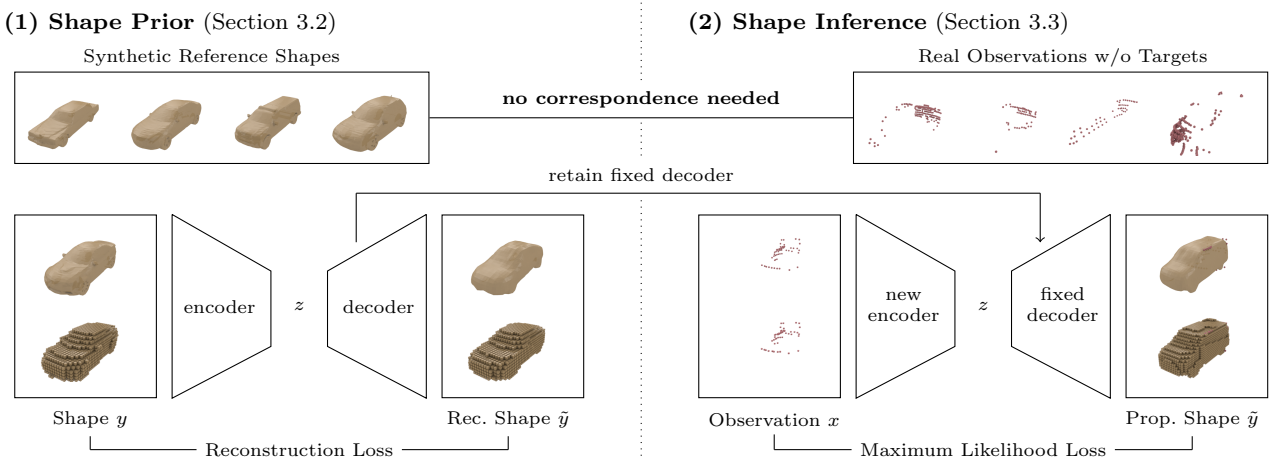

3D shape perception is a long-standing and fundamental problem both in human and computer vision \citep{Pizlo2007CAIP,Pizlo2010,Furukawa2013FTCGV} with many applications to robotics. A large body of work focuses on 3D reconstruction, \eg, reconstructing objects or scenes from one or multiple views, which is an inherently ill-posed inverse problem where many configurations of shape, color, texture and lighting may result in the very same image. While the primary goal of human vision is to understand how the human visual system accomplishes such tasks, research in computer vision and robotics is focused on the task of devising 3D reconstruction systems. Generally, work by \cite{Pizlo2010} suggests that the constraints and priors used for 3D perception are innate and not learned. Similarly, in computer vision, cues and priors are commonly built into 3D reconstruction pipelines through explicit assumptions. Recently, however -- leveraging the success of deep learning -- researchers started to \emph{learn} shape models from large collections of data, as for example ShapeNet~\citep{Chang2015ARXIV}. Predominantly generative models have been used to learn how to generate, manipulate and reason about 3D shapes  \citep{Girdhar2016ECCV,Brock2016ARXIV,Sharma2016ARXIV,Wu2016NIPS,Wu2015CVPR}.

In this paper, we focus on the specific problem of inferring and completing 3D shapes based on sparse and  noisy 3D point observations as illustrated in \figref{fig:introduction}. This problem occurs when only a single view of an individual object is provided or large parts of the object are occluded as common in robotic applications. For example, autonomous vehicles are commonly equipped with LiDAR scanners providing a 360 degree point cloud of the surrounding environment in real-time. This point cloud is inherently incomplete: back and bottom of objects are typically occluded and -- depending on material properties -- the observations are sparse and noisy, see \figref{fig:introduction} (top-right) for an illustration. Similarly, indoor robots are generally equipped with low-cost, real-time RGB-D sensors providing noisy point clouds of the observed scene. In order to make informed decisions (\eg, for path planning and navigation), it is of utmost importance to efficiently establish a representation of the environment which is as complete as possible.

Existing approaches to 3D shape completion can be categorized into data-driven and learning-based methods. The former usually rely on learned shape priors and formulate shape completion as an optimization problem over the corresponding (lower-dimensional) latent space \citep{Rock2015CVPR,Haene2014CVPR,Li2015CGF,Engelmann2016GCPR,Nan2012TG,Bao2013CVPR,Dame2013CVPR,Ngyuen2016CVPR}. These approaches have demonstrated good performance on real data, \eg, on KITTI \citep{Geiger2012CVPR}, but are often slow in practice.

Learning-based approaches, in contrast, assume a fully supervised setting in order to directly learn shape completion on synthetic data \citep{Riegler2017THREEDV,Smith2017ARXIV,Dai2017CVPRa,Sharma2016ARXIV,Fan2017CVPR,Rezende2016ARXIV,Yang2018ARXIVb,Wang2017ICCV,Varley2017IROS,Han2017ICCV}. They offer advantages in terms of efficiency as prediction can be performed in a single forward pass, however, require full supervision during training. Unfortunately, even multiple, aggregated observations (\eg, from multiple views) will not be fully complete due to occlusion, sparse sampling of views and noise, see \figref{fig:results-real} (right column) for an example.

In this paper, we propose an amortized maximum likelihood approach for 3D shape completion (\cf \figref{fig:method}) avoiding the slow optimization problem of data-driven approaches and the required supervision of learning-based approaches. Specifically, we first learn a shape prior on synthetic shapes using a (denoising) variational auto-encoder \citep{Im2017AAAI,Kingma2014ICLR}. Subsequently, 3D shape completion can be formulated as a maximum likelihood problem. However, instead of maximizing the likelihood independently for distinct observations, we follow the idea of amortized inference \citep{Gersham2014COGSCI} and \emph{learn} to predict the maximum likelihood solutions directly. Towards this goal, we train a new encoder which embeds the observations in the same latent space using an unsupervised maximum likelihood loss. This allows us to learn 3D shape completion in challenging real-world situations, \eg, on KITTI, and obtain sub-voxel accurate results using signed distance functions at resolutions up to $64^3$ voxels. For experimental evaluation, we introduce two novel, synthetic shape completion benchmarks based on ShapeNet and ModelNet \citep{Wu2015CVPR}. We compare our approach to the data-driven approach by \cite{Engelmann2016GCPR}, a baseline inspired by \cite{Gupta2015CVPR} and the fully-supervised learning-based approach by \cite{Dai2017CVPRa}; we additionally present experiments on real data from KITTI and \Kinect \citep{Yang2018ARXIVb}. Experiments show that our approach outperforms data-driven techniques and rivals learning-based techniques while significantly reducing inference time and using only a fraction of supervision.

A preliminary version of this work has been published at CVPR'18 \citep{Stutz2018CVPR}. However, we improved the proposed shape completion method, the constructed datasets and present more extensive experiments. In particular, we extended our weakly-supervised amortized maximum likelihood approach to enforce more variety and increase visual quality significantly. On ShapeNet and ModelNet, we use volumetric fusion to obtain more detailed, watertight meshes and manually selected -- per object-category -- $220$ high-quality models to synthesize challenging observations. We additionally increased the spatial resolution and consider two additional baselines \citep{Dai2017CVPRa,Gupta2015CVPR}. Our code and datasets will be made publicly available\footnote{\url{https://avg.is.tuebingen.mpg.de/research_projects/3d-shape-completion}.}.

The paper is structured as follows: We discuss related work in \secref{sec:related-work}. In \secref{sec:method} we introduce the weakly-supervised shape completion problem and describe the proposed amortized maximum likelihood approach. Subsequently, we introduce our synthetic shape completion benchmarks and discuss the data preparation for KITTI and \Kinect in \secref{sec:data}. Next, we discuss evaluation in \secref{sec:evaluation}, our training procedure in \secref{sec:training}, and the evaluated baselines in \secref{sec:baselines}. Finally, we present experimental results in \secref{sec:experiments} and conclude in \secref{sec:conclusion}.
\section{Related Work}
\label{sec:related-work}

\subsection{3D Shape Completion and Single-View 3D Reconstruction}

In general, 3D shape completion is a special case of single-view 3D reconstruction where we assume point cloud observations to be available, \eg from laser-based sensors as on KITTI \citep{Geiger2012CVPR}.

\boldparagraph{3D Shape Completion}
Following \cite{Sung2015TG}, classical shape completion approaches can roughly be categorized into symmetry-based methods and data-driven methods. The former leverage observed symmetry to complete shapes; representative works include \citep{Thrun2005ICCV,Pauly2008TG,Zheng2010TG,Kroemer2012HUMANOIDS,Law2011CVIU}. Data-driven approaches, in contrast, as pioneered by \cite{Pauly2005SGP}, pose shape completion as retrieval and alignment problem. While \cite{Pauly2005SGP} allow shape deformations, \cite{Gupta2015CVPR}, use the iterative closest point (\ICP) algorithm \citep{Besl1992PAMI} for fitting rigid shapes. Subsequent work usually avoids explicit shape retrieval by learning a latent space of shapes \citep{Rock2015CVPR,Haene2014CVPR,Li2015CGF,Engelmann2016GCPR,Nan2012TG,Bao2013CVPR,Dame2013CVPR,Ngyuen2016CVPR}. Alignment is then formulated as optimization problem over the learned, low-dimensional latent space. For example, \cite{Bao2013CVPR} parameterize the shape prior through anchor points with respect to a mean shape, while \cite{Engelmann2016GCPR} and \cite{Dame2013CVPR} directly learn the latent space using principal component analysis and Gaussian process latent variable models \citep{Prisacariu2011CVPR}, respectively. In these cases, shapes are usually represented by signed distance functions (SDFs). \cite{Ngyuen2016CVPR} use 3DShapeNets \citep{Wu2015CVPR}, a deep belief network trained on occupancy grids, as shape prior. In general, data-driven approaches are applicable to real data assuming knowledge about the object category. However, inference involves a possibly complex optimization problem, which we avoid by amortizing, \ie, \emph{learning}, the inference procedure. Additionally, we also consider multiple object categories.

With the recent success of deep learning, several learning-based approaches have been proposed \citep{Firman2016CVPR,Smith2017ARXIV,Dai2017CVPRa,Sharma2016ARXIV,Rezende2016ARXIV,Fan2017CVPR,Riegler2017THREEDV,Han2017ICCV,Yang2017ARXIV,Yang2018ARXIVb}. Strictly speaking, these are data-driven, as well; however, shape retrieval and fitting are \emph{both} avoided by directly learning shape completion end-to-end, under full supervision -- usually on synthetic data from ShapeNet \citep{Chang2015ARXIV} or ModelNet \citep{Wu2015CVPR}. \cite{Riegler2017THREEDV} additionally leverage octrees to predict higher-resolution shapes; most other approaches use low resolution occupancy grids (\eg, $32^3$ voxels). Instead, \cite{Han2017ICCV} use a patch-based approach to obtain high-resolution results. In practice, however, full supervision is often not available; thus, existing models are primarily evaluated on synthetic datasets. In order to learn shape completion without full supervision, we utilize a learned shape prior to constrain the space of possible shapes. In addition, we use SDFs to obtain sub-voxel accuracy at higher resolutions (up to $48 \ntimes 108 \ntimes 48$ or $64^3$ voxels) without using patch-based refinement or octrees. We also consider significantly sparser observations.

\boldparagraph{Single-View 3D Reconstruction}
Single-view 3D reconstruction has received considerable attention over the last years; we refer to \citep{Oswal2013ISAMA} for an overview and focus on recent deep learning approaches, instead. Following \cite{Tulsiani2018ARXIV}, these can be categorized by the level of supervision. For example, \citep{Girdhar2016ECCV,Choy2016ECCV,Wu2016NIPS,Haene2017ARXIV} require full supervision, \ie, pairs of images and ground truth 3D shapes. These are generally derived synthetically. More recent work \citep{Yan2016NIPS,Tulsiani2017CVPR,Tulsiani2018ARXIV,Kato2017ARXIV,Lin2017ARXIV,Fan2017CVPR,Tatarchenko2017ICCV,Wu2016ECCV}, in contrast, self-supervise the problem by enforcing consistency across multiple input views. \cite{Tulsiani2018ARXIV}, for example, use a differentiable ray consistency loss; and in \citep{Yan2016NIPS,Kato2017ARXIV,Lin2017ARXIV}, differentiable rendering allows to define reconstruction losses on the images directly. While most of these approaches utilize occupancy grids, \cite{Fan2017CVPR} and \cite{Lin2017ARXIV} predict point clouds instead. \cite{Tatarchenko2017ICCV} use octrees to predict higher-resolution shapes. Instead of employing multiple views as weak supervision, however, we do not assume any additional views in our approach. Instead, knowledge about the object category is sufficient. In this context, concurrent work by \cite{Gwak2017ARXIV} is more related to ours: a set of reference shapes implicitly defines a prior of shapes which is enforced using an adversarial loss. In contrast, we use a denoising variational auto-encoder (\DVAE) \citep{Kingma2014ICLR,Im2017AAAI} to explicitly learn a prior for 3D shapes.

\subsection{Shape Models}

Shape models and priors found application in a wide variety of different tasks. In 3D reconstruction, in general, shape priors are commonly used to resolve ambiguities or specularities \citep{Dame2013CVPR,Guney2015CVPR,Kar2015CVPR}. Furthermore, pose estimation \citep{Sandhu2011PAMI,Sandhu2009CVPR,Prisacariu2013ACCV,Aubry2014CVPR}, tracking \citep{Ma2014ECCV,Leotta2009CVPR}, segmentation \citep{Sandhu2011PAMI,Sandhu2009CVPR,Prisacariu2013ACCV}, object detection \citep{Zia2013PAMI,Zia2014CVPR,Pepik2015CVPR,Song2014ECCV,Zheng2015GCPR} or recognition \citep{Lin2014ECCV} -- to name just a few -- have been shown to benefit from shape models. While most of these works use hand-crafted shape models, for example based on anchor points or part annotations \citep{Zia2013PAMI,Zia2014CVPR,Pepik2015CVPR,Lin2014ECCV}, recent work \citep{Liu2017ARXIV,Sharma2016ARXIV,Girdhar2016ECCV,Wu2016NIPS,Wu2015CVPR,Smith2017ARXIV,Nash2017SGP,Liu2017ARXIV} has shown that generative models such as \VAEs \citep{Kingma2014ICLR} or generative adversarial networks (\GANs) \citep{Goodfellow2014NIPS} allow to efficiently generate, manipulate and reason about 3D shapes. We use these more expressive models to obtain high-quality shape priors for various object categories.

\subsection{Amortized Inference}

To the best of our knowledge, the notion of amortized inference was introduced by \cite{Gersham2014COGSCI} and picked up repeatedly in different contexts \citep{Rezende2015ICML,Wang2016ARXIV,Ritchie2016ARXIV}. Generally, it describes the idea of \emph{learning to infer} (or learning to sample). We refer to \citep{Wang2016ARXIV} for a broader discussion of related work. In our context, a \VAE can be seen as specific example of learned variational inference \citep{Kingma2014ICLR,Rezende2015ICML}. Besides using a \VAE as shape prior, we also amortize the maximum likelihood problem corresponding to our 3D shape completion task.
\section{Method}
\label{sec:method}

In the following, we introduce the mathematical formulation of the weakly-supervised 3D shape completion problem. Subsequently, we briefly discuss denoising variational auto-encoders (\DVAEs) \citep{Kingma2014ICLR,Im2017AAAI} which we use to learn a strong shape prior \red{that embeds a set of reference shapes in a low-dimensional latent space.} Then, we formally derive our proposed amortized maximum likelihood (\AML) approach. \red{Here, we use maximum likelihood to learn an embedding of the observations within the same latent space -- thereby allowing to perform shape completion.} The overall approach is also illustrated in \figref{fig:method}.

\subsection{Problem Formulation}
\label{subsec:method-problem}

\newcommand{\shapeneta}{330} 
\newcommand{\shapenetb}{66}
\newcommand{\shapenetc}{132}
\newcommand{\shapenetd}{198}
\newcommand{\shapenete}{726}

\begin{figure}
    \vspace*{-\figskipabove px}
    \centering
    
    \begin{subfigure}[t]{0.5\textwidth}
        \begin{subfigure}{0.19\textwidth}
            \centering\vspace{0px}
            \includegraphics[height=1.5cm,trim={\cropleft cm \croplower cm \cropright cm \cropupper cm},clip]{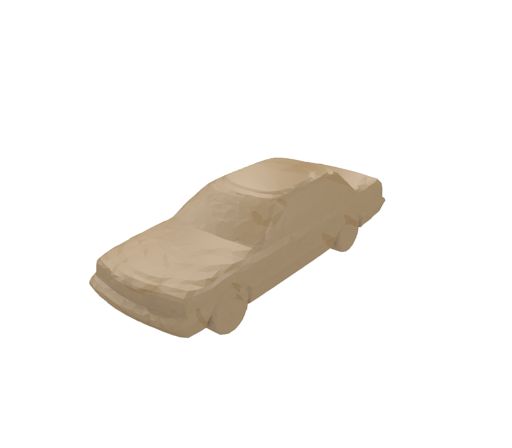}
        \end{subfigure}
        \begin{subfigure}{0.19\textwidth}
            \centering\vspace{0px}
            \includegraphics[height=1.5cm,trim={\cropleft cm \croplower cm \cropright cm \cropupper cm},clip]{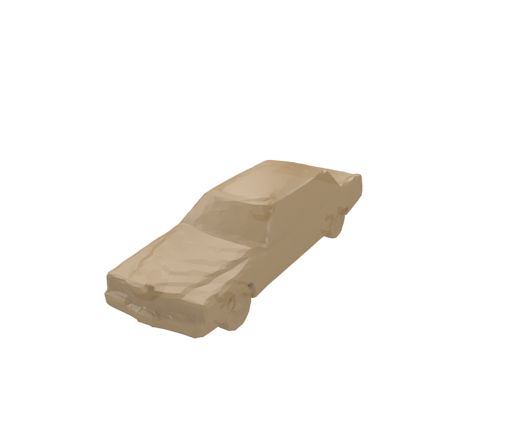}
        \end{subfigure}
        \begin{subfigure}{0.19\textwidth}
            \centering\vspace{0px}\textbf{}
            \includegraphics[height=1.5cm,trim={\cropleft cm \croplower cm \cropright cm \cropupper cm},clip]{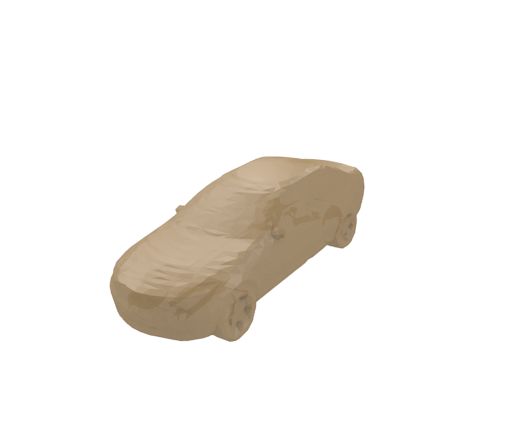}
        \end{subfigure}
        \begin{subfigure}{0.19\textwidth}
            \centering\vspace{0px}
            \includegraphics[height=1.5cm,trim={\cropleft cm \croplower cm \cropright cm \cropupper cm},clip]{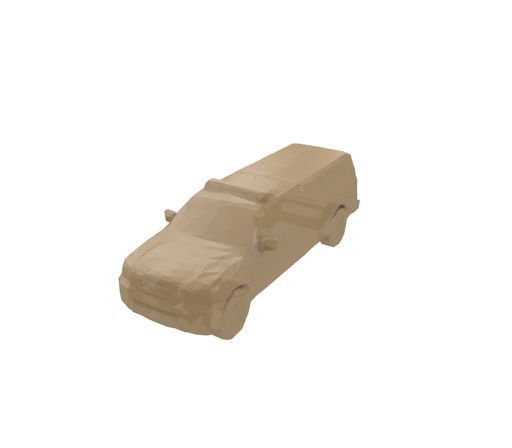}
        \end{subfigure}
        \begin{subfigure}{0.19\textwidth}
            \centering\vspace{0px}
            \includegraphics[height=1.5cm,trim={\cropleft cm \croplower cm \cropright cm \cropupper cm},clip]{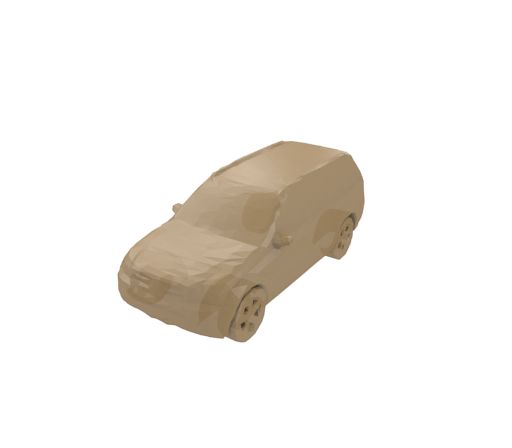}
        \end{subfigure}
        \subcaption{Reference Shapes $\mathcal{Y}$}
    \end{subfigure}
    \\
    \begin{subfigure}[t]{0.29\textwidth}
        \begin{subfigure}[t]{0.49\textwidth}
            \centering\vspace{0px}
            \includegraphics[height=2cm,trim={\cropleft cm \croplower cm \cropright cm \cropupper cm},clip]{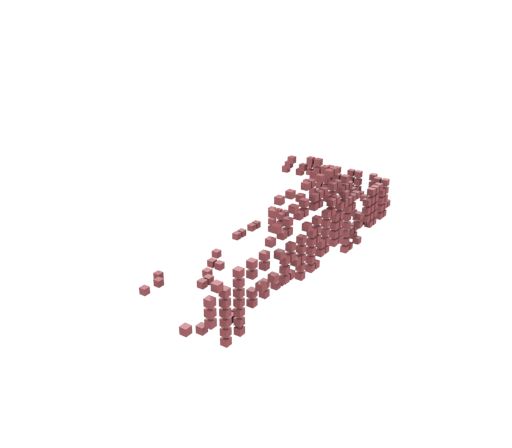}
        \end{subfigure}
        \begin{subfigure}[t]{0.49\textwidth}
            \centering\vspace{0px}
            \includegraphics[height=2cm,trim={\cropleft cm \croplower cm \cropright cm \cropupper cm},clip]{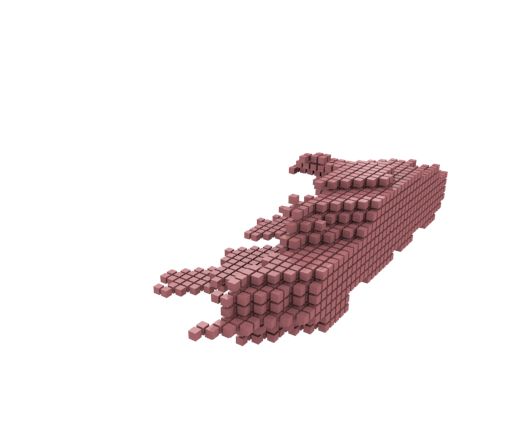}
        \end{subfigure}
        \subcaption{Observation $x_n$}
    \end{subfigure}
    \hfill
    \begin{subfigure}[t]{0.18\textwidth}
        \centering\vspace{0px}
        \includegraphics[height=2cm,trim={\cropleft cm \croplower cm \cropright cm \cropupper cm},clip]{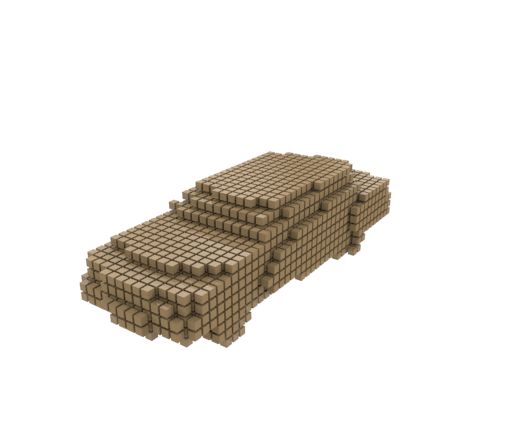}
        \subcaption{Ground Truth $y_n^*$}
    \end{subfigure}
    \vspace*{-\figskipcaption px}
    \caption{{\bf Weakly-Supervised Shape Completion.} Given reference shapes $\mathcal{Y}$ and incomplete observations $\mathcal{X}$, we want to learn a mapping $x_n \mapsto \tilde{y}(x_n)$ such that $\tilde{y}(x_n)$ matches the \emph{unknown} ground truth shape $y_n^*$ as close as possible. The observations $x_n$ are split into free space (\ie, $x_{n,i} = 0$, right) and point observations (\ie, $x_{n,i} = 1$, left). Shapes are shown in {\color{rbeige}beige} and observations in {\color{rred}red}.}
    \label{fig:method-problem}
    \vspace*{-\figskipbelow px}
\end{figure}

In a supervised setting, the task of 3D shape completion can be described as follows: Given a set of incomplete observations $\mathcal{X} = \{x_n\}_{n = 1}^N \subseteq \mathbb{R}^R$ and corresponding ground truth shapes $\mathcal{Y}^* = \{y_n^*\}_{n = 1}^N \subseteq \mathbb{R}^R$, learn a mapping $x_n \mapsto y_n^*$ that is able to generalize to previously unseen observations and possibly across object categories. We assume $\mathbb{R}^R$ to be a suitable representation of observations and shapes; in practice, we resort to occupancy grids and signed distance functions (SDFs) defined on regular grids, \ie, $x_n, y_n^* \in \mathbb{R}^{H \ntimes W \ntimes D} \simeq \mathbb{R}^R$. Specifically, occupancy grids indicate occupied space, \ie, voxel $y_{n,i}^* = 1$ if and only if the voxel lies on or inside the shape's surface. To represent shapes with sub-voxel accuracy, SDFs hold the distance of each voxel's center to the surface; for voxels inside the shape's surface, we use negative sign.
Finally, for the (incomplete) observations, we write $x_n \in \{0, 1, \uk\}^R$ to make missing information explicit; in particular, $x_{n,i} = \uk$ corresponds to unobserved voxels, while $x_{n,i} = 1$ and $x_{n,i} = 0$ correspond to occupied and unoccupied voxels, respectively.

On real data, \eg, KITTI \citep{Geiger2012CVPR}, supervised learning is often not possible as obtaining ground truth annotations is labor intensive, \cf \citep{Menze2015CVPR,Xie2016CVPR}. Therefore, we target a weakly-supervised variant of the problem instead: Given observations $\mathcal{X}$ and reference shapes $\mathcal{Y} = \{y_m\}_{m = 1}^M \subseteq \mathbb{R}^R$ both of the same, known object category, learn a mapping $x_n \mapsto \tilde{y}(x_n)$ such that the predicted shape $\tilde{y}(x_n)$ matches the unknown ground truth shape $y_n^*$ as close as possible -- or, in practice, the sparse observation $x_n$ while being plausible considering the set of reference shapes, \cf \figref{fig:method-problem}. Here, supervision is provided in the form of the known object category. Alternatively, the reference shapes $\mathcal{Y}$ can also include multiple object categories resulting in an even weaker notion of supervision as the correspondence between observations and object categories is unknown. \red{Except for the object categories, however, the set of reference shapes $\mathcal{Y}$, and its size $M$, is completely independent of the set of observations $\mathcal{X}$, and its size $N$, as also highlighted in \figref{fig:method}.} On real data, \eg, KITTI, we additionally assume the object locations to be given in the form of 3D bounding boxes in order to extract the corresponding observations $\mathcal{X}$. In practice, the reference shapes $\mathcal{Y}$ are derived from watertight, triangular meshes, \eg, from ShapeNet \citep{Chang2015ARXIV} or ModelNet \citep{Wu2015CVPR}.

\subsection{Shape Prior}
\label{subsec:method-prior}

\red{We approach the weakly-supervised shape completion problem by first learning a shape prior using a denoising variational auto-encoder (\DVAE). Later, this prior constrains shape inference (see \secref{subsec:method-inference}) to predict reasonable shapes. In the following, we briefly discuss the standard variational auto-encoder (\VAE), as introduced by \cite{Kingma2014ICLR}, as well as its denoising extension, as proposed by \cite{Im2017AAAI}.}

\boldparagraph{Variational Auto-Encoder (\VAE)}
We propose to use the provided reference shapes $\mathcal{Y}$ to learn a \red{generative} model of possible 3D shapes over a low-dimensional latent space $\mathcal{Z} = \mathbb{R}^Q$, \ie,  $Q \ll R$. \red{In the framework of \VAEs, the joint distribution $p(y, z)$ of shapes $y$ and latent codes $z$} decomposes into $p(y | z)p(z)$ with $p(z)$ being a unit Gaussian, \ie, $\mathcal{N}(z;0, I_Q)$ and $I_Q \in \mathbb{R}^{R \times R}$ being the identity matrix. \red{This decomposition allows to sample $z \sim p(z)$ and $y \sim p(y|z)$ to generate random shapes.}
\red{For training, however, we additionally need to approximate the posterior $p(z | y)$.} \red{To this end, the so-called recognition model $q(z | y) \approx p(z | y)$} takes the form
\begin{align}
q(z | y) &= \mathcal{N}(z; \mu(y), \text{diag}(\sigma^2(y)))\label{eq:encoder-decoder}
\end{align}
where $\mu(y), \sigma^2(y) \in \mathbb{R}^Q$ are predicted using the encoder neural network. \red{The generative model $p(y|z)$ decomposes over voxels $y_i$; the corresponding probabilities $p(y_i | z)$ are represented using Bernoulli distributions for occupancy grids or Gaussian distributions for SDFs:}
\begin{align}
    \begin{split}
        p(y_i | z) &= \text{Ber}(y_i ; \theta_i(z))\quad\text{or}\\
        p(y_i | z) &= \mathcal{N}(y_i ; \mu_i(z), \sigma^2).\label{eq:decoder}
    \end{split}
\end{align}
In both cases, the parameters, \ie, $\theta_i(z)$ or $\mu_i(z)$, are predicted using the decoder neural network. \red{For SDFs, we explicitly set $\sigma^2$ to be constant (see \secref{sec:training}). Then, $\sigma^2$ merely scales the corresponding loss, thereby implicitly defining the importance of accurate SDFs relative to occupancy grids as described below.}

In the framework of variational inference, the parameters of the encoder and the decoder \red{neural networks} are found by maximizing the likelihood $p(y)$. \red{In practice, the likelihood is usually intractable and the evidence lower bound is maximized instead, see \citep{Kingma2014ICLR,Blei2016ARXIV}. This results in the following loss to be minimized:}
\begin{align}
\mathcal{L}_{\text{VAE}}(w) = - \mathbb{E}_{q(z |y)}[\ln p(y|z)] + \text{KL}(q(z | y)| p(z)).\label{eq:vae}
\end{align}
\red{Here,} $w$ are the weights of the encoder and decoder \red{hidden in the recognition model $q(z | y)$ and the generative model $p(y | z)$, respectively}. \red{The Kullback-Leibler divergence $\text{KL}$ can be computed analytically as described in the appendix of \citep{Kingma2014ICLR}.} The negative log-likelihood $-\ln p(y|z)$ corresponds to a binary cross-entropy error for occupancy grids \red{and} a scaled sum-of-squared error for SDFs. The loss $\mathcal{L}_{\text{VAE}}$ is minimized using stochastic gradient descent (SGD) by approximating the expectation \red{using samples}:
\begin{align}
- \mathbb{E}_{q(z |y)}[\ln p(y|z)] \approx - \sum_{l = 1}^L \ln p(y | z^{(l)})
\end{align}
\red{The required samples $z^{(l)} \sim q(z|y)$ are computed using the so-called reparameterization trick,
\begin{align}
z^{(l)} = \mu(y) + \epsilon^{(l)} \sigma(y)\quad\text{with}\quad\epsilon^{(l)} \sim \mathcal{N}(\epsilon; 0, I_Q),\label{eq:repa}
\end{align}
in order to make $\mathcal{L}_{\text{VAE}}$, specifically the sampling process, differentiable.} \red{In practice, we found $L = 1$ samples to be sufficient -- which conforms with results by \cite{Kingma2014ICLR}. At test time, the sampling process $z\sim q(z|y)$ is replaced by the predicted mean $\mu(y)$.}
\red{Overall, the standard VAE allows us to embed the reference shapes in a low-dimensional latent space. In practice, however, the learned prior might still include unreasonable shapes.}

\boldparagraph{Denoising \VAE (\DVAE)}
\red{In order to avoid inappropriate shapes to be included in our shape prior, we consider a denoising variant of the \VAE allowing to obtain a tighter bound on the likelihood $p(y)$.} More specifically, a corruption process $y' \sim p(y' | y)$ is considered and the corresponding evidence lower bound results in the following loss:
\begin{align}
    \begin{split}
        \mathcal{L}_{\text{DVAE}}(w) = &- \mathbb{E}_{q(z | y')}[\ln p(y|z)]\\
        &+ \text{KL}(q(z | y')| p(z)).
    \end{split}
\end{align}
\red{Note that the reconstruction error $-\ln p(y|z)$ is still computed with respect to the uncorrupted shape $y$ while $z$, in contrast to \eqnref{eq:vae}, is sampled conditioned on the corrupted shape $y'$.} In practice, the corruption process $p(y' | y)$ is modeled using Bernoulli noise for occupancy grids and Gaussian noise for SDFs.
In experiments, we found \DVAEs to learn more robust latent spaces \red{-- meaning the prior is less likely to contain unreasonable shapes. In the following, we always use \DVAEs as shape priors.}

\subsection{Shape Inference}
\label{subsec:method-inference}

\begin{figure}[t]
	\vspace*{-\figskipabove px}
	\centering
	\hfill
	\begin{subfigure}[t]{0.25\linewidth}
		\vspace{0px}
		\centering
		\includegraphics[height=4.5cm]{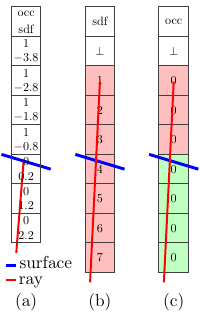}
	\end{subfigure}
	\begin{subfigure}[t]{0.6\linewidth}
		\vspace{3px}
		\centering
		\hspace*{-12px}
		\includegraphics[height=4.5cm]{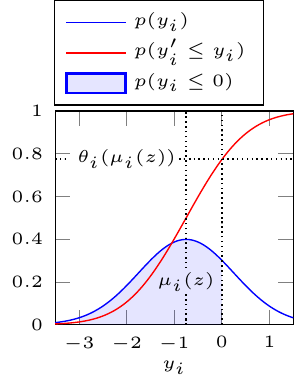}
	\end{subfigure}
	\hfill
	\vspace*{-8px}
	\caption{{{\bf Left: Problem with SDF Observations.} Illustration of a ray ({\color{red}red line}) correctly hitting a surface ({\color{blue}blue line}) causing the (signed) distance values and occupancy values computed for voxels along the ray to be correct (\cf (a)). A noisy ray, however, causes all voxels along the ray to be assigned incorrect distance values (marked {\colorbox{red!25}{red}}) \wrt to the true surface ({\color{blue}blue line}) because the ray ends far behind the actual surface (\cf (b)). When using occupancy only, in contrast, only the voxels behind the surface are assigned invalid occupancy states (marked {\colorbox{red!25}{red}}); the remaining voxels are labeled correctly (marked {\colorbox{green!25}{green}}; \cf (c)).
			{\bf Right: Proposed Gaussian-to-Bernoulli Transformation.} For $p(y_i) := p(y_i | z) = \mathcal{N}(y_i;\mu_i(z), \sigma^2)$ ({\color{blue}blue}), we illustrate the transformation discussed in \secref{subsec:method-inference} allowing to use the binary observations $x_i$ (for $x_i \neq \uk$) to supervise the SDF predictions. This is achieved by transforming the predicted Gaussian distribution to a Bernoulli distribution with occupancy probability $\theta_i(\mu_i(z)) = p(y_i \leq 0)$ ({\color{blue}blue area}).}}
	\label{fig:method-sdf}
	\vspace*{-\figskipbelow px}
\end{figure}

After learning the shape prior, \red{defining the joint distribution $p(y, z)$ of shapes $y$ and latent codes $z$ as product of generative model $p(y|z)$ and prior $p(z)$}, shape completion can be formulated as a maximum likelihood (\ML) problem \red{for $p(y, z)$} over the lower-dimensional latent space $\mathcal{Z} = \mathbb{R}^Q$. The corresponding negative log-likelihood $-\ln p(y, z)$ to be minimized can be written as
\begin{align}
\mathcal{L}_{\text{ML}}(z) &= - \sum_{x_i \neq \uk} \ln p(y_i = x_i | z) - \ln p(z).\label{eq:ml}
\end{align}
As the prior $p(z)$ is Gaussian, the negative log-probability $- \ln p(z)$ is proportional to $\|z\|_2^2$ and \red{constrains the problem to likely, \ie, reasonable, shapes with respect to the shape prior}. As before, the generative model $p(y | z)$ decomposes over voxels; here, we can only consider actually observed voxels $x_i \neq \uk$. \red{We assume that the learned shape prior can complete the remaining, unobserved voxels $x_i = \uk$.} Instead of solving \eqnref{eq:ml} for each observation $x \in \mathcal{X}$ independently, however, we follow the idea of amortized inference \citep{Gersham2014COGSCI} and train a \red{new} encoder $z(x;w)$ to \emph{learn} \ML. To this end, we keep the generative model $p(y|z)$ fixed and train \red{only} the weights $w$ of the \red{new} encoder $z(x;w)$ using the \ML objective as loss:
\begin{align}
    \begin{split}
        \mathcal{L}_{\text{dAML}}(w) =& - \sum _{x_i \neq \uk} \ln p(y_i = x_i | z(x; w))\\
        &- \lambda \ln p(z(x; w)).\label{eq:aml}
    \end{split}
\end{align}
\red{Here,} $\lambda$ controls the importance of the shape prior. The exact form of the probabilities $p(y_i = x_i | z)$ depends on the used shape representation. For occupancy grids, this term results in a cross-entropy error as \red{both the predicted voxels $y_i$ and the observations $x_i$} are, for $x_i \neq \uk$, binary. For SDFs, however, the term is not well-defined as $p(y_i | z)$ is modeled with a continuous Gaussian distribution, while the observations $x_i$ are binary. As solution, we could compute (signed) distance values along the rays corresponding to observed points (\eg, following \citep{Steinbrucker2013ICCV}) in order to obtain continuous observations $x_i \in\mathbb{R}$ for $x_i \neq \uk$. However, as illustrated in \figref{fig:method-sdf}, noisy observations cause the distance values along the whole ray to be invalid. This can partly be avoided when relying \red{only} on occupancy to represent the observations; in this case, free space (\cf \figref{fig:method-problem}) observations are partly correct even though observed points may lie within the corresponding shapes.

For making SDFs tractable (\ie, to predict sub-voxel accurate, visually smooth and appealing shapes, see \secref{sec:experiments}) \red{while using binary observations}, we propose to define $p(y_i = x_i | z)$ through a simple transformation. In particular, as $p(y_i | z)$ is modeled using a Gaussian distribution $\mathcal{N}(y_i ; \mu_i(z), \sigma^2)$ where $\mu_i(z)$ is predicted using the fixed decoder ($\sigma^2$ is constant), and $x_i$ is binary (for $x_i \neq \uk$), we introduce a mapping $\theta_i(\mu_i(z))$ transforming the predicted \red{mean SDF value to an occupancy probability} $\theta_i(\mu_i(z))$:
\begin{align}
p(y_i = x_i | z) = \text{Ber}(y_i = x_i; \theta_i(\mu_i(z)))
\end{align}
As, \red{by construction (see \secref{subsec:method-problem}), occupied voxels have negative sign or value zero in the SDF}, we can derive the occupancy probability $\theta_i(\mu_i(z))$ as the probability of a non-positive distance:
\begin{align}
\theta_i(\mu_i(z)) &= \mathcal{N}(y_i \leq 0; \mu_i(z), \sigma^2)\\[3px]
&= \frac{1}{2} \left(1 + \text{erf}\left(\frac{- \mu_i(z)}{\sigma \sqrt{\pi}}\right)\right).\label{eq:sdf}
\end{align}
Here, $\text{erf}$ is the error function which, in practice, can be approximated following~\citep{Abramowitz1974}. \eqnref{eq:sdf} is illustrated in \figref{fig:method-sdf} where the occupancy probability $\theta_i(\mu_i(z))$ is computed as the area under the Gaussian bell curve for $y_i \leq 0$. This per-voxel transformation can easily be implemented as non-linear layer and its derivative \wrt $\mu_i(z)$
is, by construction, a Gaussian. \red{Note that the transformation is correct, not approximate, based on our model assumptions and the definitions in \secref{subsec:method-problem}.} \red{Overall, this transformation allows us to easily minimize \eqnref{eq:aml} for both occupancy grids and SDFs using binary observations. The obtained encoder embeds the observations in the latent shape space to perform shape completion.}

\subsection{Practical Considerations}

\begin{figure*}[ht]
    \vspace*{-\figskipabove px}
    \vspace*{2px}
	\centering
        
	\begin{subfigure}[t]{0.32\textwidth}
        \centering\vspace{0px}
   		\begin{subfigure}[t]{0.32\textwidth}
   			\includegraphics[width=1.75cm,trim={\cropleft cm \croplower cm \cropright cm \cropupper cm},clip]{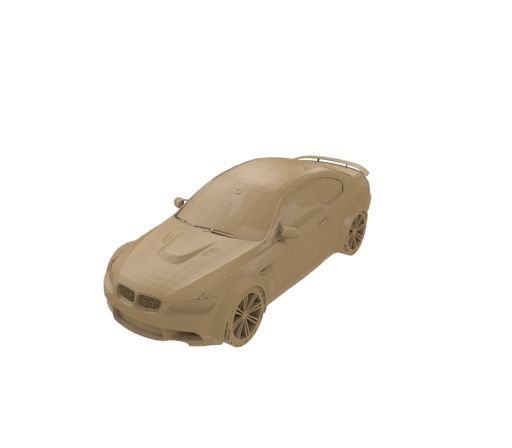}
   		\end{subfigure}
   		\begin{subfigure}[t]{0.32\textwidth}
   			\includegraphics[width=1.75cm,trim={\cropleft cm \croplower cm \cropright cm \cropupper cm},clip]{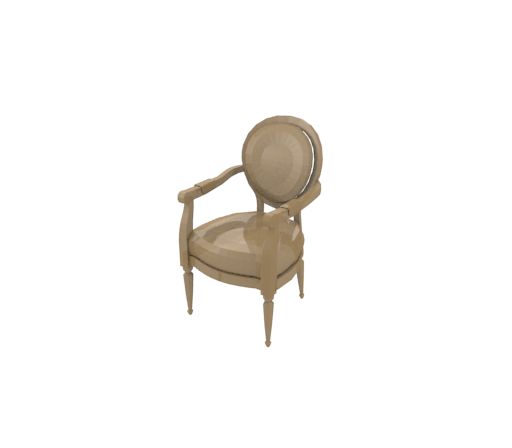}
   		\end{subfigure}
   		\begin{subfigure}[t]{0.32\textwidth}
   			\includegraphics[width=1.75cm,trim={\cropleft cm \croplower cm \cropright cm \cropupper cm},clip]{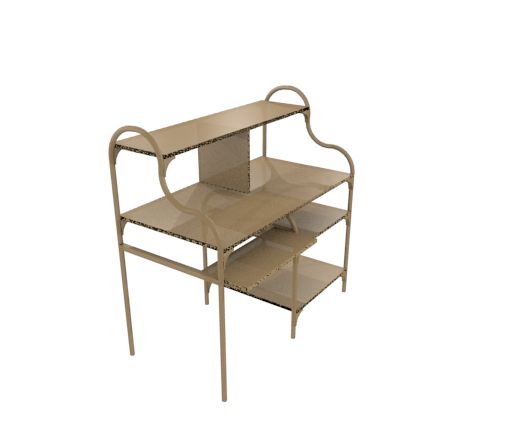}
   		\end{subfigure}
        \subcaption{Original}
		\label{fig:data-shapenet-modelnet-a}
	\end{subfigure}
	\begin{subfigure}[t]{0.32\textwidth}
        \centering\vspace{0px}
   		\begin{subfigure}[t]{0.32\textwidth}
   			\includegraphics[width=1.75cm,trim={\cropleft cm \croplower cm \cropright cm \cropupper cm},clip]{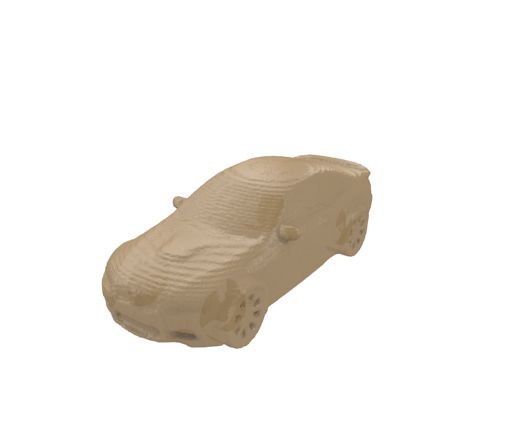}
   		\end{subfigure}
   		\begin{subfigure}[t]{0.32\textwidth}
   			\includegraphics[width=1.75cm,trim={\cropleft cm \croplower cm \cropright cm \cropupper cm},clip]{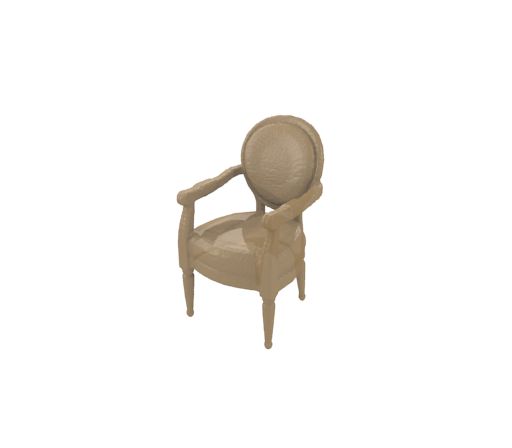}
   		\end{subfigure}
   		\begin{subfigure}[t]{0.32\textwidth}
   			\includegraphics[width=1.75cm,trim={\cropleft cm \croplower cm \cropright cm \cropupper cm},clip]{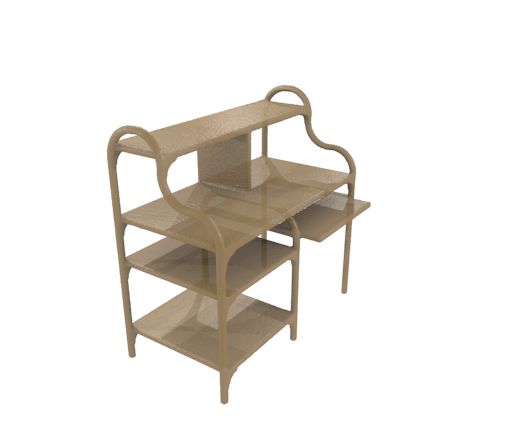}
   		\end{subfigure}
        \subcaption{TSDF Fusion, $256^3$}
		\label{fig:data-shapenet-modelnet-b}
	\end{subfigure}
	\begin{subfigure}[t]{0.32\textwidth}
		\centering\vspace{0px}
   		\begin{subfigure}[t]{0.32\textwidth}
   			\includegraphics[width=1.75cm,trim={\cropleft cm \croplower cm \cropright cm \cropupper cm},clip]{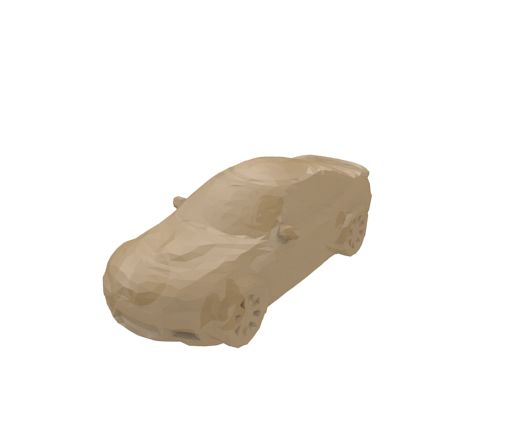}
   		\end{subfigure}
   		\begin{subfigure}[t]{0.32\textwidth}
   			\includegraphics[width=1.75cm,trim={\cropleft cm \croplower cm \cropright cm \cropupper cm},clip]{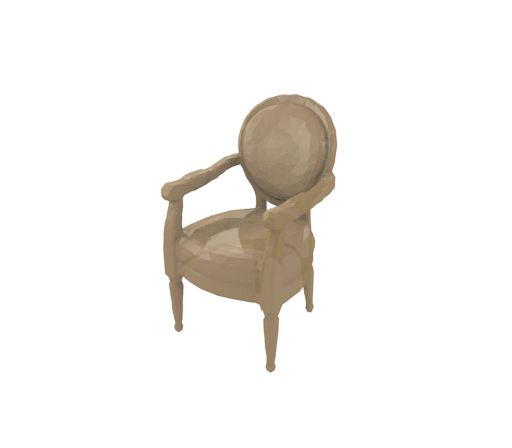}
   		\end{subfigure}
   		\begin{subfigure}[t]{0.32\textwidth}
   			\includegraphics[width=1.75cm,trim={\cropleft cm \croplower cm \cropright cm \cropupper cm},clip]{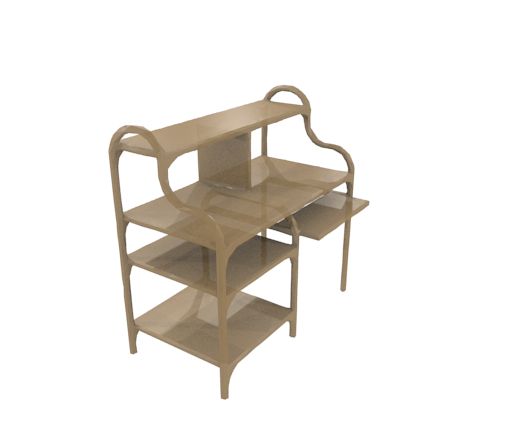}
   		\end{subfigure}
        \subcaption{Simplification, $5\text{k}$ Faces}
		\label{fig:data-shapenet-modelnet-c}
	\end{subfigure}
	\\
	\begin{subfigure}[t]{0.32\textwidth}
        \centering\vspace{0px}
   		\begin{subfigure}[t]{0.32\textwidth}
   			\includegraphics[width=1.75cm,trim={\cropleft cm \croplower cm \cropright cm \cropupper cm},clip]{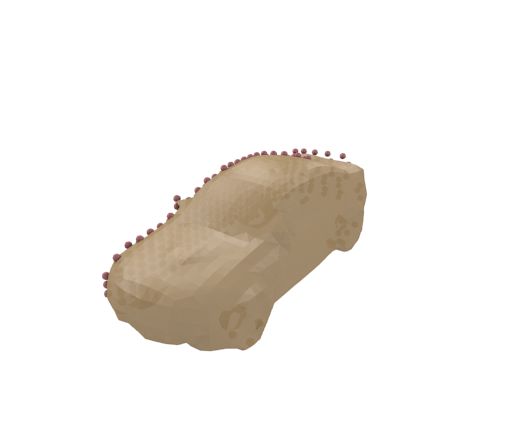}
   		\end{subfigure}
   		\begin{subfigure}[t]{0.32\textwidth}
   			\includegraphics[width=1.75cm,trim={\cropleft cm \croplower cm \cropright cm \cropupper cm},clip]{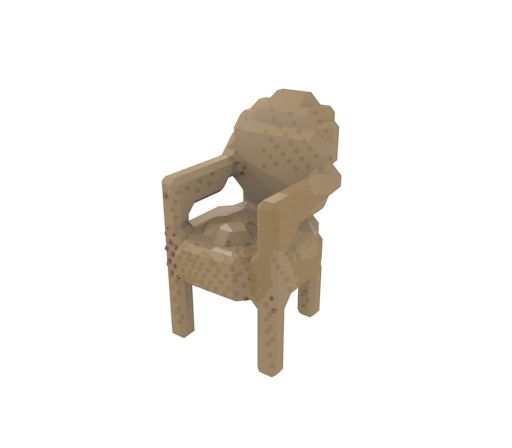}
   		\end{subfigure}
   		\begin{subfigure}[t]{0.32\textwidth}
   			\includegraphics[width=1.75cm,trim={\cropleft cm \croplower cm \cropright cm \cropupper cm},clip]{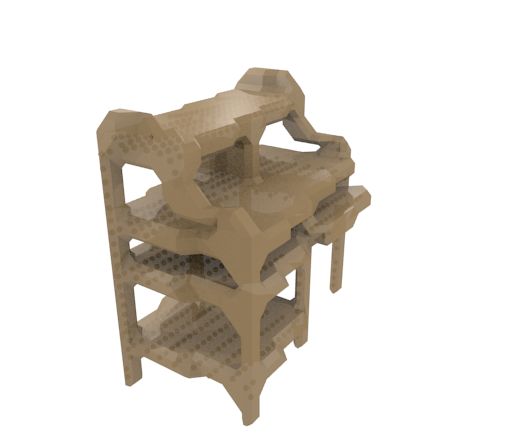}
   		\end{subfigure}
        \subcaption{Reconstruction, $24\ntimes54\ntimes24$/$32^3$}
		\label{fig:data-shapenet-modelnet-d}
	\end{subfigure}
	\begin{subfigure}[t]{0.32\textwidth}
        \centering\vspace{0px}
   		\begin{subfigure}[t]{0.32\textwidth}
   			\includegraphics[width=1.75cm,trim={\cropleft cm \croplower cm \cropright cm \cropupper cm},clip]{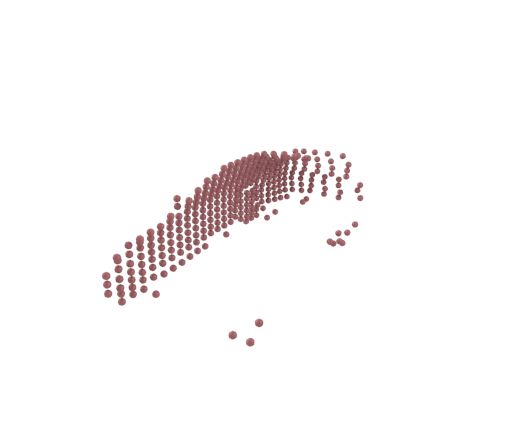}
   		\end{subfigure}
   		\begin{subfigure}[t]{0.32\textwidth}
   			\includegraphics[width=1.75cm,trim={\cropleft cm \croplower cm \cropright cm \cropupper cm},clip]{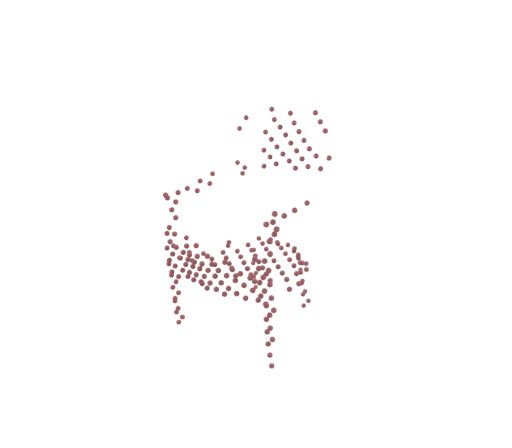}
   		\end{subfigure}
   		\begin{subfigure}[t]{0.32\textwidth}
   			\includegraphics[width=1.75cm,trim={\cropleft cm \croplower cm \cropright cm \cropupper cm},clip]{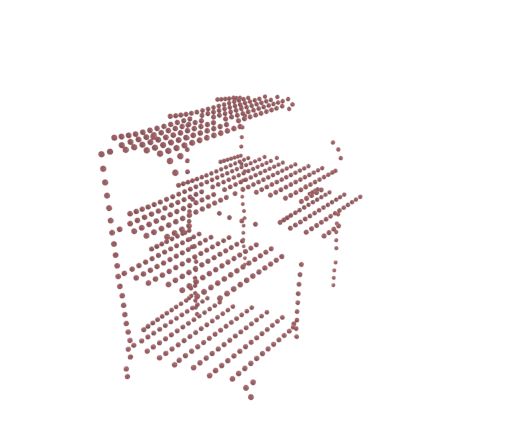}
   		\end{subfigure}
        \subcaption{Observations}
		\label{fig:data-shapenet-modelnet-e}
	\end{subfigure}
	\begin{subfigure}[t]{0.32\textwidth}
        \centering\vspace{0px}
   		\begin{subfigure}[t]{0.32\textwidth}
   			\includegraphics[width=1.75cm,trim={\cropleft cm \croplower cm \cropright cm \cropupper cm},clip]{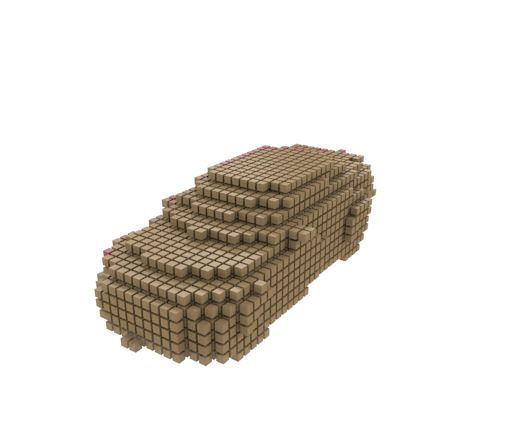}
   		\end{subfigure}
   		\begin{subfigure}[t]{0.32\textwidth}
   			\includegraphics[width=1.75cm,trim={\cropleft cm \croplower cm \cropright cm \cropupper cm},clip]{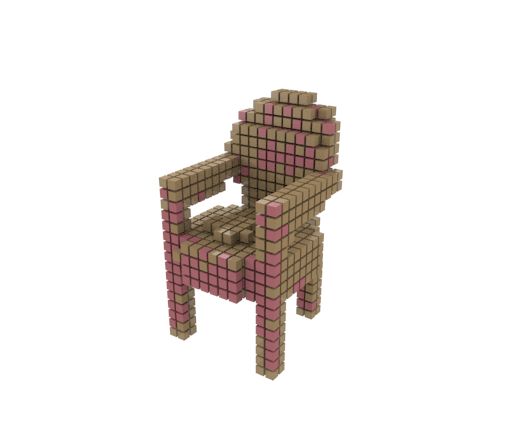}
   		\end{subfigure}
   		\begin{subfigure}[t]{0.32\textwidth}
   			\includegraphics[width=1.75cm,trim={\cropleft cm \croplower cm \cropright cm \cropupper cm},clip]{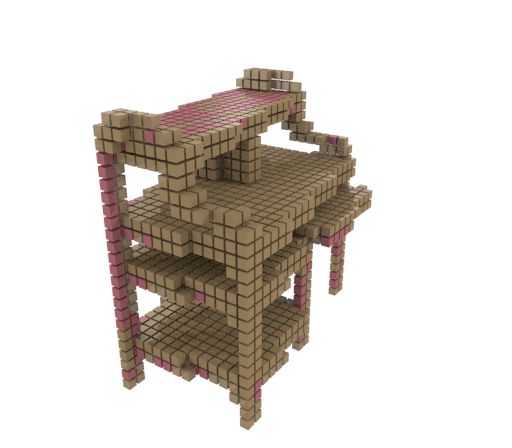}
   		\end{subfigure}
        \subcaption{Voxelization, $24\ntimes54\ntimes24$/$32^3$}
		\label{fig:data-shapenet-modelnet-f}
	\end{subfigure}
    \vspace*{-\figskipcaption px}
	\caption{{\bf ShapeNet and ModelNet Data Generation Pipeline.} On ShapeNet and ModelNet we illustrate: {\bf (\subref{fig:data-shapenet-modelnet-a})} samples from the original datasets; {\bf (\subref{fig:data-shapenet-modelnet-b})} fused watertight meshes from TSDF fusion at $256^3$ voxels resolution using \citep{Riegler2017THREEDV}; {\bf (\subref{fig:data-shapenet-modelnet-c})} simplified meshes ($5k$ faces); {\bf (\subref{fig:data-shapenet-modelnet-d})} marching cubes \citep{Lorensen1987SIGGRAPH} reconstructions from the SDFs computed from (\subref{fig:data-shapenet-modelnet-c}) (resolutions $24 \ntimes 54 \ntimes 24$ and $32^3$ voxels; note that steps (\subref{fig:data-shapenet-modelnet-b}) and (\subref{fig:data-shapenet-modelnet-c}) are necessary to derive exact SDFs); {\bf (\subref{fig:data-shapenet-modelnet-e})} observations obtained by projection into a single view; and {\bf (\subref{fig:data-shapenet-modelnet-f})} voxelized observations and shapes. Shapes (meshes and occupancy grids) {in \color{rbeige}beige} and observations in {\color{rred}red}.}
	\label{fig:data-shapenet-modelnet}
    \vspace*{-\figskipbelow px}
\end{figure*}

\boldparagraph{Encouraging Variety}
So far, our \AML formulation assumes a deterministic encoder $z(x,w)$ which predicts, given the observation $x$, a single code $z$ corresponding to a completed shape. A closer look at \eqnref{eq:aml}, however, reveals an unwanted problem: the data term scales with the number of observations, \ie, $|\{x_i \neq \uk\}|$, while the regularization term stays constant -- with less observations, the regularizer gains in importance leading to limited variety in the predicted shapes because $z(x; w)$ tends towards zero.

In order to encourage variety, we draw inspiration from the \VAE shape prior. Specifically, we use a probabilistic recognition model
\begin{align}
q(z|x) = \mathcal{N}(z; \mu(x), \text{diag}(\sigma^2(x)))
\end{align}
(\cf see \eqnref{eq:encoder-decoder}) and replace the negative log-likelihood $-\ln p(z)$ with the corresponding Kullback-Leibler divergence $\text{KL}(q(z|x)|p(z))$ with $p(z) = \mathcal{N}(z; 0, I_Q)$. Intuitively, this makes sure that the encoder's predictions ``cover'' the prior distribution -- thereby enforcing variety. Mathematically, the resulting loss, \ie, 
\begin{align}
\begin{split}
    \mathcal{L}_{\text{AML}}(w) =& - \red{\mathbb{E}_{q(z|x)}\left[\sum_{x_i \neq \uk} \ln p(y_i = x_i | z)\right]}\\
    &+ \lambda \text{KL}(q(z|x)p(z)),\label{eq:daml}
\end{split}
\end{align}
can be interpreted as the result of maximizing the evidence lower bound of a model with observation process $p(x | y)$ (analogously to the corruption process $p(y'|y)$ for \DVAEs in \citep{Im2017AAAI} and \secref{subsec:method-prior}). \red{The expectation is approximated using samples (following the reparameterization trick in \eqnref{eq:repa}) and, during testing, the sampling process $z \sim q(z|x)$ is replaced by the mean prediction $\mu(x)$.} In practice, we find that \eqnref{eq:daml} improves visual quality of the completed shapes. We compare this \AML model to its deterministic variant \dAML in \secref{sec:experiments}.

\boldparagraph{Handling Noise}
Another problem of our \AML formulation concerns noise. On KITTI, for example, specular or transparent surfaces cause invalid observations -- laser rays traversing through these surfaces cause observations to lie within shapes or not get reflected. However, our \AML framework assumes deterministic, \ie, trustworthy, observations \red{-- as can be seen in the reconstruction error in \eqnref{eq:daml}.} Therefore, we introduce per-voxel weights $\kappa_i$ computed using the reference shapes $\mathcal{Y} = \{y_m\}_{m=1}^M$:
\begin{align}
\kappa_i = 1 - \left(\frac{1}{M} \sum_{m = 1}^M y_{m,i}\right) \in [0,1]
\end{align}
where $y_{m,i} = 1$ if and only if the corresponding voxel is occupied. Applied to observations $x_i = 0$, these are trusted less if they are unlikely under the shape prior. Note that for point observations, \ie, $x_i = 1$, this is not necessary as we explicitly consider ``filled'' shapes (see \secref{sec:data}). This can also be interpreted as imposing an additional \red{mean shape prior on the predicted shapes with respect to the observed free space}. In addition, we use a corruption process $p(x' | x)$ consisting of Bernoulli and Gaussian noise during training (analogously to the \DVAE shape prior).

\section{Experiments}

\subsection{Data}
\label{sec:data}

\begin{figure}[t]
    \vspace*{-\figskipabove px}
    \vspace*{4px}
    \centering
    \begin{subfigure}[t]{0.235\textwidth}
        \vspace{0px}\centering
        \begin{subfigure}[t]{0.48\textwidth}
            \includegraphics[width=1.75cm,trim={\cropleft cm \croplower cm \cropright cm \cropupper cm},clip]{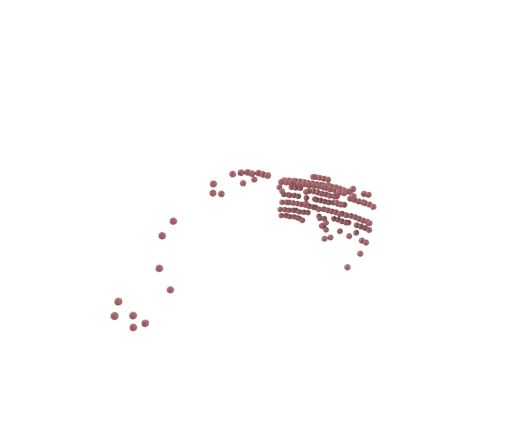}
        \end{subfigure}
        \begin{subfigure}[t]{0.48\textwidth}
            \includegraphics[width=1.75cm,trim={\cropleft cm \croplower cm \cropright cm \cropupper cm},clip]{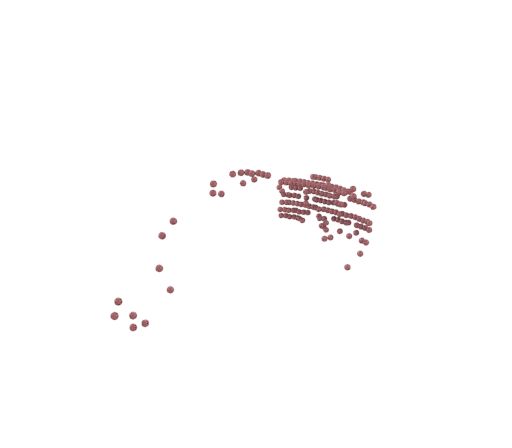}
        \end{subfigure}
        \\
        \begin{subfigure}[t]{0.48\textwidth}
            \includegraphics[width=1.75cm,trim={\cropleft cm \croplower cm \cropright cm \cropupper cm},clip]{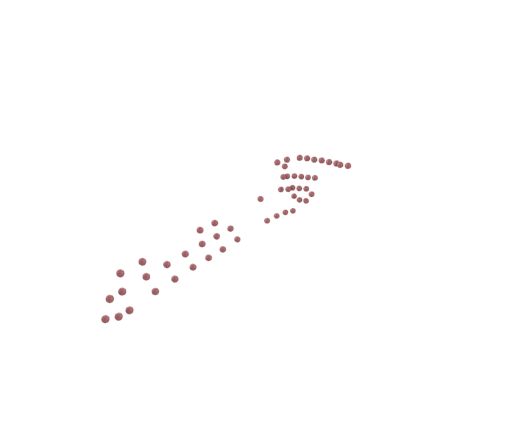}
        \end{subfigure}
        \begin{subfigure}[t]{0.48\textwidth}
            \includegraphics[width=1.75cm,trim={\cropleft cm \croplower cm \cropright cm \cropupper cm},clip]{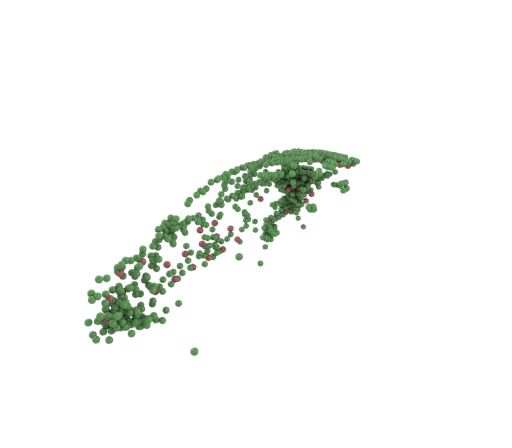}
        \end{subfigure}
        \subcaption{KITTI, Point Clouds}
    \end{subfigure}
    \begin{subfigure}[t]{0.235\textwidth}
        \vspace{0px}\centering
        \begin{subfigure}[t]{0.48\textwidth}
            \includegraphics[width=1.75cm,trim={\cropleft cm \croplower cm \cropright cm \cropupper cm},clip]{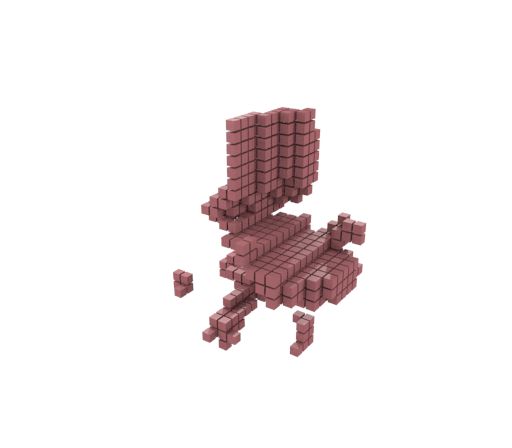}
        \end{subfigure}
        \begin{subfigure}[t]{0.48\textwidth}
            \includegraphics[width=1.75cm,trim={\cropleft cm \croplower cm \cropright cm \cropupper cm},clip]{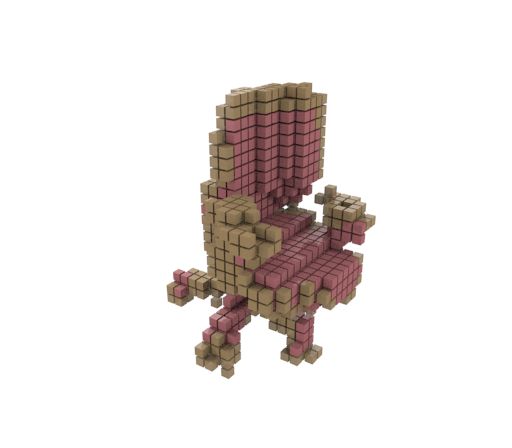}
        \end{subfigure}
        \\
        \begin{subfigure}[t]{0.48\textwidth}
            \includegraphics[width=1.75cm,trim={\cropleft cm \croplower cm \cropright cm \cropupper cm},clip]{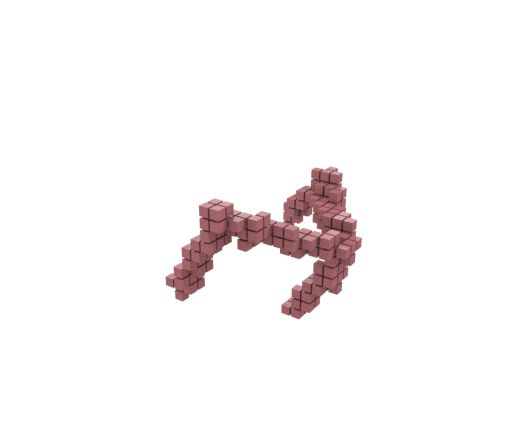}
        \end{subfigure}
        \begin{subfigure}[t]{0.48\textwidth}
            \includegraphics[width=1.75cm,trim={\cropleft cm \croplower cm \cropright cm \cropupper cm},clip]{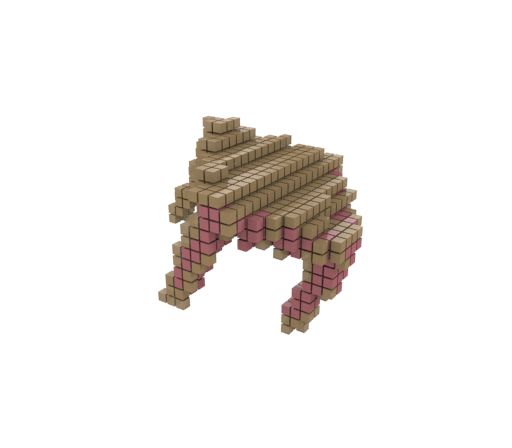}
        \end{subfigure}
        \subcaption{\Kinect, Occupancy Grids}
    \end{subfigure}
    \vspace*{-\figskipcaption px}
	\caption{{{\bf Extracted KITTI and \Kinect Data.} For KITTI, we show observed points in {\color{rred}red} and the accumulated, partial ground truth in {\color{rgreen}green}. Note that for the first example ground truth is not available due to missing past/future observations. For \Kinect, we show observations in {\color{rred}red} and ElasticFusion \citep{Whelan2015RSS} ground truth in {\color{rbeige}beige}. Note that the objects are rotated and not aligned as in ModelNet (\cf \figref{fig:data-shapenet-modelnet}).}}
	\label{fig:data-kitti-yang}
    \vspace*{-\figskipbelow px}
\end{figure}

We briefly introduce our synthetic shape completion benchmarks, derived from ShapeNet \citep{Chang2015ARXIV} and ModelNet \citep{Wu2015CVPR} (\cf \figref{fig:data-shapenet-modelnet}), and our data preparation for KITTI \citep{Geiger2012CVPR} and \Kinect \citep{Yang2018ARXIVb} (\cf \figref{fig:data-kitti-yang}); \tabref{tab:data} summarizes key statistics \red{including the level of supervision computed as the fraction of observed voxels, \ie $\nicefrac{|\{x_{n,i} \neq \uk\}|}{HWD}$, averaged over observations $x_n$.}

\boldparagraph{ShapeNet}
We utilize the truncated SDF (TSDF) fusion approach of \cite{Riegler2017THREEDV} to obtain watertight versions of the provided car shapes allowing to reliably and efficiently compute occupancy grids and SDFs. Specifically, we use $100$ depth maps of $640 \ntimes 640$ pixels resolution, distributed uniformly on the sphere around the shape, and perform TSDF fusion at a resolution of $256^3$ voxels. Detailed watertight meshes, without inner structures, can then be extracted using marching cubes \citep{Lorensen1987SIGGRAPH} and simplified to $5\text{k}$ faces using MeshLab's quadratic simplification algorithm \citep{Cignoni2008EICC}, see \figref{fig:data-shapenet-modelnet-a} to \subref{fig:data-shapenet-modelnet-c}. Finally, we manually selected $220$ shapes from this collection, removing exotic cars, unwanted configurations, or shapes with large holes (\eg, missing floors or open windows).

The shapes are splitted into $|\mathcal{Y}| = 100$ reference shapes, $|\mathcal{Y}^*| = 100$ shapes for training the inference model, and $20$ test shapes. We randomly perturb rotation and scaling to obtain $5$ variants of each shape, voxelize them using triangle-voxel intersections and subsequently ``fill'' the obtained volumes using a connected components algorithm \citep{Jones2001}. For computing SDFs we use SDFGen\footnote{\url{https://github.com/christopherbatty/SDFGen}.}. We use three different resolutions: $H \ntimes W \ntimes D = 24 \ntimes 54 \ntimes 24$, $32 \ntimes 72 \ntimes 32$ and $48 \ntimes 108 \ntimes 48$ voxels. Examples are shown in \figref{fig:data-shapenet-modelnet-d} to \subref{fig:data-shapenet-modelnet-f}.

Finally, we use the OpenGL renderer of \cite{Guney2015CVPR} to obtain $10$ depth maps per shape. The incomplete observations $\mathcal{X}$ are obtained by re-projecting them into 3D and marking voxels with at least one point as occupied and voxels between occupied voxels and the camera center as free space. We obtain more dense point clouds at $48 \ntimes 64$ pixels resolution and sparser point clouds using depth maps of $24 \ntimes 32$ pixels resolution. For the latter, more challenging case we also add exponentially distributed noise (with rate parameter $70$) to the depth values, or randomly (with probability $0.075$) set them to the maximum depth to simulate the deficiencies of point clouds captured with real sensors, \eg, on KITTI. These two variants are denoted {\bf\clean} and {\bf\noisy}.
The obtained observations are illustrated in \figref{fig:data-shapenet-modelnet-e}.

\boldparagraph{KITTI}
We extract observations from KITTI's Velodyne point clouds using the provided ground truth 3D bounding boxes to avoid the inaccuracies of 3D object detectors (train/test split by \cite{Chen2016ARXIV}). As the 3D bounding boxes in KITTI fit very tightly, we first padded them by factor $0.25$ on all sides; afterwards, the observed points are voxelized into voxel grids of size $H \ntimes W \ntimes D = 24 \ntimes 54 \ntimes 24$, $32 \ntimes 72 \ntimes 32$ and $48 \ntimes 108 \ntimes 48$ voxels. To avoid taking points from the street, nearby walls, vegetation or other objects into account, we only consider those points lying within the original (\ie, not padded) bounding box. Finally, free space is computed using ray tracing as described above. We filter all observations to ensure that each observation contains a minimum of $50$ observations. For the bounding boxes in the test set, we additionally generated partial ground truth by accumulating the 3D point clouds of $10$ future and $10$ past frames around each observation. Examples are shown in \figref{fig:data-kitti-yang}.

\boldparagraph{ModelNet}
We use ModelNet10, comprising $10$ popular object categories (bathtub, bed, chair, desk, dresser, monitor, night stand, table, toilet) and select, for each category, the first $200$ and $20$ shapes from the provided training and test sets. Then, we follow the pipeline outlined in \figref{fig:data-shapenet-modelnet}, as on ShapeNet, using $10$ random variants per shape. Due to thin structures, however, SDF computation does not work well (especially for low resolution, \eg, $32^3$ voxels). Therefore, we approximate the SDFs using a 3D distance transform on the occupancy grids. Our experiments are conducted at a resolution of $H \ntimes W \ntimes D = 32^3$, $48^3$ and $64^3$ voxels. Given the increased difficulty, we use a resolution of $64^2$, $96^2$ and $128^2$ pixels for the observation generating depth maps. In our experiments, we consider bathtubs, chairs, desks and tables individually, as well as all $10$ categories together (resulting in $100\text{k}$ views overall). For \Kinect, we additionally used a dataset of rotated chairs and tables aligned with \Kinect's ground plane.

\begin{table}[t]
    \vspace*{-\figskipabove px}
    \centering
    {\small
        \begin{tabularx}{0.49\textwidth}{|@{ }X@{ }|@{ }c@{ }|@{ }c@{ }|@{ }c@{ }|@{ }c@{ }|}
            \hline
            & \multicolumn{2}{c@{ }|@{ }}{\bf Synthetic} & \multicolumn{2}{@{ }c@{ }|}{\bf Real}\\
            \hline
            & \clean/-noisy & ModelNet & KITTI & \Kinect\\
            \hline\hline
            \multicolumn{5}{|@{ }c@{ }|}{Training/Test Sets}\\[-2px]
            \multicolumn{5}{|@{ }c@{ }|}{\scriptsize\#Shapes for Shape Prior, \#Views for Shape Inference}\\
            \hline
            \#Shapes & 500/100 & 1000/200& \color{gray}-- & \color{gray}-- \\
            \#Views & 5000/1000 & 10000/2000& 8442/9194 & 30/10\\
            \hline
            \hline
            \multicolumn{5}{|@{ }c@{ }|}{Observed Voxels in \% ({\bf\color{rred}$<5\%$}) \& Resolutions}\\[-2px]
            \multicolumn{5}{|@{ }c@{ }|}{\tiny Low = $24\ntimes54\ntimes24$/$32^3$; Medium = $32\ntimes72\ntimes32$/$48^3$; High = $48\ntimes108\ntimes48$/$64^3$}\\
            \hline
            Low & 7.66/3.86 & 9.71& 6.79 & \bf\color{rred}0.87\\
            Medium & 6.1/\bf\color{rred}2.13 & 8.74& 5.24 & \color{gray}--\\
            High & \bf\color{rred}2.78/\bf\color{rred}0.93 & 8.28& \bf\color{rred}3.44 & \color{gray}--\\
            \hline
        \end{tabularx}
    }
    \vspace*{-\figskipcaption px}
	\caption{{\bf Dataset Statistics.} We report the number of (rotated and scaled) meshes, used as reference shapes, and the resulting number of observations (\ie, views, $10$ per shape). We also report the average fraction of observed voxels, \red{\ie, $\nicefrac{|\{x_i \neq \uk\}|}{HWD}$}. For ModelNet, we exemplarily report statistics for chairs; and for \Kinect, we report statistics for tables.}
	\label{tab:data}
    \vspace*{-\figskipbelow px}
\end{table}

\boldparagraph{\Kinect}
Yang \etal provide Kinect scans of various chairs and tables. They provide both single-view observations as well as ground truth from ElasticFusion \citep{Whelan2015RSS} as occupancy grids. However, the ground truth is not fully accurate, and only $40$ views are provided per object category. Still, the objects have been segmented to remove clutter and are appropriate for experiments in conjunction with ModelNet10. Unfortunately, Yang \etal do not provide SDFs; again, we use 3D distance transforms as approximation. Additionally, the observations do not indicate free space and we were required to guess an appropriate ground plane. For our experiments, we use $30$ views for training and $10$ views for testing, see \figref{fig:data-kitti-yang} for examples.
\subsection{Evaluation}
\label{sec:evaluation}

\begin{figure}
	\vspace*{-\figskipabove px}
	\centering
	\hspace*{-4px}
	\includegraphics[width=1.025\linewidth]{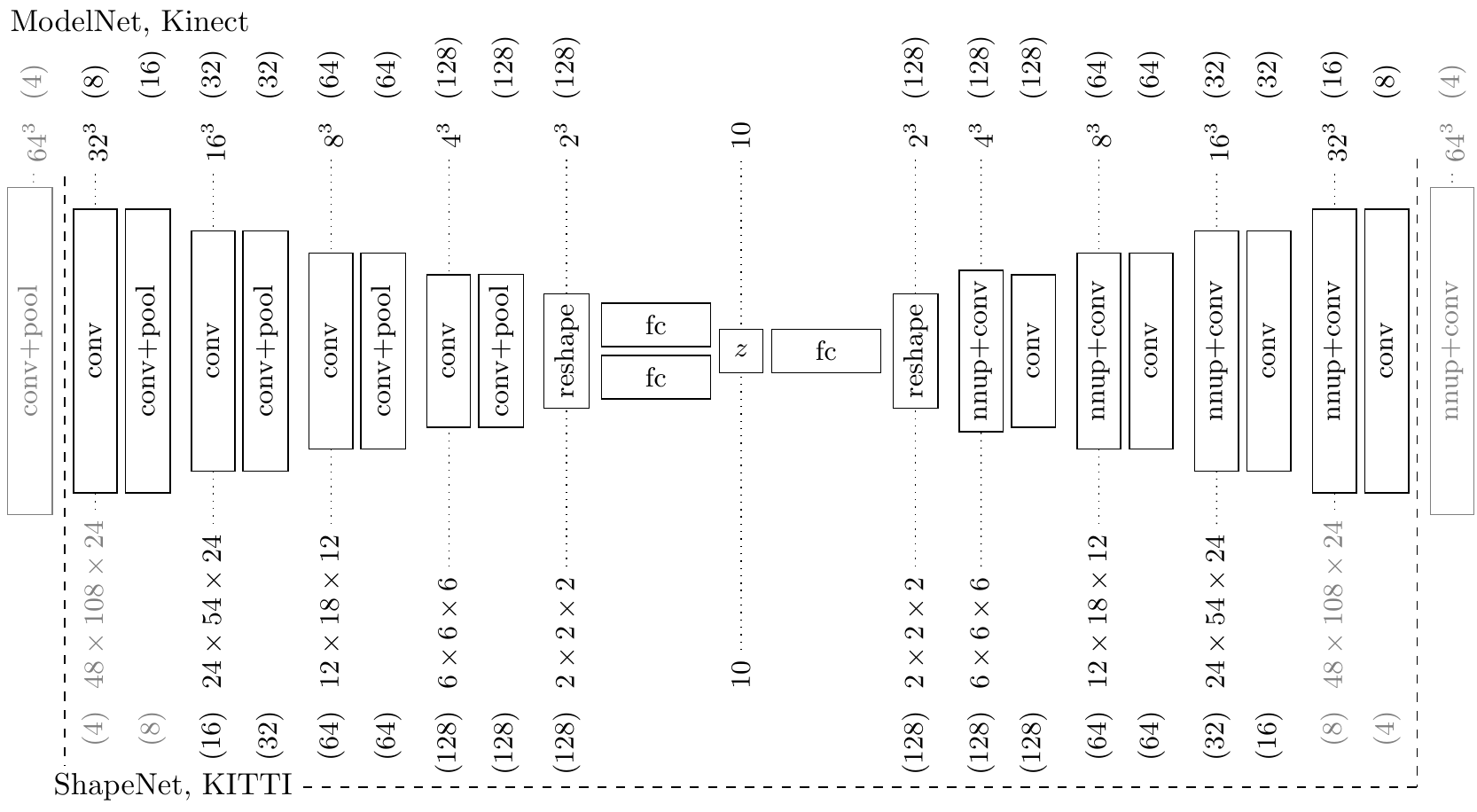}
	\vspace*{-12px}
	\caption{{\bf Network Architectures.} We use different resolutions for ShapeNet and KITTI as well as ModelNet and \Kinect (bottom and top, respectively). In both cases, architectures for higher resolutions employ one additional stage in the en- and decoder (in {\color{gray}gray}). Each convolutional layer is followed by $\text{ReLU}$ activations and batch normalization \citep{Ioffe2015ICML}; the window sizes for max pooling and nearest-neighbor upsampling can be derived from the context; the number of channels are given in parentheses.}
	\label{fig:architectures}
	\vspace*{-\figskipbelow px}
\end{figure}

For occupancy grids, we use Hamming distance (\Abs) and intersection-over-union (\IoU) between the (thresholded) predictions and the ground truth; note that lower \Abs is better, while lower \IoU is worse. For SDFs, we consider a mesh-to-mesh distance on ShapeNet and a mesh-to-point distance on KITTI. We follow \citep{Jensen2014CVPR} and consider accuracy (\Acc) and completeness (\Compl). To measure \Acc, we uniformly sample roughly $10\text{k}$  points on the reconstructed mesh and average their distance to the target mesh. Analogously, \Compl is the distance from the target mesh (or the ground truth points on KITTI) to the reconstructed mesh. Note that for both \Acc and \Compl, lower is better. On ShapeNet and ModelNet, we report both \Acc and \Compl in voxels, \ie, in multiples of the voxel edge length (\ie, in [vx], as we do not know the absolute scale of the models); on KITTI, we report \Compl in meters (\ie, in [m]).
\subsection{Architectures and Training}
\label{sec:training}

\begin{figure}[tp]
    \vspace*{-\figskipabove px}
    \vspace{4px}
    \centering
    {\scriptsize
        
    \begin{subfigure}[t]{0.5\textwidth}
        \vspace{0px}\centering
	    \begin{subfigure}[t]{0.15\textwidth}
   	    	\vspace{0px}\centering
   	    	GT,\\High\\
	        \includegraphics[width=1.5cm,trim={\cropleft cm \croplower cm \cropright cm \cropupper cm},clip]{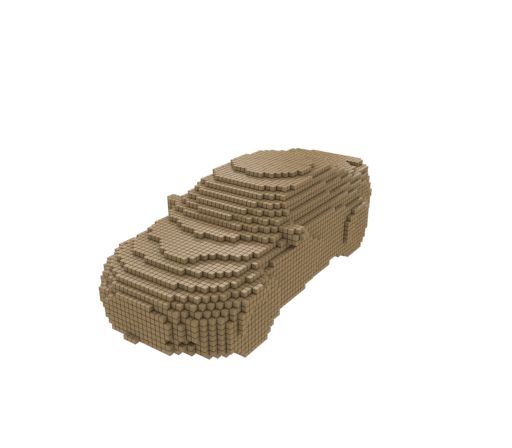}
	    \end{subfigure}
	    \begin{subfigure}[t]{0.15\textwidth}
   	    	\vspace{0px}\centering
   	    	GT\\
   	    	~\\
	        \includegraphics[width=1.5cm,trim={\cropleft cm \croplower cm \cropright cm \cropupper cm},clip]{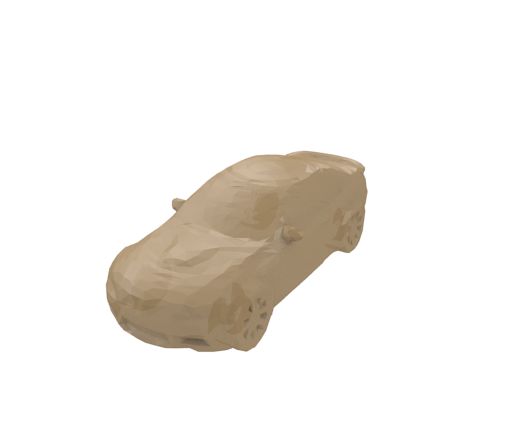}
	    \end{subfigure}
	    \begin{subfigure}[t]{0.15\textwidth}
   	    	\vspace{0px}\centering
   	    	\DVAE, Low\\
	        \includegraphics[width=1.5cm,trim={\cropleft cm \croplower cm \cropright cm \cropupper cm},clip]{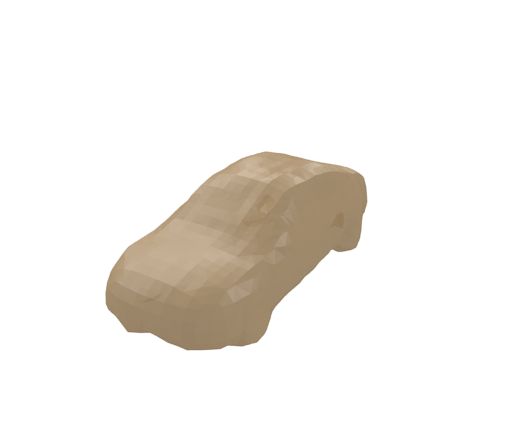}
	    \end{subfigure}
	    \begin{subfigure}[t]{0.15\textwidth}
   	    	\vspace{0px}\centering
   	    	\DVAE, High\\
   	    	\includegraphics[width=1.5cm,trim={\cropleft cm \croplower cm \cropright cm \cropupper cm},clip]{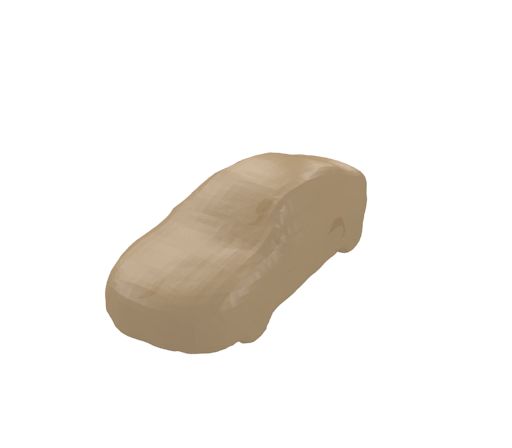}
	    \end{subfigure}
	    \begin{subfigure}[t]{0.15\textwidth}
   	    	\vspace{0px}\centering
   	    	\DVAE, Low\\
	        \includegraphics[width=1.5cm,trim={\cropleft cm \croplower cm \cropright cm \cropupper cm},clip]{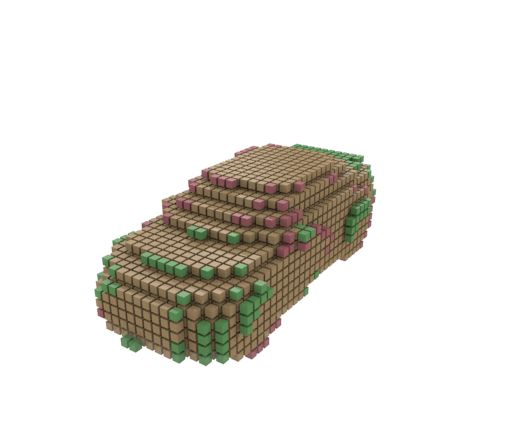}
	    \end{subfigure}
	    \begin{subfigure}[t]{0.15\textwidth}
   	    	\vspace{0px}\centering
   	    	\DVAE, High\\
	        \includegraphics[width=1.5cm,trim={\cropleft cm \croplower cm \cropright cm \cropupper cm},clip]{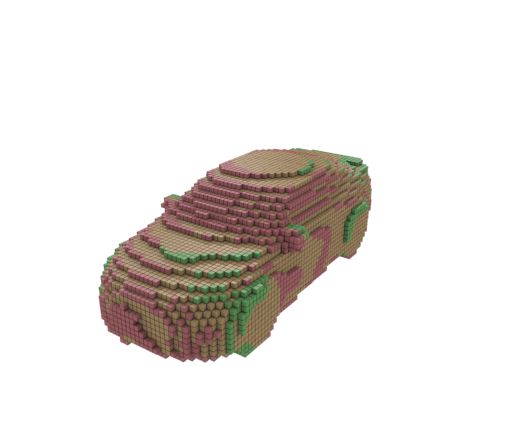}
	    \end{subfigure}
	    \\
	    \begin{subfigure}[t]{0.15\textwidth}
   	    	\vspace{0px}\centering
	        \includegraphics[width=1.5cm,trim={\cropleft cm \croplower cm \cropright cm \cropupper cm},clip]{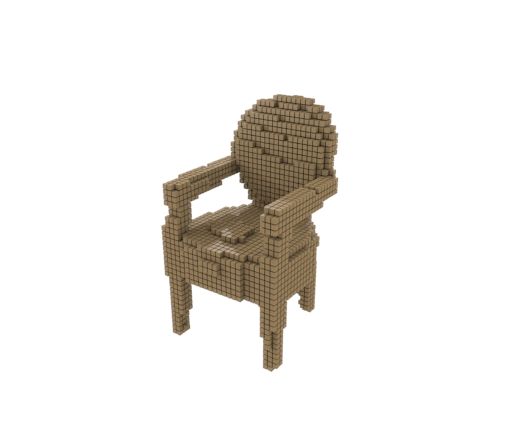}
	    \end{subfigure}
	    \begin{subfigure}[t]{0.15\textwidth}
   	    	\vspace{0px}\centering
	        \includegraphics[width=1.5cm,trim={\cropleft cm \croplower cm \cropright cm \cropupper cm},clip]{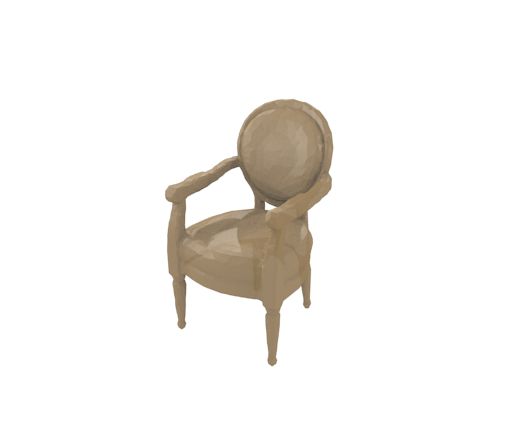}
	    \end{subfigure}
	    \begin{subfigure}[t]{0.15\textwidth}
   	    	\vspace{0px}\centering
	        \includegraphics[width=1.5cm,trim={\cropleft cm \croplower cm \cropright cm \cropupper cm},clip]{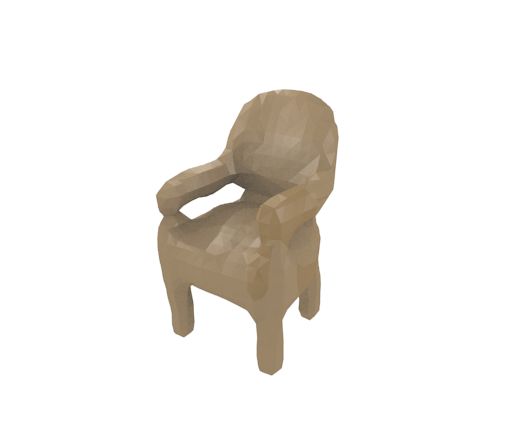}
	    \end{subfigure}
	    \begin{subfigure}[t]{0.15\textwidth}
   	    	\vspace{0px}\centering
   	    	\includegraphics[width=1.5cm,trim={\cropleft cm \croplower cm \cropright cm \cropupper cm},clip]{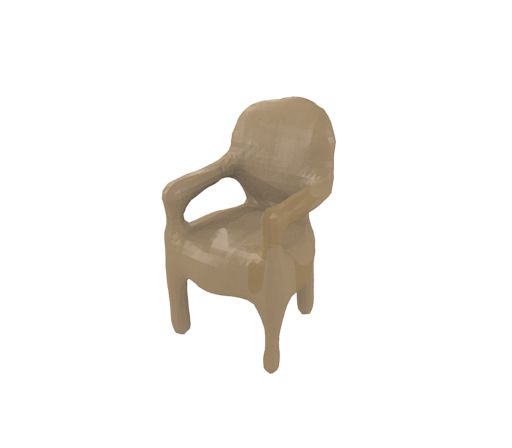}
	    \end{subfigure}
	    \begin{subfigure}[t]{0.15\textwidth}
   	    	\vspace{0px}\centering
	        \includegraphics[width=1.5cm,trim={\cropleft cm \croplower cm \cropright cm \cropupper cm},clip]{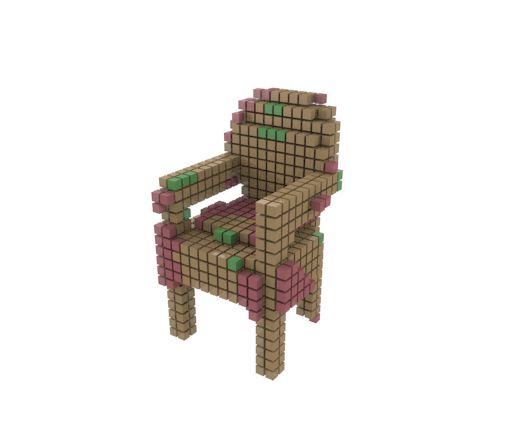}
	    \end{subfigure}
	    \begin{subfigure}[t]{0.15\textwidth}
   	    	\vspace{0px}\centering
	        \includegraphics[width=1.5cm,trim={\cropleft cm \croplower cm \cropright cm \cropupper cm},clip]{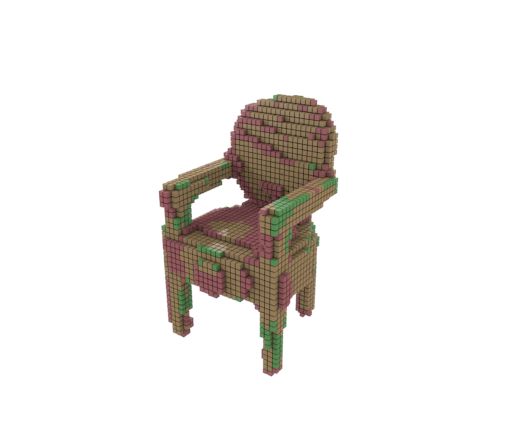}
	    \end{subfigure}
        \subcaption{Reconstructions, Low and High Resolution (\cf \tabref{tab:data})}
    \end{subfigure}
    \\[4px]
    \begin{subfigure}[t]{0.5\textwidth}
        \vspace{0px}\centering
	    \begin{subfigure}[t]{0.15\textwidth}
   	    	\vspace{0px}\centering
            Low\\
	        \includegraphics[width=1.5cm,trim={\cropleft cm \croplower cm \cropright cm \cropupper cm},clip]{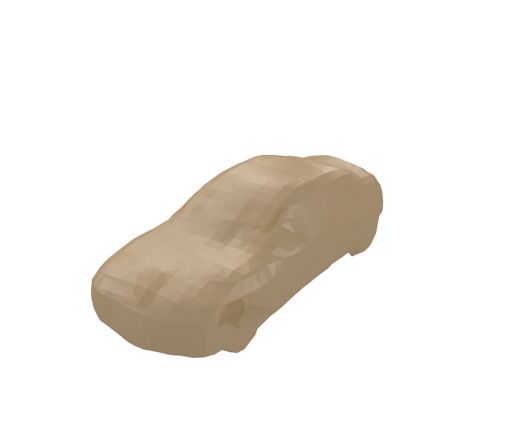}
	    \end{subfigure}
	    \begin{subfigure}[t]{0.15\textwidth}
   	    	\vspace{0px}\centering
            Low\\
	        \includegraphics[width=1.5cm,trim={\cropleft cm \croplower cm \cropright cm \cropupper cm},clip]{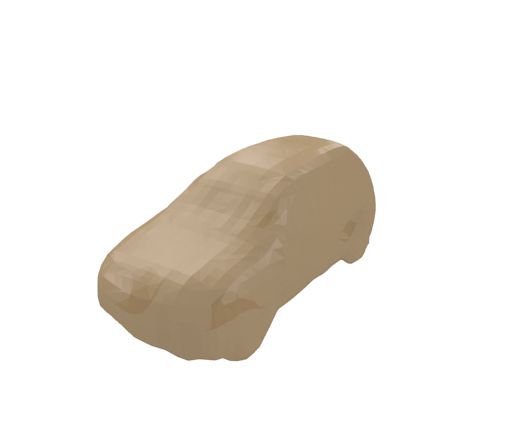}
	    \end{subfigure}
	    \begin{subfigure}[t]{0.15\textwidth}
   	    	\vspace{0px}\centering
            Low\\
	        \includegraphics[width=1.5cm,trim={\cropleft cm \croplower cm \cropright cm \cropupper cm},clip]{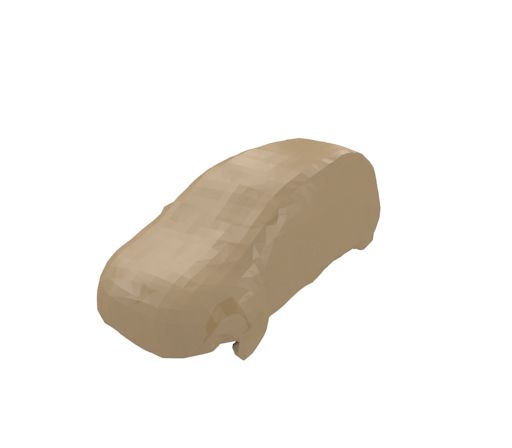}
	    \end{subfigure}
	    \begin{subfigure}[t]{0.15\textwidth}
   	    	\vspace{0px}\centering
            High\\
	        \includegraphics[width=1.5cm,trim={\cropleft cm \croplower cm \cropright cm \cropupper cm},clip]{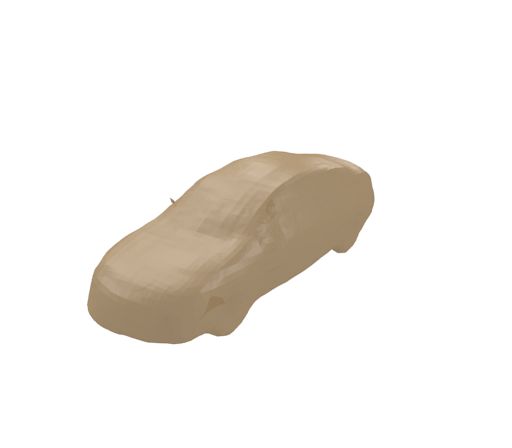}
	    \end{subfigure}
	    \begin{subfigure}[t]{0.15\textwidth}
   	    	\vspace{0px}\centering
            High\\
	        \includegraphics[width=1.5cm,trim={\cropleft cm \croplower cm \cropright cm \cropupper cm},clip]{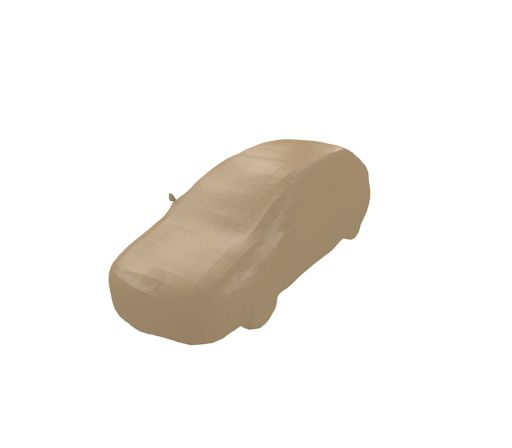}
	    \end{subfigure}
	    \begin{subfigure}[t]{0.15\textwidth}
   	    	\vspace{0px}\centering
            High\\
	        \includegraphics[width=1.5cm,trim={\cropleft cm \croplower cm \cropright cm \cropupper cm},clip]{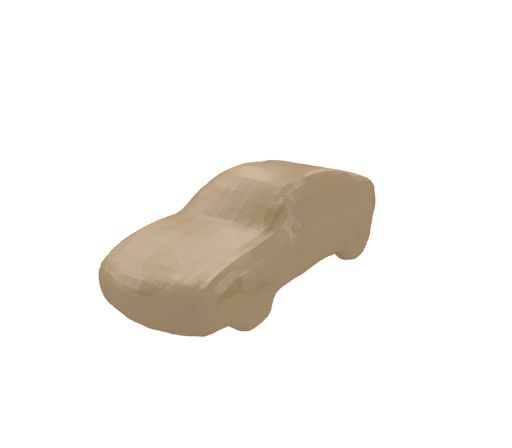}
	    \end{subfigure}
	    \\ 
	    \begin{subfigure}[t]{0.15\textwidth}
   	    	\vspace{0px}\centering
   	       	\includegraphics[width=1.5cm,trim={\cropleft cm \croplower cm \cropright cm \cropupper cm},clip]{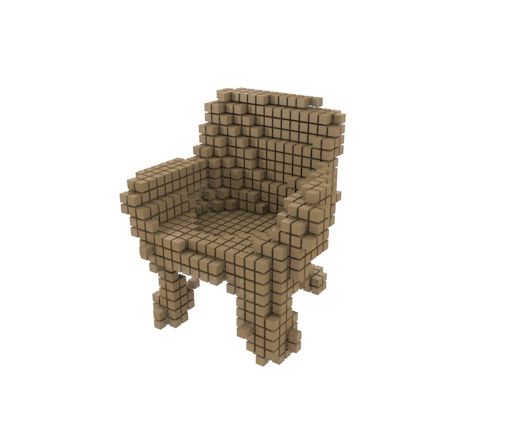}
	    \end{subfigure}
	    \begin{subfigure}[t]{0.15\textwidth}
   	    	\vspace{0px}\centering
   	       	\includegraphics[width=1.5cm,trim={\cropleft cm \croplower cm \cropright cm \cropupper cm},clip]{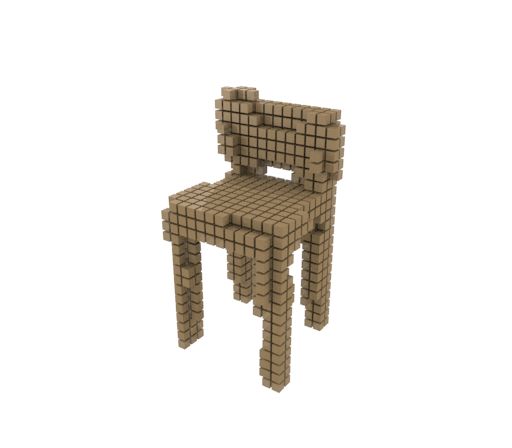}
	    \end{subfigure}
	    \begin{subfigure}[t]{0.15\textwidth}
   	    	\vspace{0px}\centering
   	       	\includegraphics[width=1.5cm,trim={\cropleft cm \croplower cm \cropright cm \cropupper cm},clip]{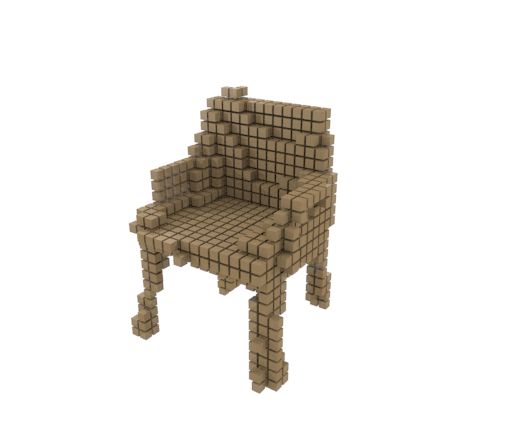}
	    \end{subfigure}
	    \begin{subfigure}[t]{0.15\textwidth}
   	    	\vspace{0px}\centering
   	       	\includegraphics[width=1.5cm,trim={\cropleft cm \croplower cm \cropright cm \cropupper cm},clip]{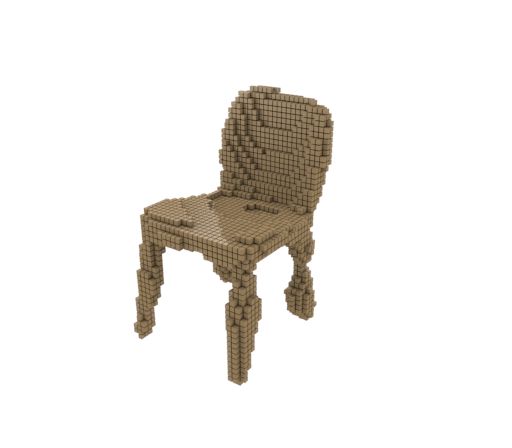}
	    \end{subfigure}
	    \begin{subfigure}[t]{0.15\textwidth}
   	    	\vspace{0px}\centering
   	       	\includegraphics[width=1.5cm,trim={\cropleft cm \croplower cm \cropright cm \cropupper cm},clip]{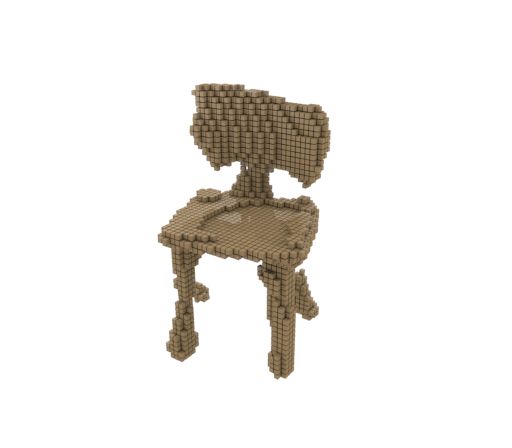}
	    \end{subfigure}
	    \begin{subfigure}[t]{0.15\textwidth}
   	    	\vspace{0px}\centering
   	       	\includegraphics[width=1.5cm,trim={\cropleft cm \croplower cm \cropright cm \cropupper cm},clip]{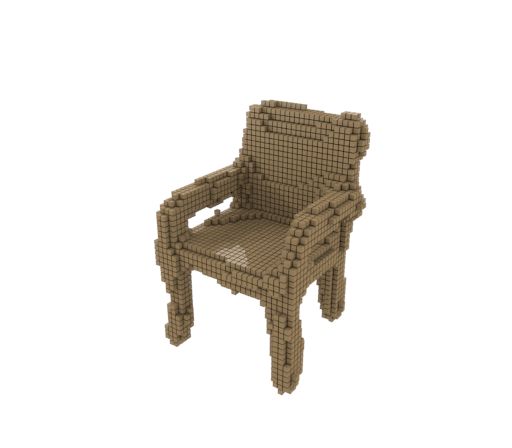}
	    \end{subfigure}
        \subcaption{Random Samples, Low and High Resolution (\cf \tabref{tab:data})}
    \end{subfigure}
    }

    \vspace*{-\figskipcaption px}
    \caption{{\bf \DVAE Shape Prior.} Reconstructions and random samples on ShapeNet and ModelNet at multiple resolutions (\cf \tabref{tab:data}); false negative and false positive voxels in {\color{rgreen}green} and {\color{rred}red}. Our \DVAE shape prior provides high-quality reconstructions and meaningful random samples across resolutions.}
    \label{fig:results-shape-prior}
    \vspace*{-\figskipbelow px}
\end{figure}

As depicted in \figref{fig:architectures}, our network architectures are kept simple and shallow. Considering a resolution of $24 \ntimes 54 \ntimes 24$ voxels on ShapeNet and KITTI, the encoder comprises three stages, each consisting of two convolutional layers (followed by $\text{ReLU}$ activations and batch normalization \citep{Ioffe2015ICML}) and max pooling; the decoder mirrors the encoder, replacing max pooling by nearest neighbor upsampling. We consistently use $3^3$ convolutional kernels. We use a latent space of size $Q = 10$ and predict occupancy using Sigmoid activations.

We found that the shape representation has a significant impact on training. Specifically, learning both occupancy grids and SDFs works better compared to training on SDFs only. Additionally, following prior art in single image depth prediction \citep{Eigen2015ICCV,Eigen2014NIPS,Laina2016THREEDV}, we consider log-transformed, truncated SDFs (logTSDFs) for training: given a signed distance $y_i$, we compute $\text{sign}(y_i)\log(1 + \min(5, |y_i|))$ as the corresponding log-transformed, truncated signed distance. TSDFs are commonly used in the literature \citep{Newcombe2011ISMAR,Riegler2017THREEDV,Dai2017CVPRa,Engelmann2016GCPR,Curless1996SIGGRAPH} and the logarithmic transformation additionally increases the relative importance of values around the surfaces (\ie, around the zero crossing).

\begin{figure}[tp]
    \vspace*{-\figskipabove px}
    \vspace{4px}
    \centering
    {\scriptsize
        
    \newcommand{\ablationa}{33}
    \newcommand{\ablationb}{99}
    \newcommand{\ablationc}{528}
    \newcommand{\ablationd}{693}
    \newcommand{\ablatione}{231}
    \newcommand{\ablationf}{396} 
    
    \begin{subfigure}[t]{0.01\textwidth}
        \vspace{0px}\centering
        \rotatebox[]{90}{\dAML\hspace*{0.25cm}}
    \end{subfigure}
    \begin{subfigure}[t]{0.07\textwidth}
        \vspace{0px}\centering
        \includegraphics[width=1.5cm,trim={\cropleft cm \croplower cm \cropright cm \cropupper cm},clip]{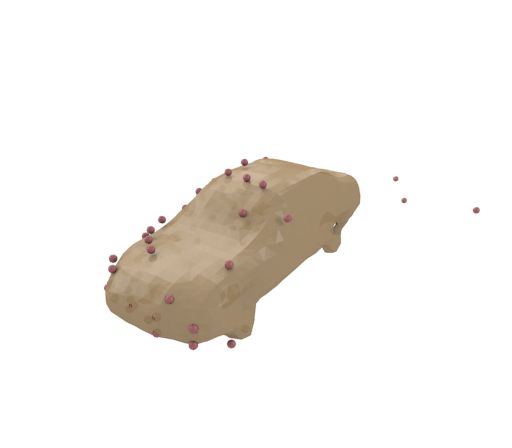}
    \end{subfigure}
    \begin{subfigure}[t]{0.07\textwidth}
        \vspace{0px}\centering
        \includegraphics[width=1.5cm,trim={\cropleft cm \croplower cm \cropright cm \cropupper cm},clip]{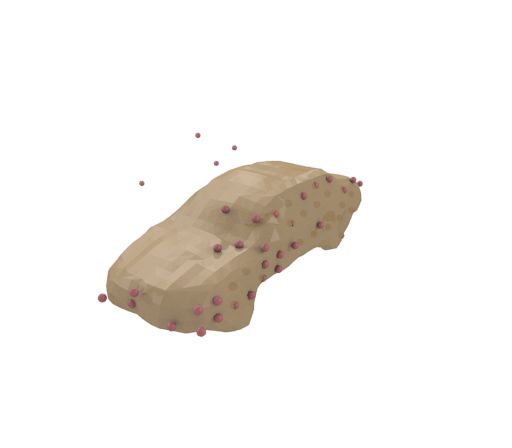}
    \end{subfigure}
    \begin{subfigure}[t]{0.07\textwidth}
        \vspace{0px}\centering
        \includegraphics[width=1.5cm,trim={\cropleft cm \croplower cm \cropright cm \cropupper cm},clip]{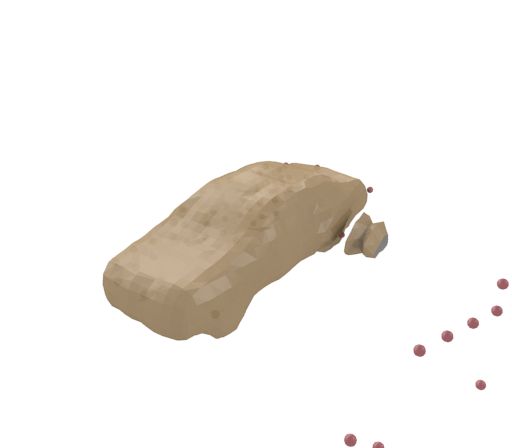}
    \end{subfigure}
    \begin{subfigure}[t]{0.07\textwidth}
        \vspace{0px}\centering
        \includegraphics[width=1.5cm,trim={\cropleft cm \croplower cm \cropright cm \cropupper cm},clip]{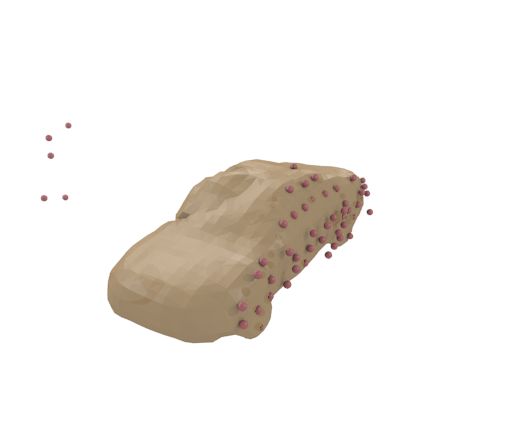}
    \end{subfigure}
    \begin{subfigure}[t]{0.07\textwidth}
        \vspace{0px}\centering
        \includegraphics[width=1.5cm,trim={\cropleft cm \croplower cm \cropright cm \cropupper cm},clip]{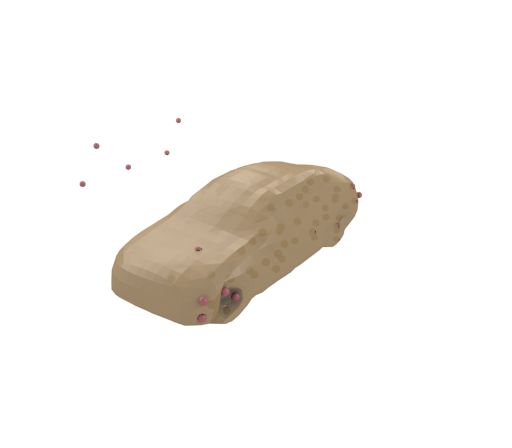}
    \end{subfigure}
    \begin{subfigure}[t]{0.07\textwidth}
        \vspace{0px}\centering
        \includegraphics[width=1.5cm,trim={\cropleft cm \croplower cm \cropright cm \cropupper cm},clip]{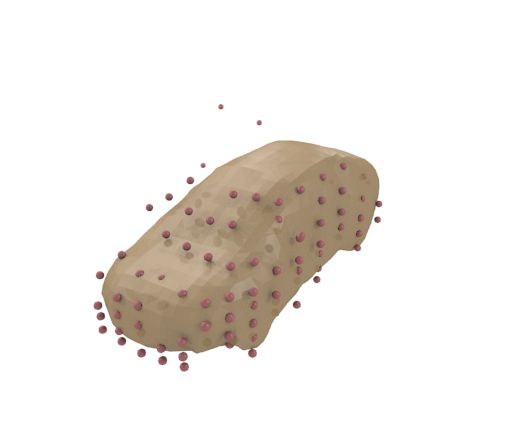}
    \end{subfigure}
    \\[-4px]
    \begin{subfigure}[t]{0.01\textwidth}
        \vspace{0px}\centering
        \rotatebox[]{90}{\AML\hspace*{0.25cm}}
    \end{subfigure}
    \begin{subfigure}[t]{0.07\textwidth}
        \vspace{0px}\centering
        \includegraphics[width=1.5cm,trim={\cropleft cm \croplower cm \cropright cm \cropupper cm},clip]{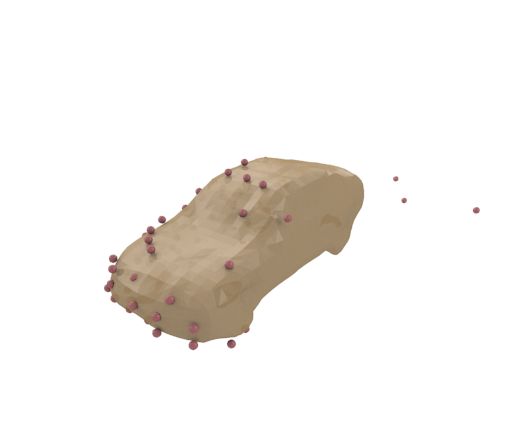}
    \end{subfigure}
    \begin{subfigure}[t]{0.07\textwidth}
        \vspace{0px}\centering
        \includegraphics[width=1.5cm,trim={\cropleft cm \croplower cm \cropright cm \cropupper cm},clip]{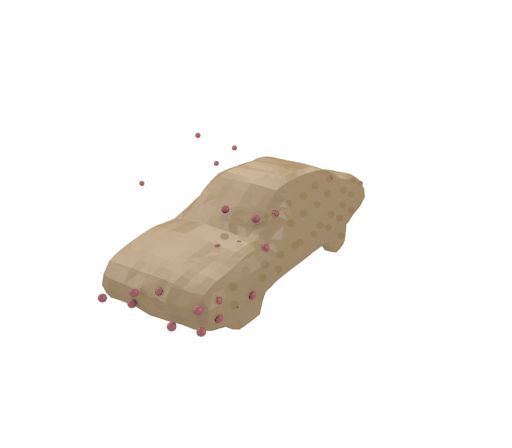}
    \end{subfigure}
    \begin{subfigure}[t]{0.07\textwidth}
        \vspace{0px}\centering
        \includegraphics[width=1.5cm,trim={\cropleft cm \croplower cm \cropright cm \cropupper cm},clip]{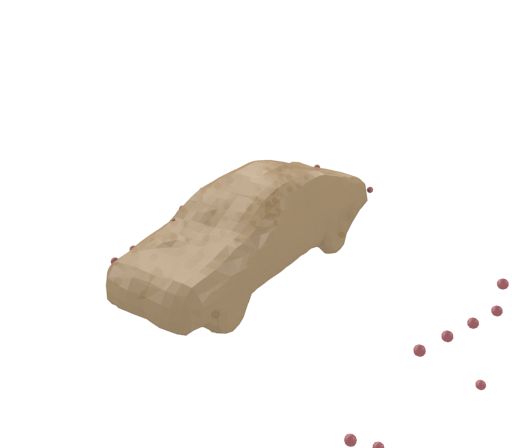}
    \end{subfigure}
    \begin{subfigure}[t]{0.07\textwidth}
        \vspace{0px}\centering
        \includegraphics[width=1.5cm,trim={\cropleft cm \croplower cm \cropright cm \cropupper cm},clip]{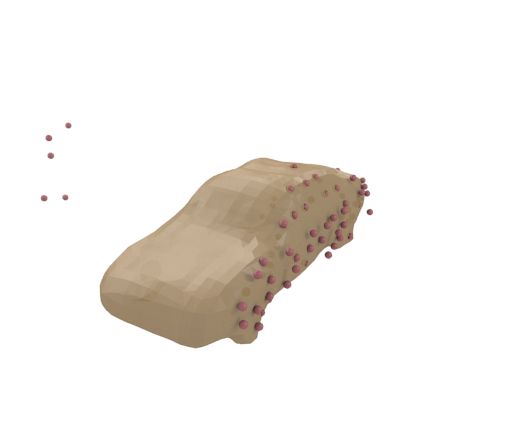}
    \end{subfigure}
    \begin{subfigure}[t]{0.07\textwidth}
        \vspace{0px}\centering
        \includegraphics[width=1.5cm,trim={\cropleft cm \croplower cm \cropright cm \cropupper cm},clip]{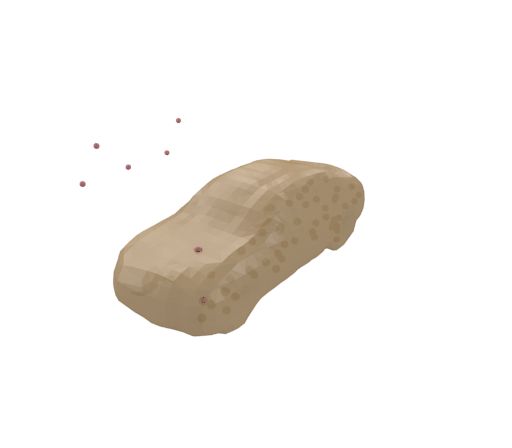}
    \end{subfigure}
    \begin{subfigure}[t]{0.07\textwidth}
        \vspace{0px}\centering
        \includegraphics[width=1.5cm,trim={\cropleft cm \croplower cm \cropright cm \cropupper cm},clip]{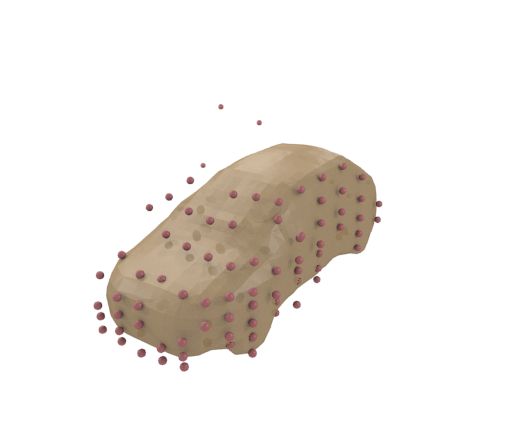}
    \end{subfigure}
    }
    \vspace*{-\figskipcaption px}
    \caption{{\bf Comparison of \AML and \dAML.} Our deterministic variant, \dAML, suffers from inferior results. Predicted shapes in {\color{rbeige}beige} and observations in {\color{rred}red} at low resolution ($24\ntimes54\ntimes24$ voxels).}
    \label{fig:results-shape-prior}
    \vspace*{-\figskipbelow px}
\end{figure}

For training, we combine occupancy grids and logTSDFs in separate feature channels and randomly translate both by up to $3$ voxels per axis. Additionally, we use Bernoulli noise (probability $0.1$) and Gaussian noise (variance $0.05$). We use Adam \citep{Kingma2015ICLR}, a batch size of $16$ and the initialization scheme by \cite{Glorot2010AISTATS}. The shape prior is trained for $3000$ to $4000$ epochs with an initial learning rate of $10^{-4}$ which is decayed by $0.925$ every $215$ iterations until a minimum of $10^{-16}$ has been reached. In addition, weight decay ($10^{-4}$) is applied. For shape inference, training takes $30$ to $50$ epochs, and an initial learning rate of $10^{-4}$ is decayed by $0.9$ every $215$ iterations. For our learning-based baselines (see \secref{sec:baselines}) we require between $300$ and $400$ epochs using the same training procedure as for the shape prior. On the \Kinect dataset, where only $30$ training examples are available, we used $5000$ epochs. We use $\log \sigma^2 = -2$ \red{as an empirically found trade-off between accuracy of the reconstructed SDFs and ease of training -- significantly lower $\log \sigma^2$ may lead to difficulties during training, including divergence.} \red{On ShapeNet, ModelNet and \Kinect, the weight $\lambda$ of the Kullback-Leibler divergence $\text{KL}$ (for both \DVAE and (d)AML) was empirically determined to be $\lambda = 2, 2.5, 3$ for low, medium and high resolution, respectively. On KITTI, we use $\lambda = 1$ for all resolutions. In practice, $\lambda$ controls the trade-off between diversity (low $\lambda$) and quality (high $\lambda$) of the completed shapes.} In addition, we reduce the weight in free space areas to one fourth on \noisy and KITTI to balance between occupied and free space. We implemented our networks in Torch \citep{Collobert2011NIPSWORK}.
\subsection{Baselines}
\label{sec:baselines}

\boldparagraph{Data-Driven Approaches}
{We consider the works by \cite{Engelmann2016GCPR} and \cite{Gupta2015CVPR} as data-driven baselines. Additionally, we consider regular maximum likelihood (\ML). \cite{Engelmann2016GCPR} -- referred to as \Engelmann\xspace-- use a principal component analysis shape prior trained on a manually selected set of car models\footnote{\url{https://github.com/VisualComputingInstitute/ShapePriors_GCPR16}}. Shape completion is posed as optimization problem considering both shape and pose. The pre-trained shape prior provided by Engelmann \etal assumes a ground plane which is, according to KITTI's LiDAR data, fixed at $1m$ height. Thus, we don't need to optimize pose on KITTI as we use the ground truth bounding boxes; on ShapeNet, in contrast, we need to optimize both pose and shape to deal with the random rotations in \clean and \noisy.}

\begin{figure}[tp]
    \vspace*{-\figskipabove px}
    \centering
    \hspace*{-22px}
    \begin{subfigure}[t]{0.49\linewidth}
        \includegraphics[height=3.1cm]{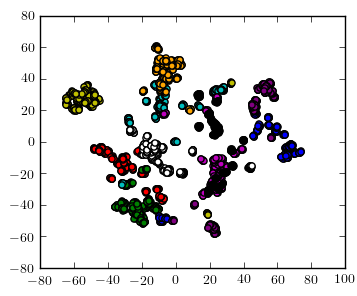}
        \subcaption{\bf \DVAE t-SNE}
        \label{fig:results-latent-space-a1}
    \end{subfigure}
    \hspace*{-12px}
    \begin{subfigure}[t]{0.49\linewidth}
        \includegraphics[height=3cm]{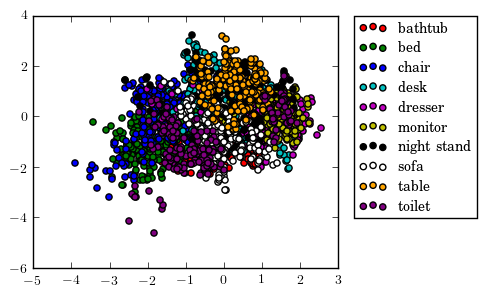}
        \subcaption{\bf \DVAE Projection}
        \label{fig:results-latent-space-a2}
    \end{subfigure}
    \\
    \hspace*{-22px}
    \begin{subfigure}[t]{0.49\linewidth}
        \includegraphics[height=3cm]{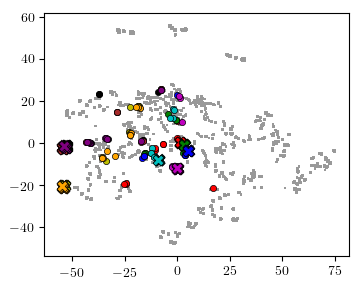}
        \subcaption{\bf \AML t-SNE}
        \label{fig:results-latent-space-b1}
    \end{subfigure}
    \hspace*{-12px}
    \begin{subfigure}[t]{0.49\linewidth}
        \includegraphics[height=3cm]{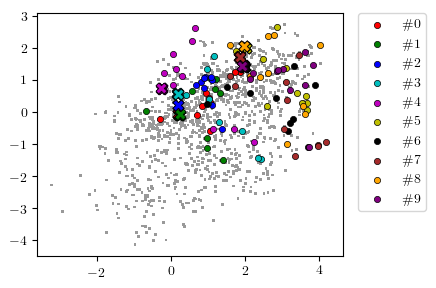}
        \subcaption{\bf \AML Projection}
        \label{fig:results-latent-space-b2}
    \end{subfigure}
    \vspace*{-\figskipcaption px}
    \caption{{\bf Learned Latent Spaces.} In (\subref{fig:results-latent-space-a1}) and (\subref{fig:results-latent-space-a2}), we show a t-SNE \citep{Maaten2008JMLR} visualization and a two-dimensional projection of the \DVAE latent space on ModelNet10. The plots illustrate that the \DVAE is able to separate the ten object categories. In (\subref{fig:results-latent-space-b1}) and (\subref{fig:results-latent-space-b2}), we show a t-SNE visualization and a projection of the latent space corresponding to our learned \AML model on \clean. We randomly picked $10$ ground truth shapes,  ``x'', and the corresponding observations ($10$ per shape), points (gray pixels indicate remaining shapes/observations). The plots illustrate that \AML is able to associate observations with the corresponding ground truth shapes under weak supervision.}
    \label{fig:results-latent-space}
    \vspace*{-\figskipbelow px}
\end{figure}
\begin{table*}[t]
    \vspace*{-\figskipabove px}
    \centering
    {\scriptsize
        \begin{tabularx}{1\textwidth}{|@{  }l@{  }|@{  }X@{  }|@{  }c@{  }c@{  }c@{  }c@{  }|@{  }c@{  }c@{  }c@{  }c@{  }|@{  }c@{  }|}
            \hline
             \multicolumn{1}{|@{  }c@{  }|@{  }}{Supervision} & \multicolumn{1}{@{  }c@{  }|@{  }}{Method} & \multicolumn{4}{@{  }c@{  }|@{  }}{\clean} & \multicolumn{4}{@{  }c@{  }|@{  }}{\noisy} & \multicolumn{1}{c|}{KITTI}\\
             \multicolumn{1}{|@{  }c@{  }|@{  }}{in \%} && \Abs {\tiny $\downarrow$} & \IoU {\tiny $\uparrow$} & \Acc [vx] {\tiny $\downarrow$} & \Compl [vx] {\tiny $\downarrow$} & \Abs {\tiny $\downarrow$} & \IoU {\tiny $\uparrow$} & \Acc [vx] {\tiny $\downarrow$} & \Compl [vx] {\tiny $\downarrow$} & \Compl [m] {\tiny $\downarrow$} \\
            \hline\hline
            \multicolumn{11}{|c|}{Low Resolution: $24 \times 54 \times 24$ voxels; * independent of resolution}\\
            \hline\hline
            {\color{darkgray}(shape prior)} & {\leavevmode\color{darkgray}\DVAE} & {\color{darkgray}0.019} & {\color{darkgray}0.885} & {\color{darkgray}0.283} & {\color{darkgray}0.527} & \multicolumn{5}{@{  }c@{  }|}{{\color{darkgray}(same shape prior as on \clean)}}\\
            \hline\hline
            \multirow{2}{*}{$\hphantom{<}100$} & \cite{Dai2017CVPRa} (\Dai) & {\bf\color{rred} 0.021} & {\bf\color{rred} 0.872} & {\bf\color{rred} 0.321} & {\bf\color{rred} 0.564} & {\bf\color{rred} 0.027} & {\bf\color{rred} 0.836} & {\bf\color{rred} 0.391} & {\bf\color{rred} 0.633} & 0.128\\
            &\Sup & 0.026 & 0.841 & 0.409 & 0.607 & 0.028 & 0.833 & 0.407 & 0.637 & {\bf\color{rred} 0.091}\\
            \hline
            \multirow{10}{*}{$<7.7$} &\BL & 0.067 & 0.596 & 0.999 & 1.335 & 0.064 & 0.609 & 0.941 & 1.29 & \color{darkgray}--\\
            &\M & 0.052 & 0.697 & 0.79 & 0.938 & 0.052 & 0.696 & 0.79 & 0.938 & \color{darkgray}--\\
            &\ML & 0.04 & 0.756 & 0.637 & 0.8 & 0.041 & 0.755 & 0.625 & 0.829 & \color{darkgray}(too slow)\\
            & *\cite{Gupta2015CVPR} (\ICP) & \multicolumn{2}{@{  }c@{  }}{\color{darkgray}(mesh only)} & 0.534 & {\bf\color{rgreen} 0.503} & \multicolumn{2}{@{  }c@{  }}{\color{darkgray}(mesh only)} & 7.551 & 6.372 & \color{darkgray}(too slow)\\
            & *\cite{Engelmann2016GCPR} (\Engelmann) & \multicolumn{2}{@{  }c@{  }}{\color{darkgray}(mesh only)} & 1.235 & 1.237 & \multicolumn{2}{@{  }c@{  }}{\color{darkgray}(mesh only)} & 1.974 & 1.312 & 0.13\\ 
            &\dAML & {\bf\color{rgreen} 0.034} & {\bf\color{rgreen} 0.784} & {\bf\color{rgreen} 0.532} & 0.741 & {\bf\color{rgreen} 0.036} & {\bf\color{rgreen} 0.772} & {\bf\color{rgreen} 0.557} & {\bf\color{rgreen} 0.76} & \color{darkgray}(see \AML)\\
            &\AML & {\bf\color{rgreen} 0.034} & 0.779 & 0.549 & 0.753 & {\bf\color{rgreen} 0.036} & 0.771 & 0.57 & 0.761 & {\bf\color{rgreen} 0.12}\\
            \hline\hline
            \multicolumn{11}{|c|}{Low Resolution: $24 \times 54 \times 24$ voxels; Multiple, $k > 1$ Fused Views}\\
            \hline\hline
            \multirow{2}{*}{$\hphantom{<}100$} & \cite{Dai2017CVPRa} (\Dai), $k = 5$ & \bf\color{rred}0.012 & \bf\color{rred}0.924 & \bf\color{rred}0.214 & \bf\color{rred}0.436 & \bf\color{rred}0.018 & \bf\color{rred}0.887 & \bf\color{rred}0.278 & \bf\color{rred}0.491 &\multirow{2}{*}{\color{darkgray}n/a}\\
            &\Sup, $k = 5$ & 0.022 & 0.866 & 0.336 & 0.566 & 0.024 & 0.86 & 0.331 & 0.573 &\\
            \hline
            $<16$ & \AML, $k = 2$ & 0.032 & 0.794 & 0.489 & 0.695 & 0.034 & 0.79 & 0.52 & 0.725 & \multirow{3}{*}{\color{darkgray}n/a}\\
            $<24$ & \AML, $k = 3$ & {\bf\color{rgreen} 0.031} & {\bf\color{rgreen} 0.809} & {\bf\color{rgreen} 0.471} & {\bf\color{rgreen} 0.667} & {\bf\color{rgreen} 0.031} & {\bf\color{rgreen} 0.81} & {\bf\color{rgreen} 0.493} & {\bf\color{rgreen} 0.67} &\\
            $<40$ & \AML, $k = 5$ & {\bf\color{rgreen} 0.031} & 0.804 & 0.502 & 0.686 & 0.035 & 0.799 & 0.523 & 0.7 &\\
            \hline\hline
            \multicolumn{11}{|c|}{Medium Resolution: $32 \times 72 \times 32$ voxels}\\
            \hline\hline
            {\color{darkgray}(shape prior)} & {\leavevmode\color{darkgray}\DVAE} & {\color{darkgray}0.019} & {\color{darkgray}0.877} & {\color{darkgray}0.24} & {\color{darkgray}0.47} & \multicolumn{5}{@{  }c@{  }|}{{\color{darkgray}(same shape prior as on \clean)}}\\
            \hline\hline 
            \multirow{2}{*}{$\hphantom{<}100$} & \cite{Dai2017CVPRa} (\Dai) & {\bf\color{rred} 0.02} & {\bf\color{rred} 0.869} & {\bf\color{rred} 0.399} & {\bf\color{rred} 0.674} & {\bf\color{rred} 0.026} & {\bf\color{rred} 0.83} & {\bf\color{rred} 0.51} & {\bf\color{rred} 0.767} & {\bf\color{rred} 0.074}\\
            &\Sup & 0.027 & 0.834 & 0.498 & 0.789 & 0.029 & 0.815 & 0.571 & 0.843 & 0.09\\
            \hline
            $\leq6.1$ & \AML & {\bf\color{rgreen} 0.031} & {\bf\color{rgreen} 0.788} & {\bf\color{rgreen} 0.415} & {\bf\color{rgreen} 0.584} & {\bf\color{rgreen} 0.036} & {\bf\color{rgreen} 0.766} & {\bf\color{rgreen} 0.721} & {\bf\color{rgreen} 0.953} & {\bf\color{rgreen} 0.083}\\
            \hline\hline
            \multicolumn{11}{|c|}{High Resolution: $48 \times 108 \times 48$ voxels}\\
            \hline\hline
            {\color{darkgray}(shape prior)} & {\leavevmode\color{darkgray}\DVAE} & {\color{darkgray}0.018} & \color{darkgray}0.87 & {\color{darkgray}0.272} & {\color{darkgray}0.434} & \multicolumn{5}{@{  }c@{  }|}{{\color{darkgray}(same shape prior as on \clean)}}\\
            \hline\hline
            \multirow{2}{*}{$\hphantom{<}100$} & \Dai & {\bf\color{rred} 0.017} & {\bf\color{rred} 0.88} & {\bf\color{rred} 0.517} & {\bf\color{rred} 0.827} & 0.054 & 0.664 & 1.559 & 2.067 & {\bf\color{rred} 0.066}\\
            &\Sup & 0.023 & 0.843 & 0.677 & 1.032 & {\bf\color{rred} 0.052} & {\bf\color{rred} 0.674} & {\bf\color{rred} 1.52} & {\bf\color{rred} 1.981} & 0.091\\
            \hline
            $<3.5$ & \AML & {\bf\color{rgreen} 0.028} & {\bf\color{rgreen} 0.796} & {\bf\color{rgreen} 0.433} & {\bf\color{rgreen} 0.579} & {\bf\color{rgreen} 0.045} & {\bf\color{rgreen} 0.659} & {\bf\color{rgreen} 1.4} & {\bf\color{rgreen} 1.957} & {\bf\color{rgreen} 0.078}\\
            \hline
        \end{tabularx}
    }
    \vspace*{-\figskipcaption px}
    \caption{{\bf Quantitative Results on ShapeNet and KITTI.} We consider Hamming distance (\Abs) and intersection over union (\IoU) for occupancy grids as well as accuracy (\Acc) and completeness (\Compl) for meshes on \clean, \noisy and KITTI. For \Abs, \Acc and \Compl, lower is better; for \IoU, higher is better. The unit of \Acc and \Compl is voxels (voxel length at $24 \ntimes 54 \ntimes 48$ voxels) or meters. Note that the \DVAE shape prior (in {\color{darkgray}gray}) is only reported as reference (\ie, bound on (d)\AML). We indicate the level of supervision in percentage, relative to the corresponding resolution \red{(see \tabref{tab:data})} and mark the best results under full supervision in {\color{rred}\bf red} and under weak supervision in {\color{rgreen}\bf green}.}
    \label{tab:results-shapenet}
    \vspace*{-\figskipbelow px}
\end{table*}

Inspired by the work by \cite{Gupta2015CVPR} we also consider a shape retrieval and fitting baseline. Specifically, we perform iterative closest point (\ICP) \citep{Besl1992PAMI} fitting on all training shapes and subsequently select the best-fitting one. To this end, we uniformly sample $1\text{Mio}$ points on the training shapes, and perform point-to-point \ICP\footnote{\url{http://www.cvlibs.net/software/libicp/}.} for a maximum of $100$ iterations using $\left[\begin{matrix}R & t\end{matrix}\right] = \left[\begin{matrix}I_3 & 0\end{matrix}\right]$ as initialization. On the training set, we verified that this approach is always able to retrieve the perfect shape.

Finally, we consider a simple \ML baseline iteratively minimizing \eqnref{eq:ml} using stochastic gradient descent (SGD). This baseline is similar to the work by Engelmann \etal, however, like ours it is bound to the voxel grid. Per example, we allow a maximum of $5000$ iterations, starting with latent code $z = 0$, learning rate $0.05$ and momentum $0.5$ (decayed every $50$ iterations at rate $0.85$ and $1.0$ until $10^{-5}$ and $0.9$ have been reached).

\boldparagraph{Learning-Based Approaches}
Learning-based approaches usually employ an encoder-decoder architecture to directly learn a mapping from observations $x_n$ to ground truth shapes $y_n^*$ in a fully supervised setting \citep{Wang2017ICCV,Varley2017IROS,Yang2018ARXIVb,Yang2017ARXIV,Dai2017CVPRa}. While existing architectures differ slightly, they usually rely on a U-net architecture \citep{Ronneberger2015MICCAI,Cicek2016ARXIV}. In this paper, we use the approach of \cite{Dai2017CVPRa}\footnote{
    We use \url{https://github.com/angeladai/cnncomplete}. On ModelNet we added one convolutional stage in the en- and decoder for larger resolutions; on ShapeNet and KITTI, we needed to adapt the convolutional strides to fit the corresponding resolutions.
} -- referred to as \Dai\xspace --
as a representative baseline for this class of approaches. In addition, we consider a custom learning-based baseline which uses the architecture of our \DVAE shape prior, \cf \figref{fig:architectures}. In contrast to \citep{Dai2017CVPRa}, this baseline is also limited by the low-dimensional ($Q = 10$) bottleneck as it does not use skip connections.
\newcommand{\cleana}{231} 
\newcommand{\cleanb}{297}

\newcommand{\noisya}{132}
\newcommand{\noisyb}{66}

\newcommand{\bathtuba}{792}
\newcommand{\bathtubb}{330}

\newcommand{\chaira}{528}
\newcommand{\chairb}{990} 

\newcommand{\deska}{264} 
\newcommand{\deskb}{858}

\newcommand{\tablea}{858}
\newcommand{\tableb}{396} 
\begin{figure*}[t]
    \vspace*{-\figskipabove px}
    \vspace*{2px}
    \centering
    {\scriptsize
    \begin{subfigure}[t]{1\textwidth}
        \begin{subfigure}[t]{0.095\textwidth}
        	\vspace{0px}\centering
        	Obs\\
        	\includegraphics[width=1.8cm,trim={\cropleft cm \croplower cm \cropright cm \cropupper cm},clip]{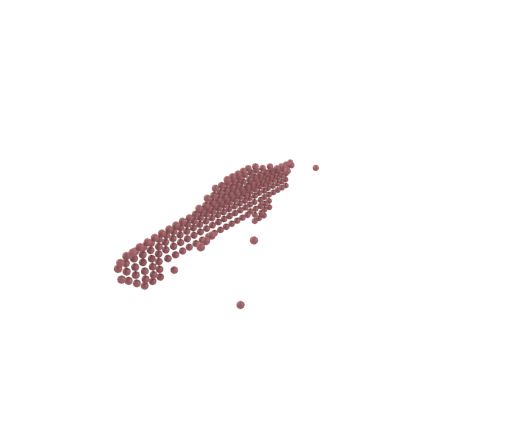}
        \end{subfigure}
        \begin{subfigure}[t]{0.095\textwidth}
        	\vspace{0px}\centering
        	\Dai\\
        	\includegraphics[width=1.8cm,trim={\cropleft cm \croplower cm \cropright cm \cropupper cm},clip]{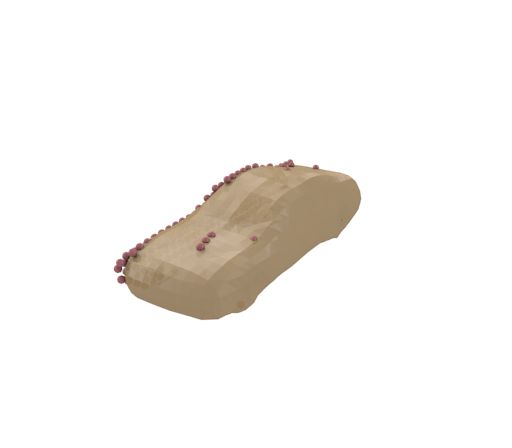}
        \end{subfigure}
        \begin{subfigure}[t]{0.095\textwidth}
            \vspace{0px}\centering
            \Dai\\
            \includegraphics[width=1.8cm,trim={\cropleft cm \croplower cm \cropright cm \cropupper cm},clip]{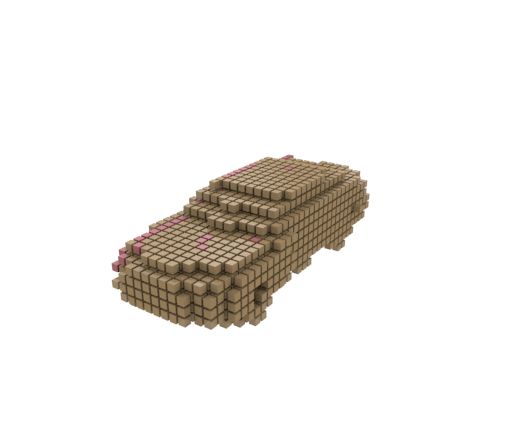}
        \end{subfigure}
        \begin{subfigure}[t]{0.095\textwidth}
            \vspace{0px}\centering
            \Engelmann\\
            \includegraphics[width=1.8cm,trim={\cropleft cm \croplower cm \cropright cm \cropupper cm},clip]{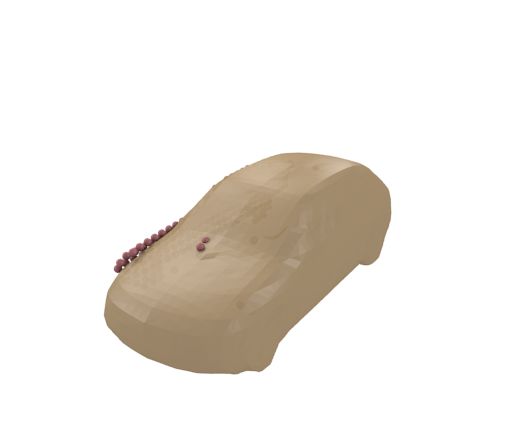}
        \end{subfigure}
        \begin{subfigure}[t]{0.095\textwidth}
            \vspace{0px}\centering
            \ML\\
            \includegraphics[width=1.8cm,trim={\cropleft cm \croplower cm \cropright cm \cropupper cm},clip]{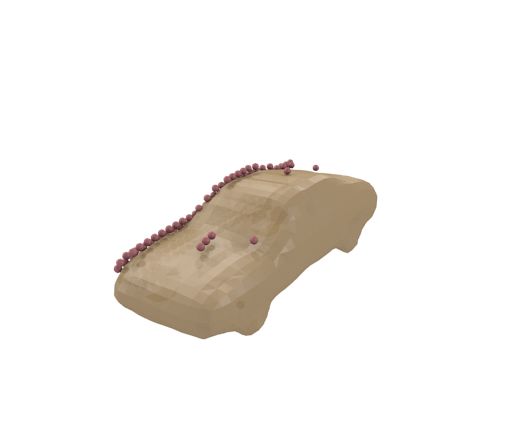}
        \end{subfigure}
        \begin{subfigure}[t]{0.095\textwidth}
            \vspace{0px}\centering
            \ML\\
            \includegraphics[width=1.8cm,trim={\cropleft cm \croplower cm \cropright cm \cropupper cm},clip]{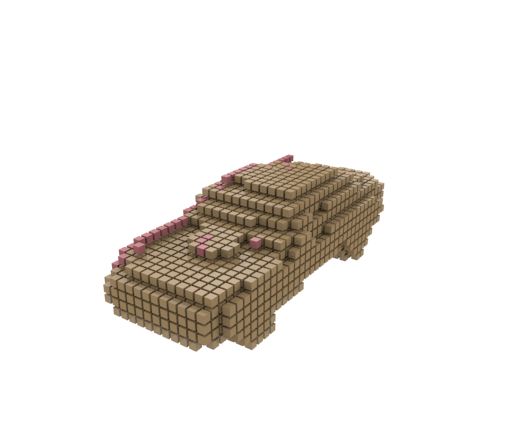}
        \end{subfigure}
        \begin{subfigure}[t]{0.095\textwidth}
        	\vspace{0px}\centering
        	\AML\\
        	\includegraphics[width=1.8cm,trim={\cropleft cm \croplower cm \cropright cm \cropupper cm},clip]{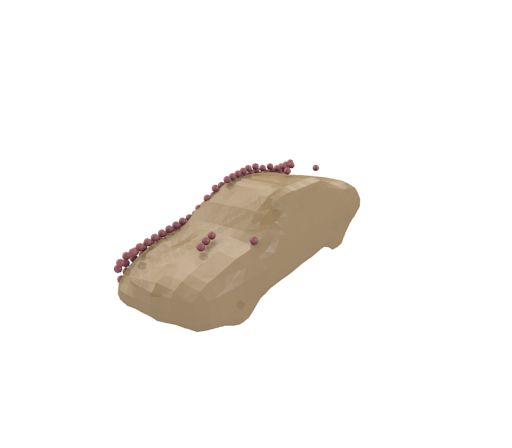}
        \end{subfigure}
        \begin{subfigure}[t]{0.095\textwidth}
            \vspace{0px}\centering
            \AML\\
            \includegraphics[width=1.8cm,trim={\cropleft cm \croplower cm \cropright cm \cropupper cm},clip]{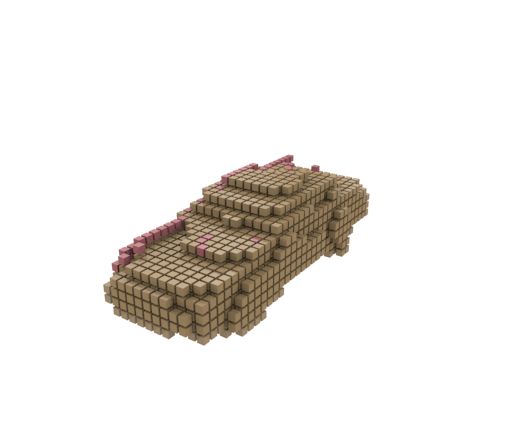}
        \end{subfigure}
        \begin{subfigure}[t]{0.095\textwidth}
        	\vspace{0px}\centering
        	GT\\
        	\includegraphics[width=1.8cm,trim={\cropleft cm \croplower cm \cropright cm \cropupper cm},clip]{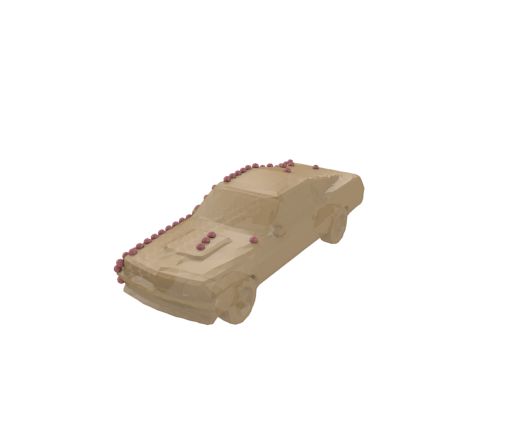}
        \end{subfigure}
        \begin{subfigure}[t]{0.095\textwidth}
            \vspace{0px}\centering
            GT\\
            \includegraphics[width=1.8cm,trim={\cropleft cm \croplower cm \cropright cm \cropupper cm},clip]{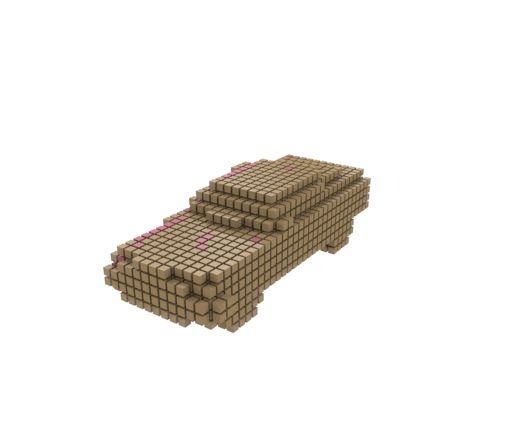}
        \end{subfigure}
        \\[-2px]
        \begin{subfigure}[t]{0.095\textwidth}
        	\vspace{0px}\centering
        	\includegraphics[width=1.8cm,trim={\cropleft cm \croplower cm \cropright cm \cropupper cm},clip]{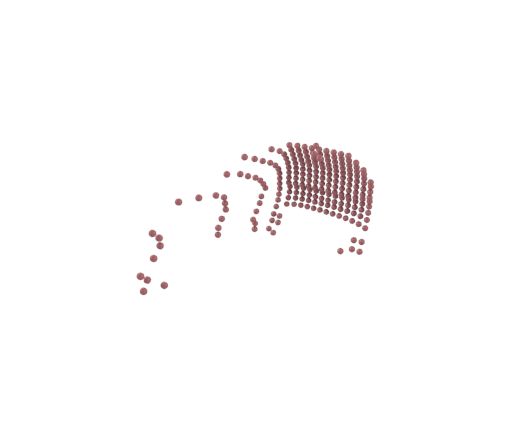}
        \end{subfigure}
        \begin{subfigure}[t]{0.095\textwidth}
        	\vspace{0px}\centering
        	\includegraphics[width=1.8cm,trim={\cropleft cm \croplower cm \cropright cm \cropupper cm},clip]{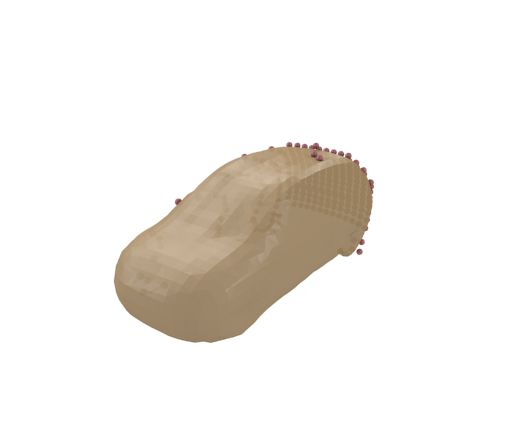}
        \end{subfigure}
        \begin{subfigure}[t]{0.095\textwidth}
        	\vspace{0px}\centering
        	\includegraphics[width=1.8cm,trim={\cropleft cm \croplower cm \cropright cm \cropupper cm},clip]{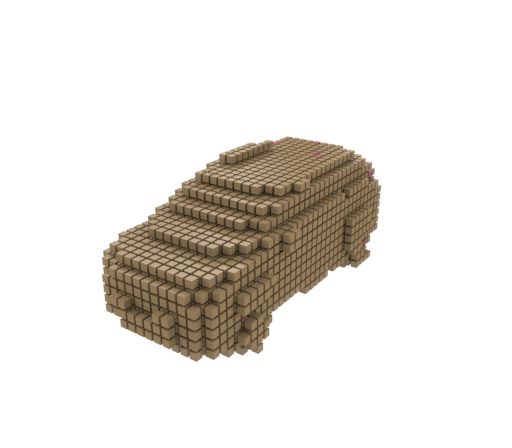}
        \end{subfigure}
        \begin{subfigure}[t]{0.095\textwidth}
        	\vspace{0px}\centering
        	\includegraphics[width=1.8cm,trim={\cropleft cm \croplower cm \cropright cm \cropupper cm},clip]{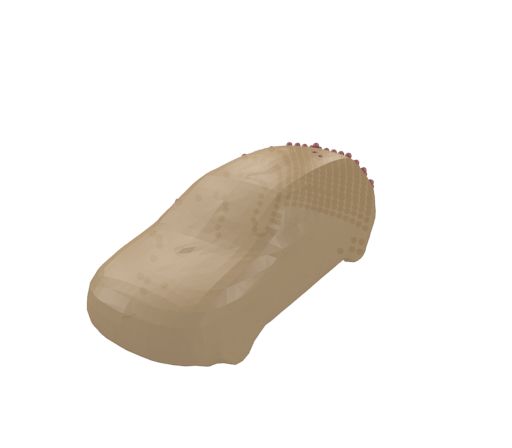}
        \end{subfigure}
        \begin{subfigure}[t]{0.095\textwidth}
            \vspace{0px}\centering
            \includegraphics[width=1.8cm,trim={\cropleft cm \croplower cm \cropright cm \cropupper cm},clip]{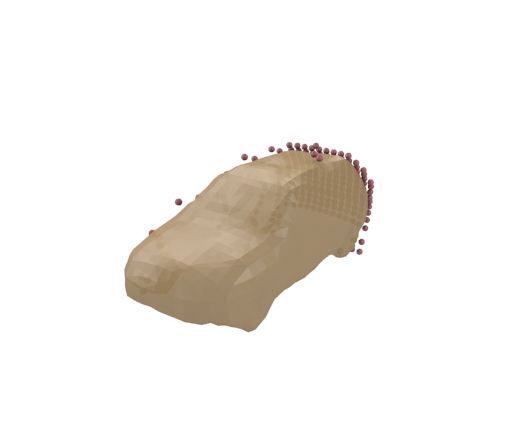}
        \end{subfigure}
        \begin{subfigure}[t]{0.095\textwidth}
        	\vspace{0px}\centering
        	\includegraphics[width=1.8cm,trim={\cropleft cm \croplower cm \cropright cm \cropupper cm},clip]{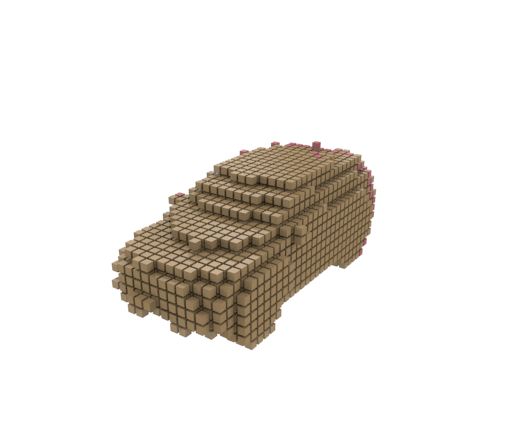}
        \end{subfigure}
        \begin{subfigure}[t]{0.095\textwidth}
        	\vspace{0px}\centering
        	\includegraphics[width=1.8cm,trim={\cropleft cm \croplower cm \cropright cm \cropupper cm},clip]{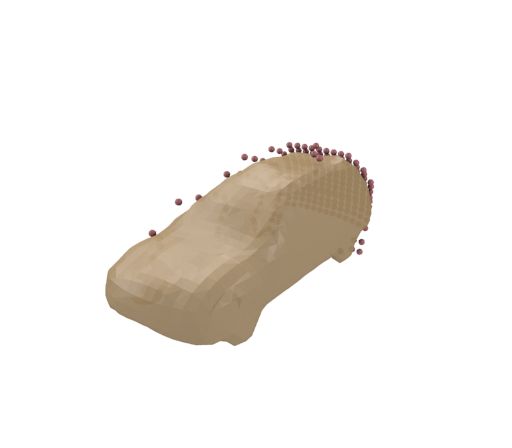}
        \end{subfigure}
        \begin{subfigure}[t]{0.095\textwidth}
        	\vspace{0px}\centering
        	\includegraphics[width=1.8cm,trim={\cropleft cm \croplower cm \cropright cm \cropupper cm},clip]{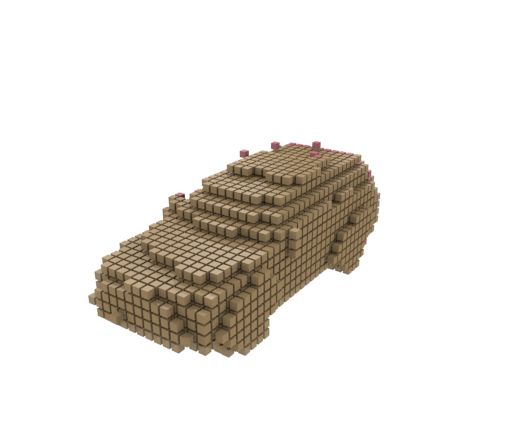}
        \end{subfigure}
        \begin{subfigure}[t]{0.095\textwidth}
        	\vspace{0px}\centering
        	\includegraphics[width=1.8cm,trim={\cropleft cm \croplower cm \cropright cm \cropupper cm},clip]{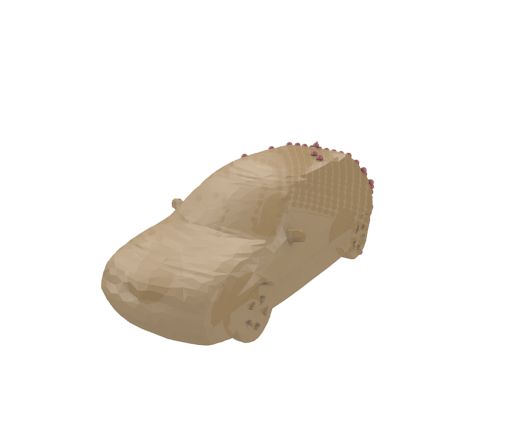}
        \end{subfigure}
        \begin{subfigure}[t]{0.095\textwidth}
        	\vspace{0px}\centering
        	\includegraphics[width=1.8cm,trim={\cropleft cm \croplower cm \cropright cm \cropupper cm},clip]{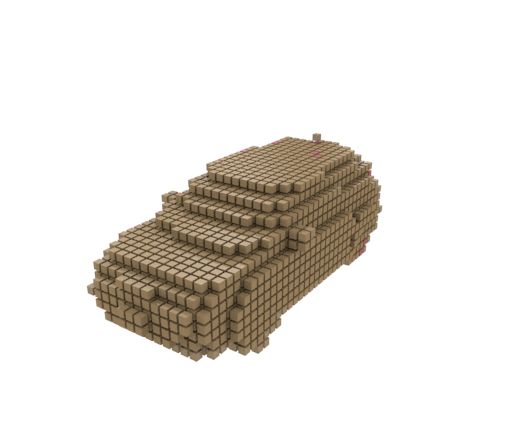}
        \end{subfigure}
        \begin{subfigure}[t]{0.095\textwidth}
        	\vspace{0px}\centering
        	\includegraphics[width=1.8cm,trim={\cropleft cm \croplower cm \cropright cm \cropupper cm},clip]{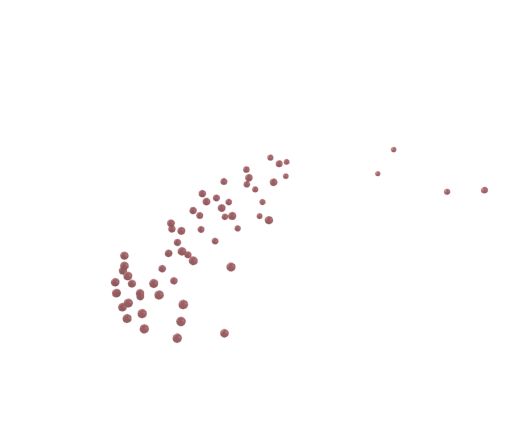}
        \end{subfigure}
        \begin{subfigure}[t]{0.095\textwidth}
        	\vspace{0px}\centering
        	\includegraphics[width=1.8cm,trim={\cropleft cm \croplower cm \cropright cm \cropupper cm},clip]{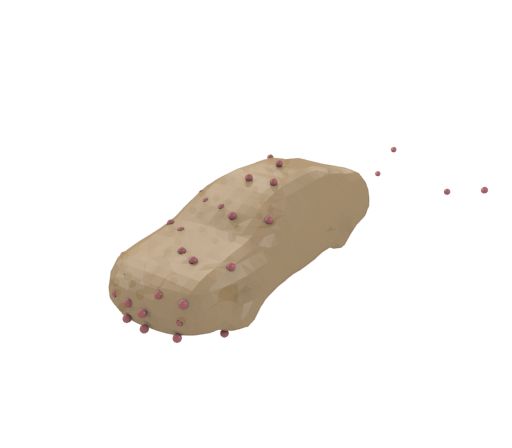}
        \end{subfigure}
        \begin{subfigure}[t]{0.095\textwidth}
            \vspace{0px}\centering
            \includegraphics[width=1.8cm,trim={\cropleft cm \croplower cm \cropright cm \cropupper cm},clip]{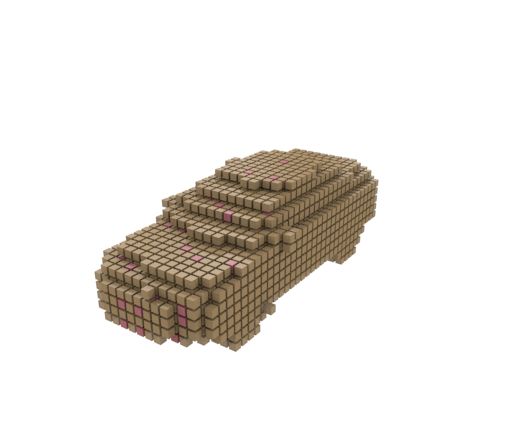}
        \end{subfigure}
        \begin{subfigure}[t]{0.095\textwidth}
        	\vspace{0px}\centering
        	\includegraphics[width=1.8cm,trim={\cropleft cm \croplower cm \cropright cm \cropupper cm},clip]{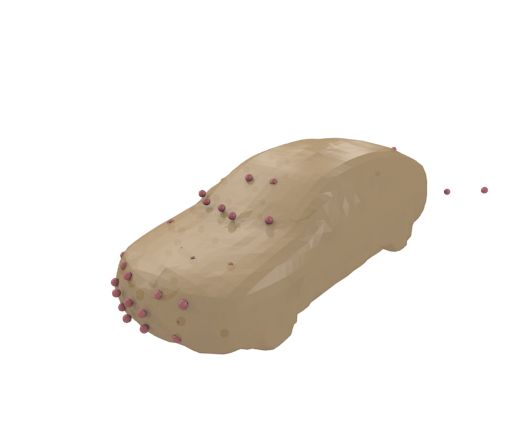}
        \end{subfigure}
        \begin{subfigure}[t]{0.095\textwidth}
            \vspace{0px}\centering
            \includegraphics[width=1.8cm,trim={\cropleft cm \croplower cm \cropright cm \cropupper cm},clip]{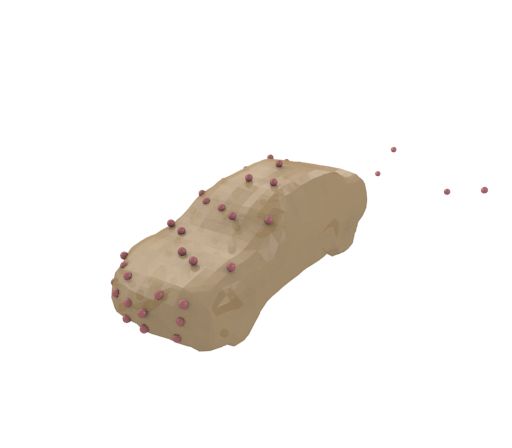}
        \end{subfigure}
        \begin{subfigure}[t]{0.095\textwidth}
            \vspace{0px}\centering
            \includegraphics[width=1.8cm,trim={\cropleft cm \croplower cm \cropright cm \cropupper cm},clip]{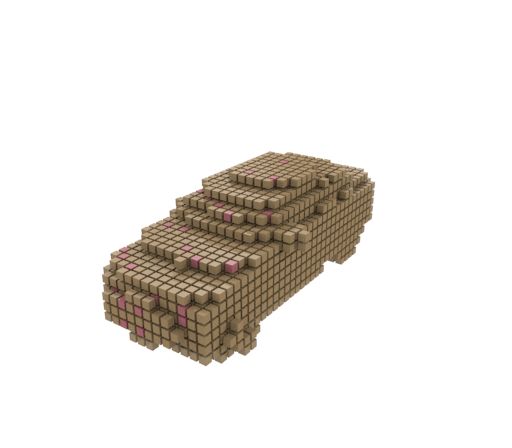}
        \end{subfigure}
        \begin{subfigure}[t]{0.095\textwidth}
        	\vspace{0px}\centering
        	\includegraphics[width=1.8cm,trim={\cropleft cm \croplower cm \cropright cm \cropupper cm},clip]{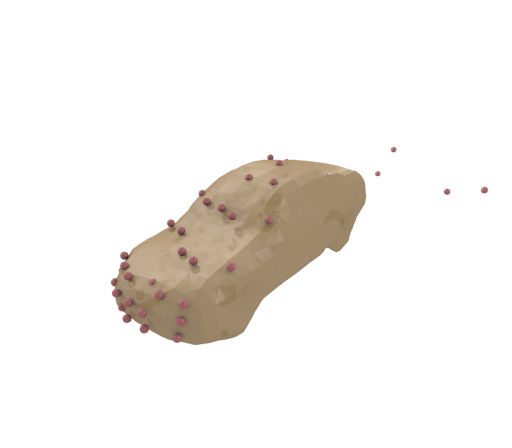}
        \end{subfigure}
        \begin{subfigure}[t]{0.095\textwidth}
        	\vspace{0px}\centering
        	\includegraphics[width=1.8cm,trim={\cropleft cm \croplower cm \cropright cm \cropupper cm},clip]{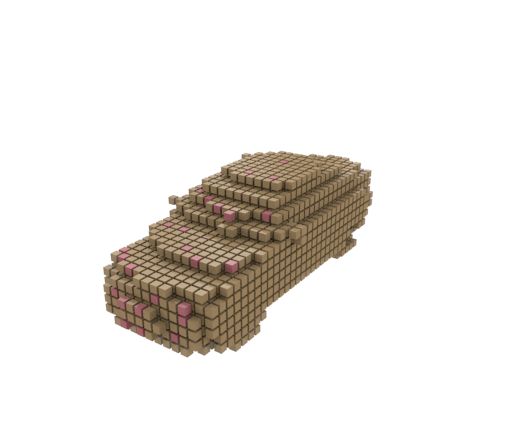}
        \end{subfigure}
        \begin{subfigure}[t]{0.095\textwidth}
        	\vspace{0px}\centering
        	\includegraphics[width=1.8cm,trim={\cropleft cm \croplower cm \cropright cm \cropupper cm},clip]{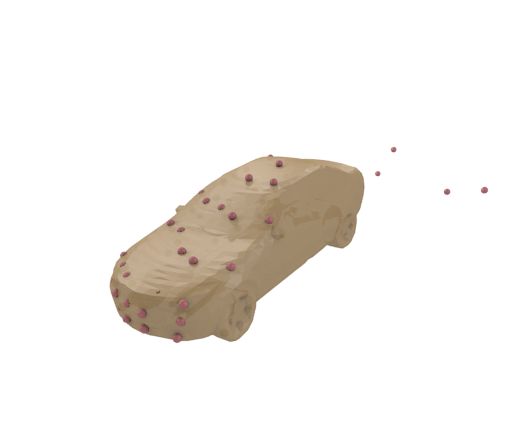}
        \end{subfigure}
        \begin{subfigure}[t]{0.095\textwidth}
            \vspace{0px}\centering
            \includegraphics[width=1.8cm,trim={\cropleft cm \croplower cm \cropright cm \cropupper cm},clip]{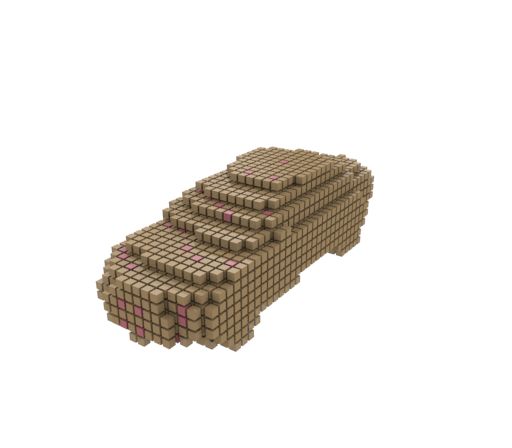}
        \end{subfigure}
        \\[-2px]
        \begin{subfigure}[t]{0.095\textwidth}
        	\vspace{0px}\centering
        	\includegraphics[width=1.8cm,trim={\cropleft cm \croplower cm \cropright cm \cropupper cm},clip]{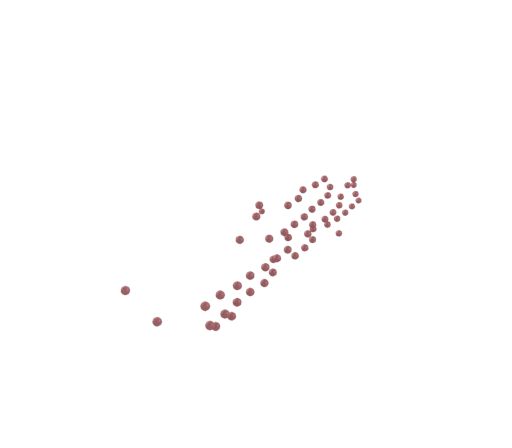}
        \end{subfigure}
        \begin{subfigure}[t]{0.095\textwidth}
        	\vspace{0px}\centering
        	\includegraphics[width=1.8cm,trim={\cropleft cm \croplower cm \cropright cm \cropupper cm},clip]{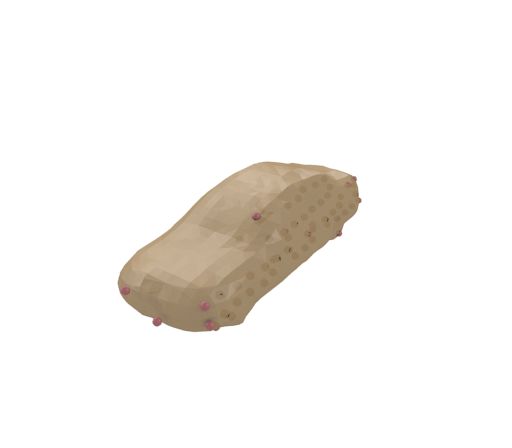}
        \end{subfigure}
        \begin{subfigure}[t]{0.095\textwidth}
        	\vspace{0px}\centering
        	\includegraphics[width=1.8cm,trim={\cropleft cm \croplower cm \cropright cm \cropupper cm},clip]{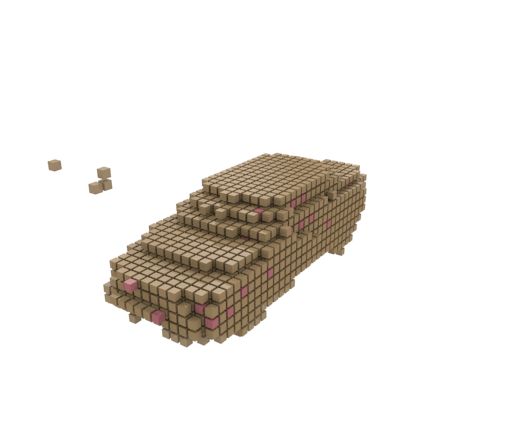}
        \end{subfigure}
        \begin{subfigure}[t]{0.095\textwidth}
        	\vspace{0px}\centering
        	\includegraphics[width=1.8cm,trim={\cropleft cm \croplower cm \cropright cm \cropupper cm},clip]{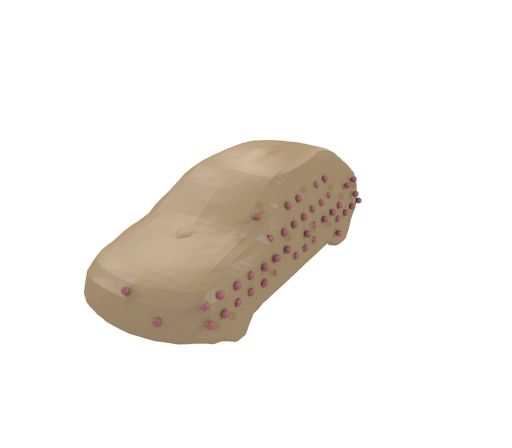}
        \end{subfigure}
        \begin{subfigure}[t]{0.095\textwidth}
            \vspace{0px}\centering
            \includegraphics[width=1.8cm,trim={\cropleft cm \croplower cm \cropright cm \cropupper cm},clip]{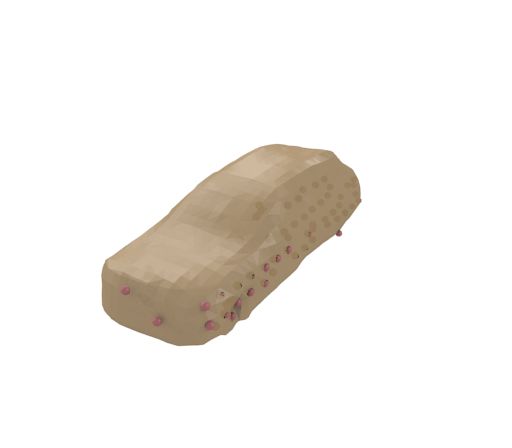}
        \end{subfigure}
        \begin{subfigure}[t]{0.095\textwidth}
        	\vspace{0px}\centering
        	\includegraphics[width=1.8cm,trim={\cropleft cm \croplower cm \cropright cm \cropupper cm},clip]{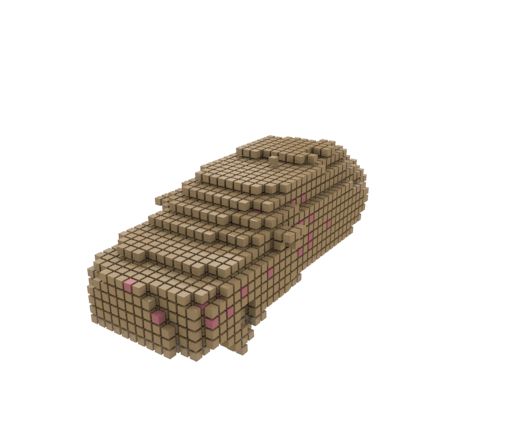}
        \end{subfigure}
        \begin{subfigure}[t]{0.095\textwidth}
        	\vspace{0px}\centering
        	\includegraphics[width=1.8cm,trim={\cropleft cm \croplower cm \cropright cm \cropupper cm},clip]{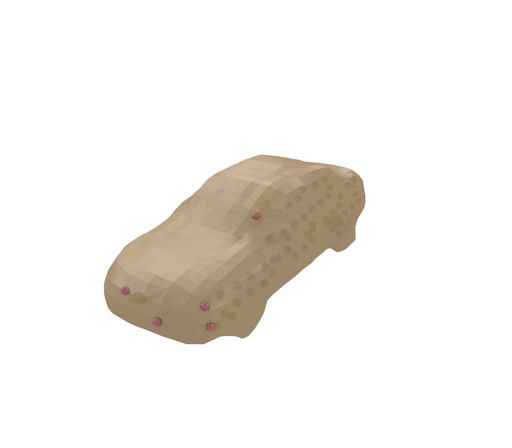}
        \end{subfigure}
        \begin{subfigure}[t]{0.095\textwidth}
        	\vspace{0px}\centering
        	\includegraphics[width=1.8cm,trim={\cropleft cm \croplower cm \cropright cm \cropupper cm},clip]{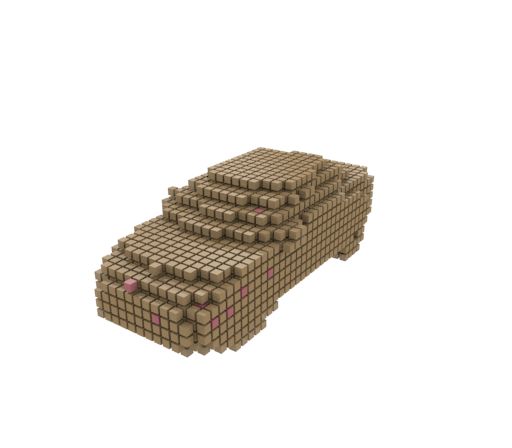}
        \end{subfigure}
        \begin{subfigure}[t]{0.095\textwidth}
        	\vspace{0px}\centering
        	\includegraphics[width=1.8cm,trim={\cropleft cm \croplower cm \cropright cm \cropupper cm},clip]{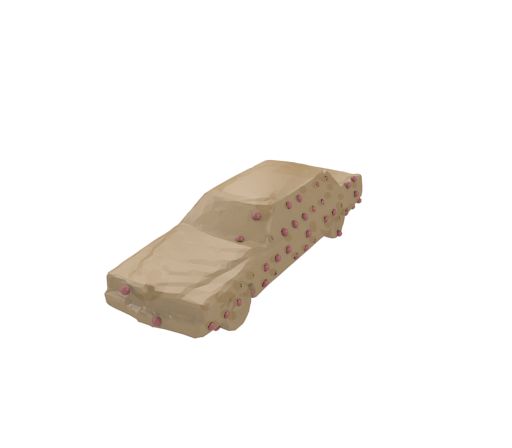}
        \end{subfigure}
        \begin{subfigure}[t]{0.095\textwidth}
        	\vspace{0px}\centering
        	\includegraphics[width=1.8cm,trim={\cropleft cm \croplower cm \cropright cm \cropupper cm},clip]{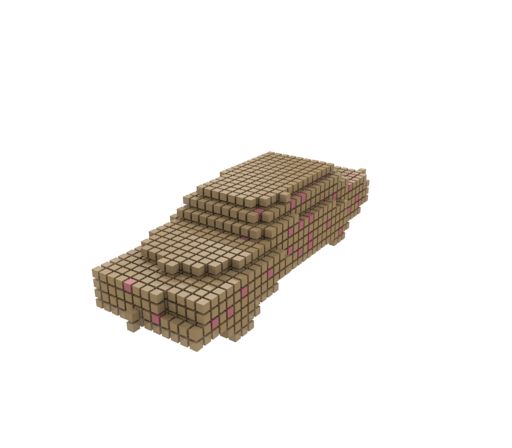}
        \end{subfigure}
        \subcaption{\clean (Top) and \noisy (Bottom), Low Resolution ($24\ntimes54\ntimes24$)}
    \end{subfigure}
    \\[4px]
    \begin{subfigure}[t]{1\textwidth}
   		\begin{subfigure}[t]{0.095\textwidth}
   			\vspace{0px}\centering
   			Obs\\
   			\includegraphics[width=1.8cm,trim={\cropleft cm \croplower cm \cropright cm \cropupper cm},clip]{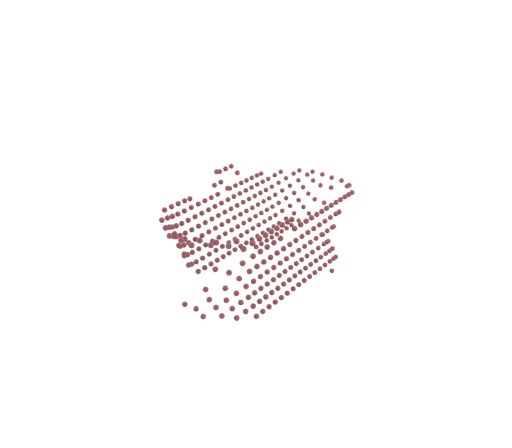}
   		\end{subfigure}
   		\begin{subfigure}[t]{0.095\textwidth}
   			\vspace{0px}\centering
   			\Dai\\
   			\includegraphics[width=1.8cm,trim={\cropleft cm \croplower cm \cropright cm \cropupper cm},clip]{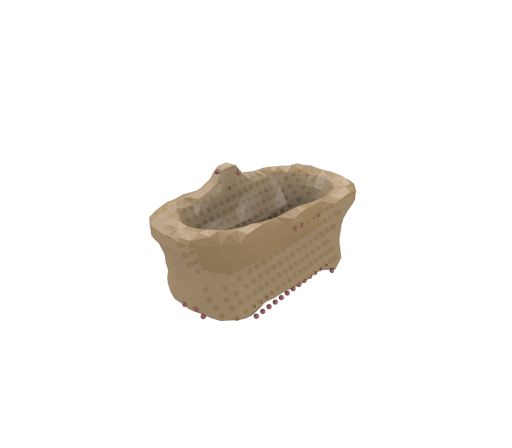}
   		\end{subfigure}
   		\begin{subfigure}[t]{0.095\textwidth}
   			\vspace{0px}\centering
   			\Dai\\
   			\includegraphics[width=1.8cm,trim={\cropleft cm \croplower cm \cropright cm \cropupper cm},clip]{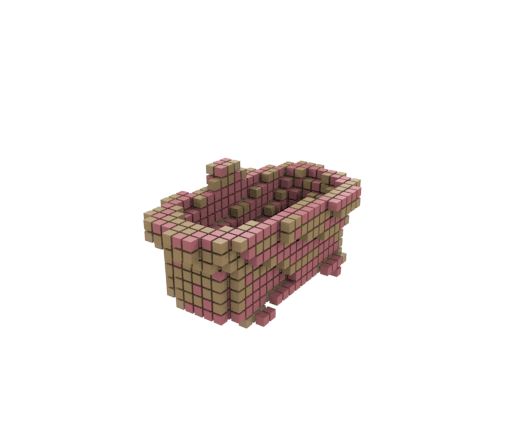}
   		\end{subfigure}
   		\begin{subfigure}[t]{0.095\textwidth}
   			\vspace{0px}\centering
   			\ICP\\
   			\includegraphics[width=1.8cm,trim={\cropleft cm \croplower cm \cropright cm \cropupper cm},clip]{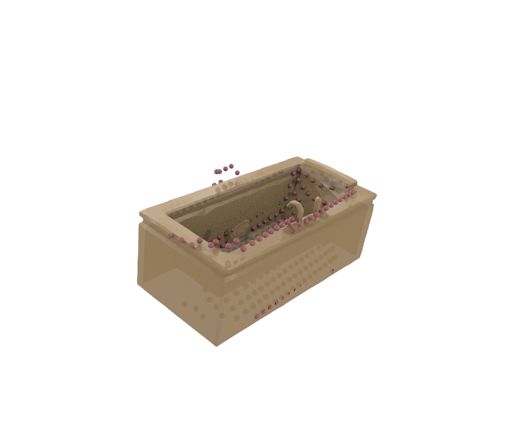}
   		\end{subfigure}
        \begin{subfigure}[t]{0.095\textwidth}
            \vspace{0px}\centering
            \ML\\
            \includegraphics[width=1.8cm,trim={\cropleft cm \croplower cm \cropright cm \cropupper cm},clip]{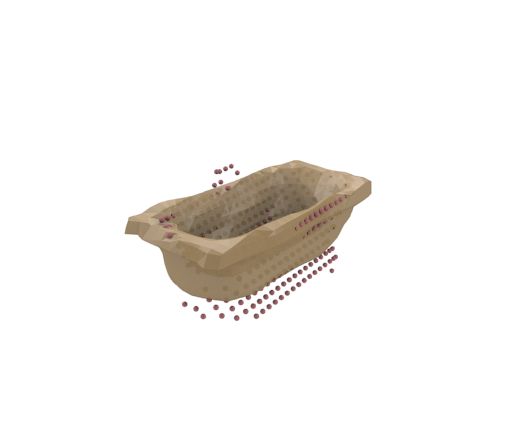}
        \end{subfigure}
   		\begin{subfigure}[t]{0.095\textwidth}
   			\vspace{0px}\centering
   			\ML\\
   			\includegraphics[width=1.8cm,trim={\cropleft cm \croplower cm \cropright cm \cropupper cm},clip]{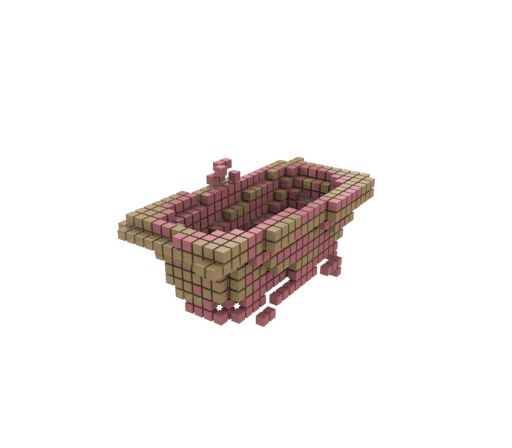}
   		\end{subfigure}
   		\begin{subfigure}[t]{0.095\textwidth}
   			\vspace{0px}\centering
   			\AML\\
   			\includegraphics[width=1.8cm,trim={\cropleft cm \croplower cm \cropright cm \cropupper cm},clip]{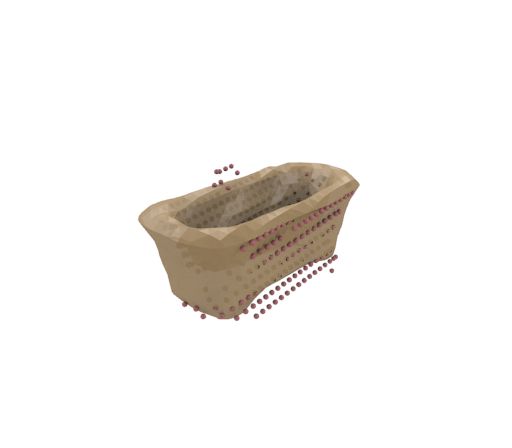}
   		\end{subfigure}
   		\begin{subfigure}[t]{0.095\textwidth}
   			\vspace{0px}\centering
   			\AML\\
   			\includegraphics[width=1.8cm,trim={\cropleft cm \croplower cm \cropright cm \cropupper cm},clip]{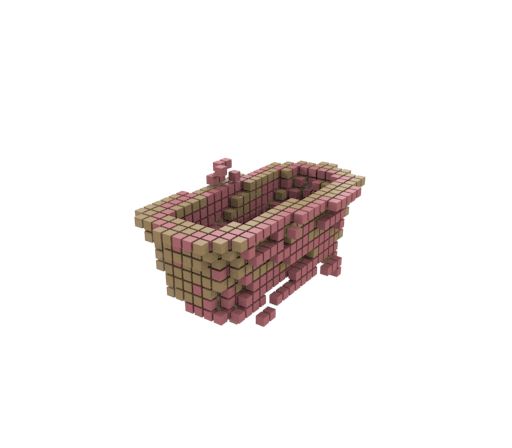}
   		\end{subfigure}
   		\begin{subfigure}[t]{0.095\textwidth}
   			\vspace{0px}\centering
   			GT\\
   			\includegraphics[width=1.8cm,trim={\cropleft cm \croplower cm \cropright cm \cropupper cm},clip]{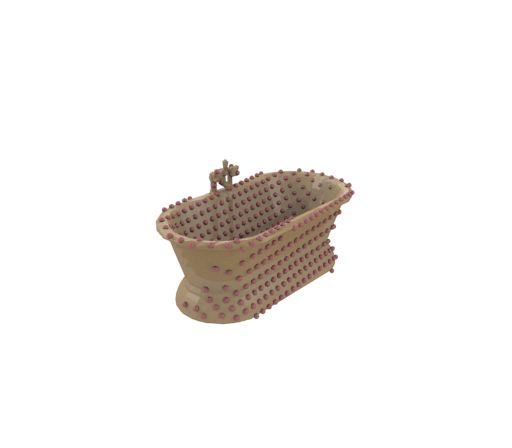}
   		\end{subfigure}
   		\begin{subfigure}[t]{0.095\textwidth}
   			\vspace{0px}\centering
   			GT\\
   			\includegraphics[width=1.8cm,trim={\cropleft cm \croplower cm \cropright cm \cropupper cm},clip]{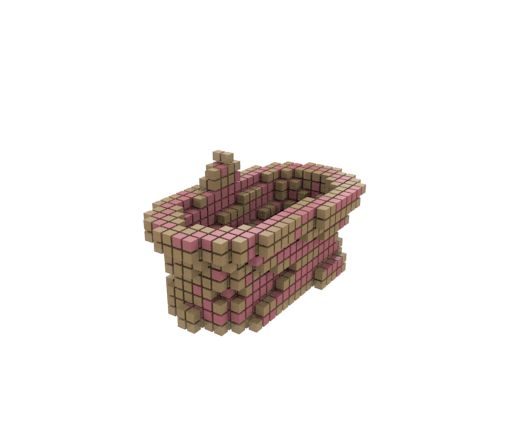}
   		\end{subfigure}
   		\\[2px]
   		\begin{subfigure}[t]{0.095\textwidth}
   			\vspace{0px}\centering
   			\includegraphics[width=1.8cm,trim={\cropleft cm \croplower cm \cropright cm \cropupper cm},clip]{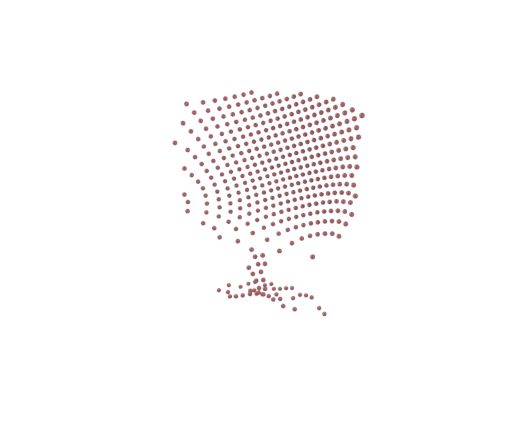}
   		\end{subfigure}
   		\begin{subfigure}[t]{0.095\textwidth}
   			\vspace{0px}\centering
   			\includegraphics[width=1.8cm,trim={\cropleft cm \croplower cm \cropright cm \cropupper cm},clip]{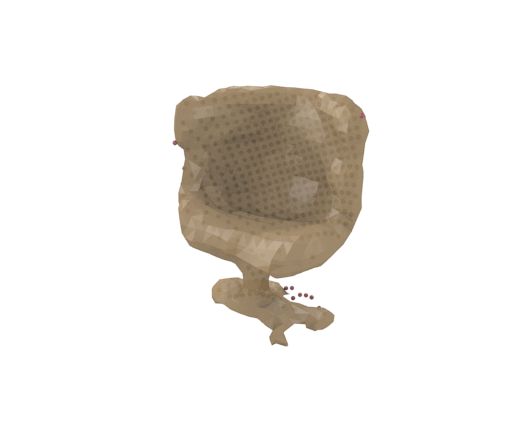}
   		\end{subfigure}
   		\begin{subfigure}[t]{0.095\textwidth}
   			\vspace{0px}\centering
   			\includegraphics[width=1.8cm,trim={\cropleft cm \croplower cm \cropright cm \cropupper cm},clip]{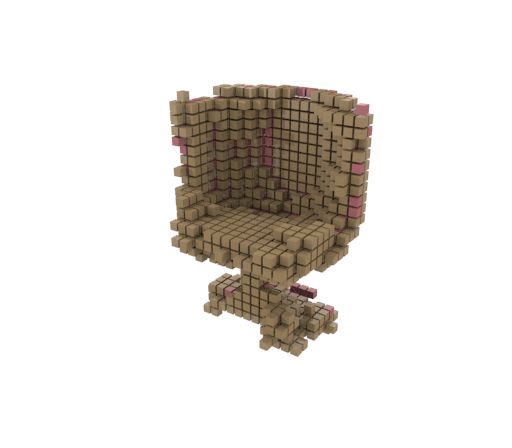}
   		\end{subfigure}
   		\begin{subfigure}[t]{0.095\textwidth}
   			\vspace{0px}\centering
   			\includegraphics[width=1.8cm,trim={\cropleft cm \croplower cm \cropright cm \cropupper cm},clip]{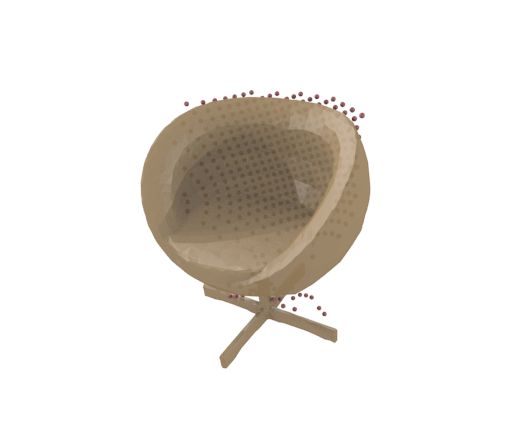}
   		\end{subfigure}
        \begin{subfigure}[t]{0.095\textwidth}
            \vspace{0px}\centering
            \includegraphics[width=1.8cm,trim={\cropleft cm \croplower cm \cropright cm \cropupper cm},clip]{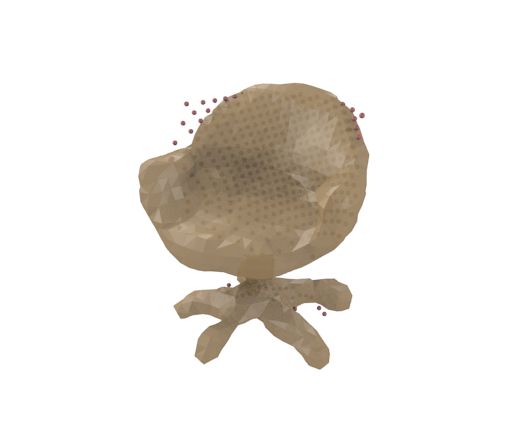}
        \end{subfigure}
   		\begin{subfigure}[t]{0.095\textwidth}
   			\vspace{0px}\centering
   			\includegraphics[width=1.8cm,trim={\cropleft cm \croplower cm \cropright cm \cropupper cm},clip]{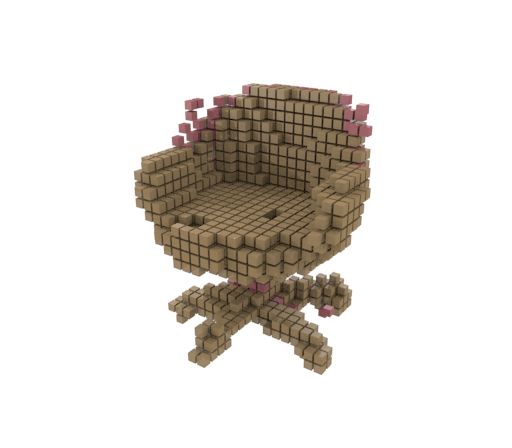}
   		\end{subfigure}
   		\begin{subfigure}[t]{0.095\textwidth}
   			\vspace{0px}\centering
   			\includegraphics[width=1.8cm,trim={\cropleft cm \croplower cm \cropright cm \cropupper cm},clip]{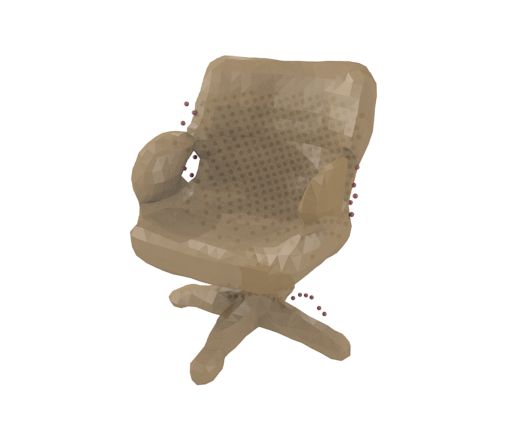}
   		\end{subfigure}
   		\begin{subfigure}[t]{0.095\textwidth}
   			\vspace{0px}\centering
   			\includegraphics[width=1.8cm,trim={\cropleft cm \croplower cm \cropright cm \cropupper cm},clip]{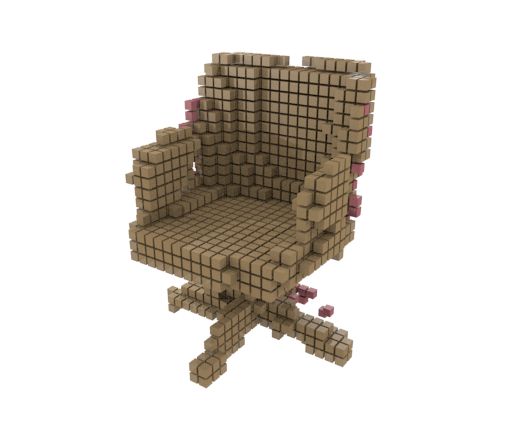}
   		\end{subfigure}
   		\begin{subfigure}[t]{0.095\textwidth}
   			\vspace{0px}\centering
   			\includegraphics[width=1.8cm,trim={\cropleft cm \croplower cm \cropright cm \cropupper cm},clip]{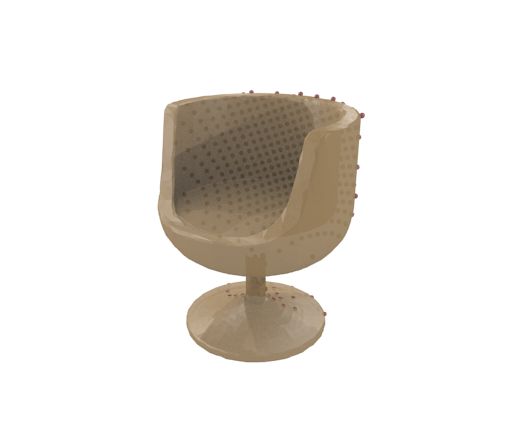}
   		\end{subfigure}
   		\begin{subfigure}[t]{0.095\textwidth}
   			\vspace{0px}\centering
   			\includegraphics[width=1.8cm,trim={\cropleft cm \croplower cm \cropright cm \cropupper cm},clip]{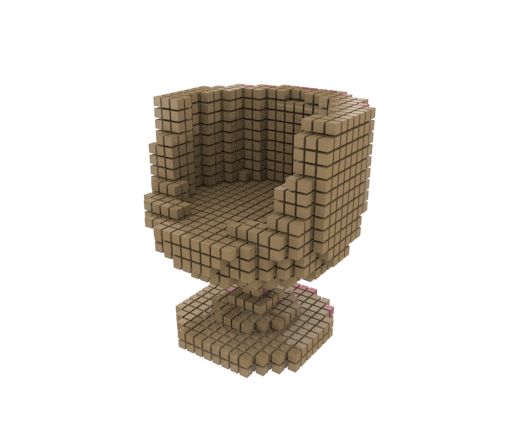}
   		\end{subfigure}
   		\\[2px]
   		\begin{subfigure}[t]{0.095\textwidth}
   			\vspace{0px}\centering
   			\includegraphics[width=1.8cm,trim={\cropleft cm \croplower cm \cropright cm \cropupper cm},clip]{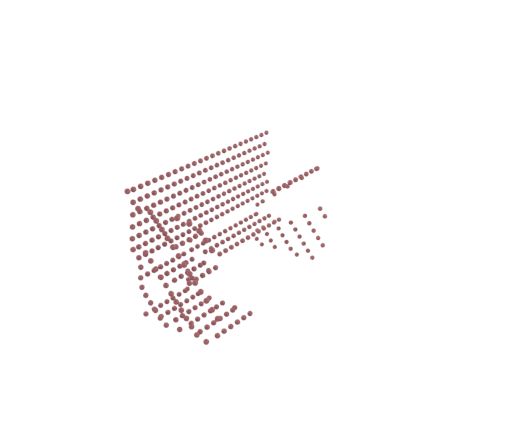}
   		\end{subfigure}
   		\begin{subfigure}[t]{0.095\textwidth}
   			\vspace{0px}\centering
   			\includegraphics[width=1.8cm,trim={\cropleft cm \croplower cm \cropright cm \cropupper cm},clip]{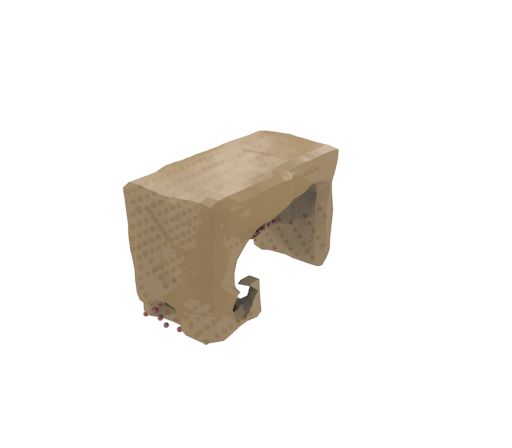}
   		\end{subfigure}
   		\begin{subfigure}[t]{0.095\textwidth}
   			\vspace{0px}\centering
   			\includegraphics[width=1.8cm,trim={\cropleft cm \croplower cm \cropright cm \cropupper cm},clip]{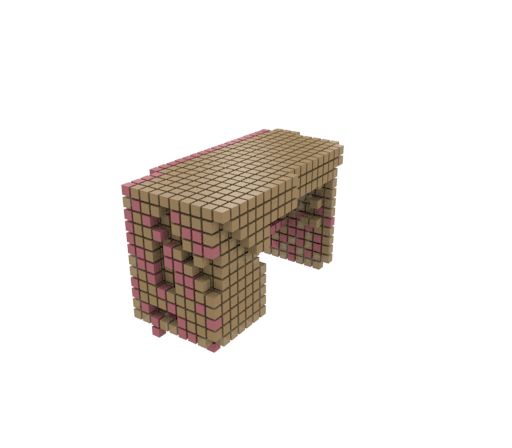}
   		\end{subfigure}
   		\begin{subfigure}[t]{0.095\textwidth}
   			\vspace{0px}\centering
   			\includegraphics[width=1.8cm,trim={\cropleft cm \croplower cm \cropright cm \cropupper cm},clip]{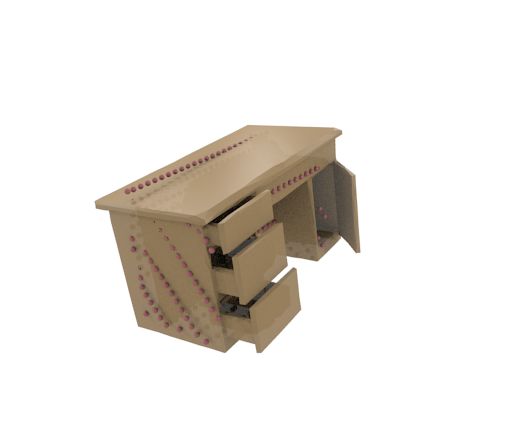}
   		\end{subfigure}
        \begin{subfigure}[t]{0.095\textwidth}
            \vspace{0px}\centering
            \includegraphics[width=1.8cm,trim={\cropleft cm \croplower cm \cropright cm \cropupper cm},clip]{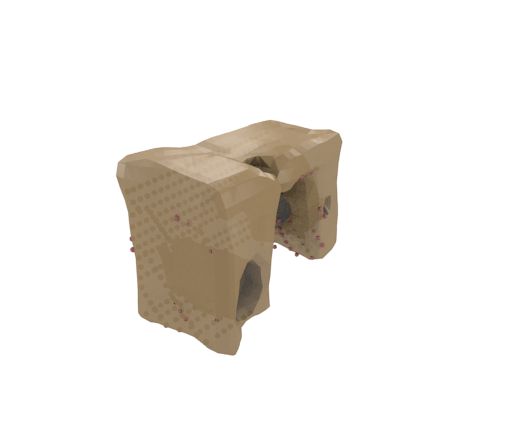}
        \end{subfigure}
   		\begin{subfigure}[t]{0.095\textwidth}
   			\vspace{0px}\centering
   			\includegraphics[width=1.8cm,trim={\cropleft cm \croplower cm \cropright cm \cropupper cm},clip]{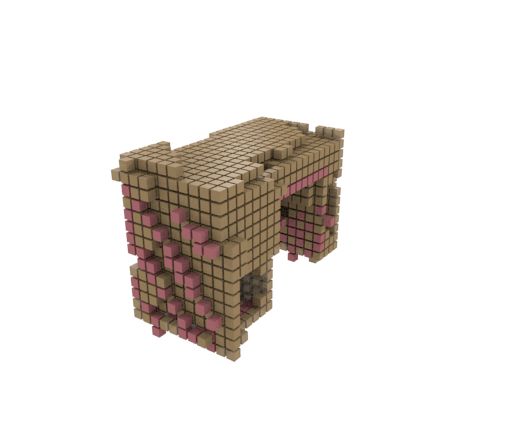}
   		\end{subfigure}
   		\begin{subfigure}[t]{0.095\textwidth}
   			\vspace{0px}\centering
   			\includegraphics[width=1.8cm,trim={\cropleft cm \croplower cm \cropright cm \cropupper cm},clip]{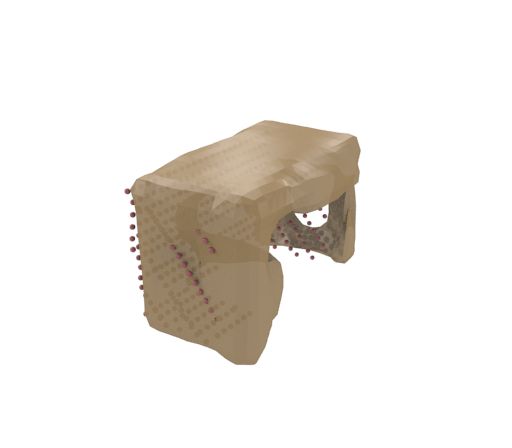}
   		\end{subfigure}
   		\begin{subfigure}[t]{0.095\textwidth}
   			\vspace{0px}\centering
   			\includegraphics[width=1.8cm,trim={\cropleft cm \croplower cm \cropright cm \cropupper cm},clip]{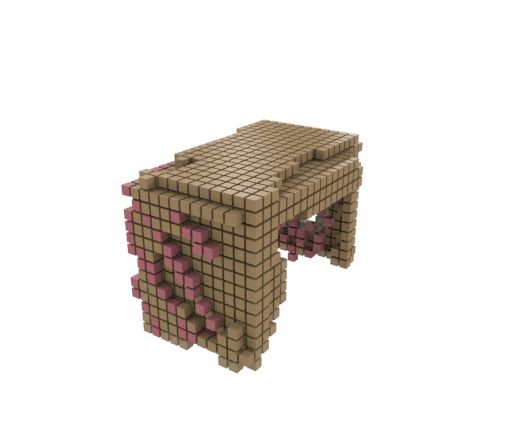}
   		\end{subfigure}
   		\begin{subfigure}[t]{0.095\textwidth}
   			\vspace{0px}\centering
   			\includegraphics[width=1.8cm,trim={\cropleft cm \croplower cm \cropright cm \cropupper cm},clip]{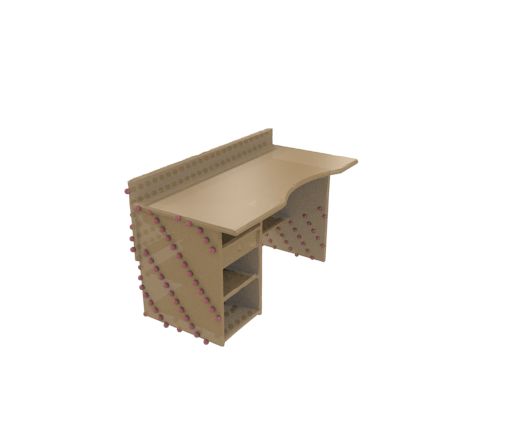}
   		\end{subfigure}
   		\begin{subfigure}[t]{0.095\textwidth}
   			\vspace{0px}\centering
   			\includegraphics[width=1.8cm,trim={\cropleft cm \croplower cm \cropright cm \cropupper cm},clip]{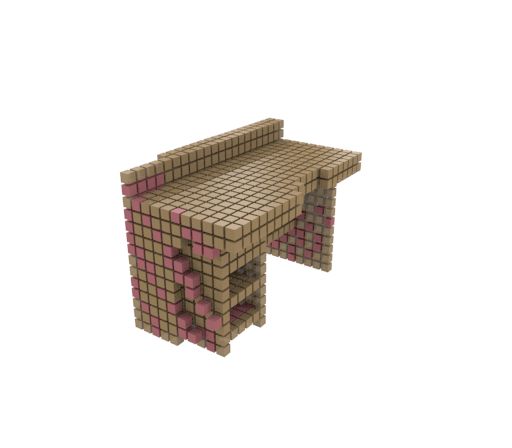}
   		\end{subfigure}
   		\\[2px]
   		\begin{subfigure}[t]{0.095\textwidth}
   			\vspace{0px}\centering
   			\includegraphics[width=1.8cm,trim={\cropleft cm \croplower cm \cropright cm \cropupper cm},clip]{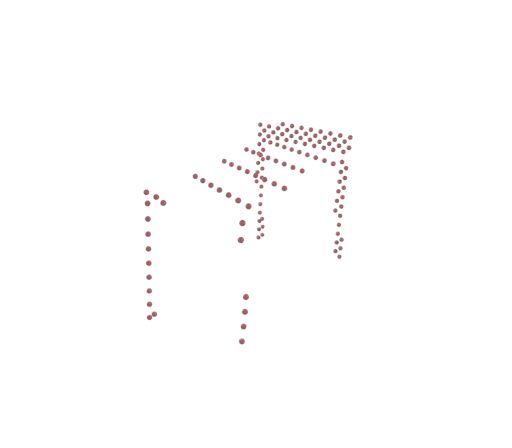}
   		\end{subfigure}
   		\begin{subfigure}[t]{0.095\textwidth}
   			\vspace{0px}\centering
   			\includegraphics[width=1.8cm,trim={\cropleft cm \croplower cm \cropright cm \cropupper cm},clip]{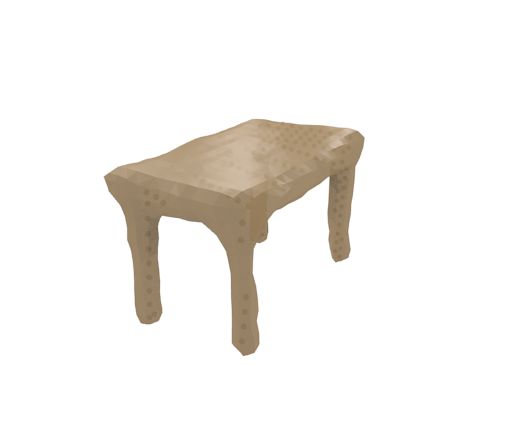}
   		\end{subfigure}
   		\begin{subfigure}[t]{0.095\textwidth}
   			\vspace{0px}\centering
   			\includegraphics[width=1.8cm,trim={\cropleft cm \croplower cm \cropright cm \cropupper cm},clip]{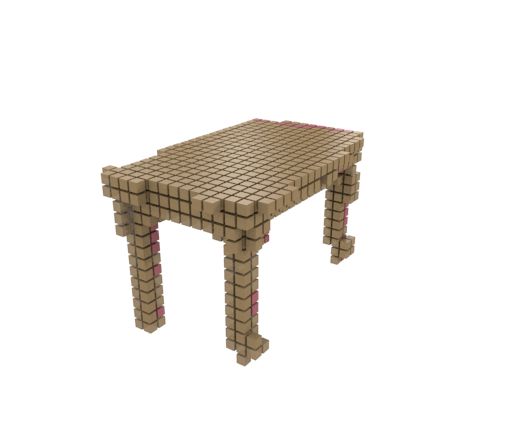}
   		\end{subfigure}
   		\begin{subfigure}[t]{0.095\textwidth}
   			\vspace{0px}\centering
   			\includegraphics[width=1.8cm,trim={\cropleft cm \croplower cm \cropright cm \cropupper cm},clip]{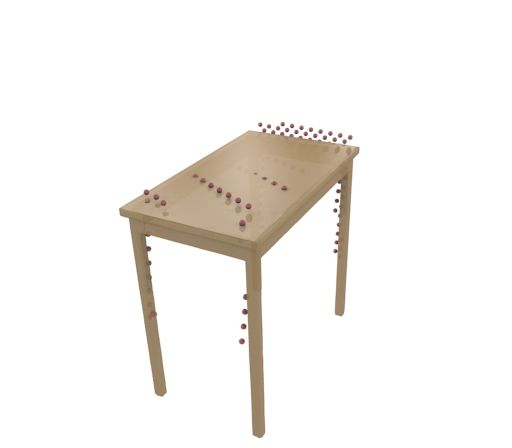}
   		\end{subfigure}
        \begin{subfigure}[t]{0.095\textwidth}
            \vspace{0px}\centering
            \includegraphics[width=1.8cm,trim={\cropleft cm \croplower cm \cropright cm \cropupper cm},clip]{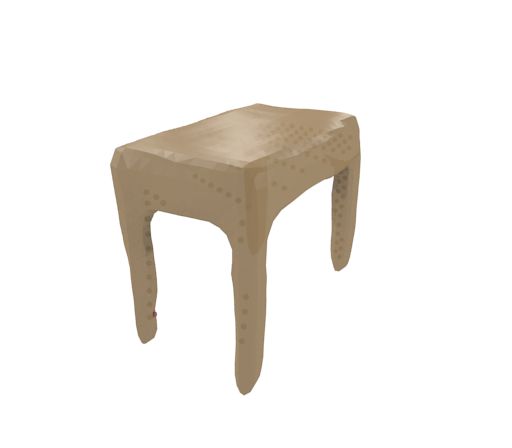}
        \end{subfigure}
   		\begin{subfigure}[t]{0.095\textwidth}
   			\vspace{0px}\centering
   			\includegraphics[width=1.8cm,trim={\cropleft cm \croplower cm \cropright cm \cropupper cm},clip]{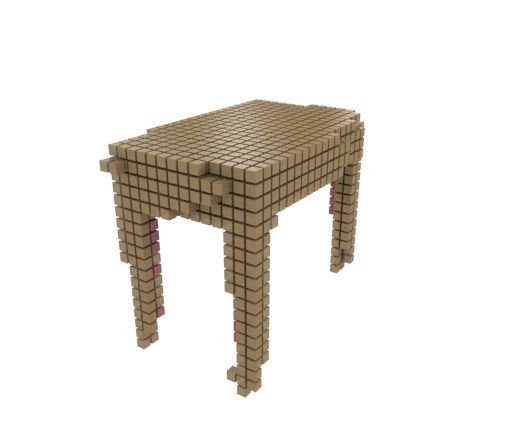}
   		\end{subfigure}
   		\begin{subfigure}[t]{0.095\textwidth}
   			\vspace{0px}\centering
   			\includegraphics[width=1.8cm,trim={\cropleft cm \croplower cm \cropright cm \cropupper cm},clip]{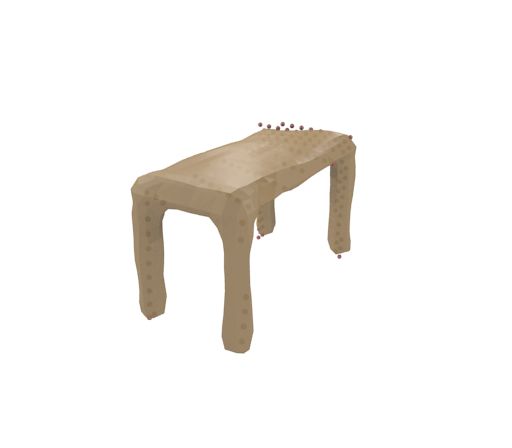}
   		\end{subfigure}
   		\begin{subfigure}[t]{0.095\textwidth}
   			\vspace{0px}\centering
   			\includegraphics[width=1.8cm,trim={\cropleft cm \croplower cm \cropright cm \cropupper cm},clip]{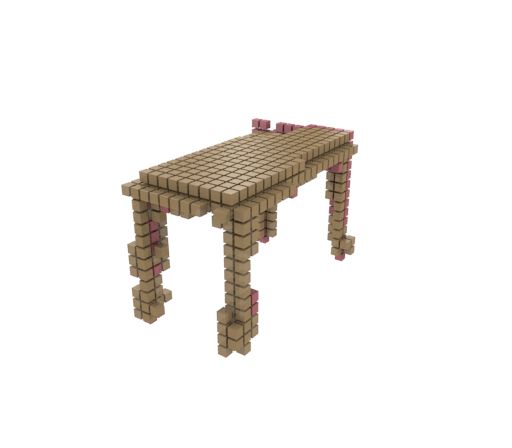}
   		\end{subfigure}
   		\begin{subfigure}[t]{0.095\textwidth}
   			\vspace{0px}\centering
   			\includegraphics[width=1.8cm,trim={\cropleft cm \croplower cm \cropright cm \cropupper cm},clip]{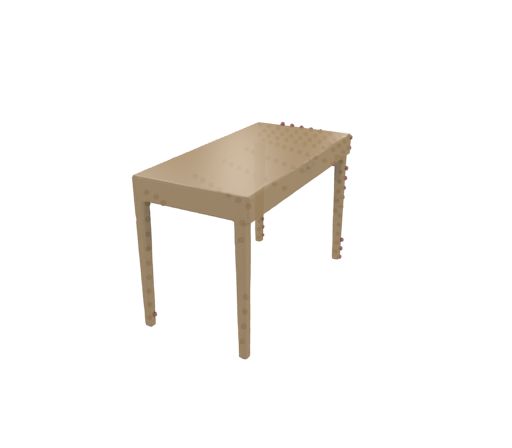}
   		\end{subfigure}
   		\begin{subfigure}[t]{0.095\textwidth}
   			\vspace{0px}\centering
   			\includegraphics[width=1.8cm,trim={\cropleft cm \croplower cm \cropright cm \cropupper cm},clip]{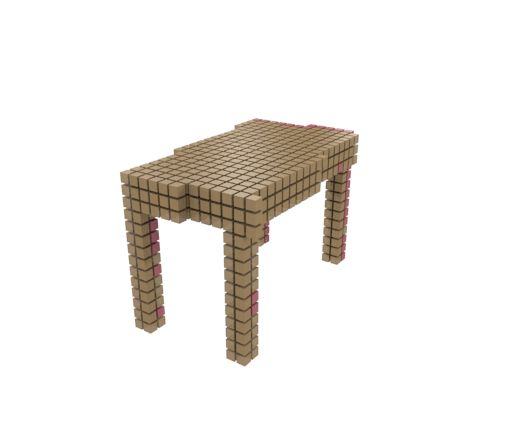}
   		\end{subfigure}
       \subcaption{ModelNet Bathtubs, Chairs, Desks and Tables, Low Resolution ($32^3$)}
   	\end{subfigure}
    }
    \vspace*{-\figskipcaption px}
    \caption{{\bf Qualitative Results on ShapeNet and ModelNet.} Results for \AML, \Dai, \Engelmann, \ICP and \ML on \clean, \noisy and ModelNet's bathtubs, chairs, desks and tables. \AML outperforms data-driven approaches (\ML, \Engelmann, \ICP) and rivals \Dai while requiring significantly less supervision. Occupancy grids and meshes in {\color{rbeige}beige}, observations in {\color{rred}red}.}
    \label{fig:results-synthetic}
    \vspace*{-\figskipbelow px}
\end{figure*}
\begin{table*}[t]
    \vspace*{-\figskipabove px}
    \centering
    {\scriptsize
        
    \begin{tabularx}{1\linewidth}{|@{  }l@{  }|@{  }X@{  }|@{  }c@{  }c@{  }|@{  }c@{  }c@{  }c@{  }c@{  }|@{  }c@{  }c@{  }|@{  }c@{  }c@{  }|c@{  }c@{  }|}
        \hline
        Supervision & Method & \multicolumn{2}{@{  }c@{  }|@{  }}{bathtub} & \multicolumn{4}{@{  }c@{  }|@{  }}{chair} & \multicolumn{2}{@{  }c@{  }|@{  }}{desk} & \multicolumn{2}{@{  }c@{  }|}{table} & \multicolumn{2}{@{  }c@{  }|}{ModelNet10}\\
        \multicolumn{1}{|@{  }c@{  }|@{  }}{in \%} && \Abs{\tiny $\downarrow$}& \IoU{\tiny $\uparrow$} & \Abs{\tiny $\downarrow$} & \IoU{\tiny $\uparrow$}  & \Acc [vx]{\tiny $\downarrow$} & \Compl [vx]{\tiny $\downarrow$} & \Abs{\tiny $\downarrow$}& \IoU{\tiny $\uparrow$} & \Abs{\tiny $\downarrow$} & \IoU{\tiny $\uparrow$} & \Abs{\tiny $\downarrow$} & \IoU{\tiny $\uparrow$}\\
        \hline\hline
        \multicolumn{14}{|c|}{Low Resolution: $32^3$ voxels; * independent of resolution}\\
        \hline\hline
        \color{darkgray}(shape prior) & \leavevmode\color{darkgray}\DVAE & \color{darkgray}0.015 & \color{darkgray}0.699 & \color{darkgray}0.025 & \color{darkgray}0.517 & \color{darkgray}0.884 & \color{darkgray}0.72 & \color{darkgray}0.028 & \color{darkgray}0.555 & \color{darkgray}011 & \color{darkgray}0.608 & \color{darkgray}0.023 & \color{darkgray}0.714\\
        \hline\hline
        \multirow{2}{*}{$\hphantom{<}100$} & \cite{Dai2017CVPRa} (\Dai) & \bf\color{rred}0.022 & 0.59 & \bf\color{rred}0.019 & \bf\color{rred}0.61 & \bf\color{rred}0.663 & \bf\color{rred}0.671 & \bf\color{rred}0.027 & \bf\color{rred}0.568 & \bf\color{rred}0.011 & \bf\color{rred}0.648 & \bf\color{rred}0.03 & \bf\color{rred}0.646\\
        & \Sup & 0.023 & \bf\color{rred}0.618 & 0.03 & 0.478 & 0.873 & 0.813 & 0.036 & 0.458 & 0.017 & 0.497 & 0.038 & 0.589\\
        \hline
        \multirow{3}{*}{$<10$} & * \cite{Gupta2015CVPR} (\ICP) & \multicolumn{2}{@{  }c@{  }|@{  }}{\color{darkgray}(mesh only)} & \multicolumn{2}{@{  }c@{  }}{\color{darkgray}(mesh only)} & 1.483 & 0.89 & \multicolumn{2}{@{  }c@{  }|@{  }}{\color{darkgray}(mesh only)} & \multicolumn{2}{@{  }c@{  }|}{\color{darkgray}(mesh only)} & \multicolumn{2}{@{  }c@{  }|}{\color{darkgray}(mesh only)}\\
        & \ML & 0.028 & \bf\color{rgreen}0.503 & \bf\color{rgreen}0.033 & \bf\color{rgreen}0.414 & 1.489 & 1.065 & 0.048 & 0.323 & 0.029 & 0.318 &\multicolumn{2}{@{  }c@{  }|}{\color{darkgray} (too slow)}\\
        & \AML & \bf\color{rgreen}0.026 & \bf\color{rgreen}0.503 & \bf\color{rgreen}0.033 & 0.373 & \bf\color{rgreen}1.088 & \bf\color{rgreen}0.785 & \bf\color{rgreen}0.041 & \bf\color{rgreen}0.389 & \bf\color{rgreen}0.018 & \bf\color{rgreen}0.423 & \bf\color{rgreen}0.04 & \bf\color{rgreen}0.509\\
        \cline{13-14}
        \hhline{============~~}
        \multicolumn{12}{|@{  }c@{  }|}{Medium Resolution: $48^3$ voxels}&\multicolumn{2}{c}{\multirow{1}{*}{}}\\
        \hhline{============~~}
        \color{darkgray}(shape prior) & \leavevmode\color{darkgray}\DVAE & \color{darkgray}0.014 & \color{darkgray}0.671 & \color{darkgray}0.021 & \color{darkgray}0.491 & \color{darkgray}0.748 & \color{darkgray}0.697 & \color{darkgray}0.025 & \color{darkgray}0.525 & \color{darkgray}0.01 & \color{darkgray}0.548&\multicolumn{2}{c}{\multirow{1}{*}{}}\\
        \hhline{============~~}
        $\hphantom{<}100$ & \cite{Dai2017CVPRa} (\Dai) & \bf\color{rred}0.018 & \bf\color{rred}0.609 & \bf\color{rred}0.016 & \bf\color{rred}0.576 & \bf\color{rred}0.513 & \bf\color{rred}0.508 & \bf\color{rred}0.023 & \bf\color{rred}0.532 & \bf\color{rred}0.008 & \bf\color{rred}0.65&\multicolumn{2}{c}{\multirow{1}{*}{}}\\
        \cline{1-12}
        $<9$ & \AML & \bf\color{rgreen}0.024 & \bf\color{rgreen}0.459 & \bf\color{rgreen}0.029 & \bf\color{rgreen}0.347 & \bf\color{rgreen}1.025 & \bf\color{rgreen}0.805& \bf\color{rgreen}0.034 & \bf\color{rgreen}0.361 & \bf\color{rgreen}0.015 & \bf\color{rgreen}0.384&\multicolumn{2}{c}{\multirow{1}{*}{}}\\
        \cline{1-12}\cline{1-12}
        \multicolumn{12}{|@{  }c@{  }|}{High Resolution: $64^3$ voxels}&\multicolumn{2}{c}{\multirow{1}{*}{}}\\
        \hhline{============~~}
        \color{darkgray}(shape prior) & \leavevmode\color{darkgray}\DVAE & \color{darkgray}0.014 & \color{darkgray}0.644 & \color{darkgray}0.02 & \color{darkgray}0.474 & \color{darkgray}0.702 & \color{darkgray}0.705 & \color{darkgray}0.024 & \color{darkgray}0.506 & \color{darkgray}0.009 & \color{darkgray}0.548&\multicolumn{2}{c}{\multirow{1}{*}{}}\\
        \cline{1-12}\cline{1-12}
        $\hphantom{<}100$ & \cite{Dai2017CVPRa} (\Dai) & \bf\color{rred}0.018 & \bf\color{rred}0.54 & \bf\color{rred}0.016 & \bf\color{rred}0.548 & \bf\color{rred}0.47 & \bf\color{rred}0.53 & \bf\color{rred}0.021 & \bf\color{rred}0.525 & \bf\color{rred}0.007 & \bf\color{rred}0.673&\multicolumn{2}{c}{\multirow{1}{*}{}}\\
        \cline{1-12}
        $<9$ & \AML & \bf\color{rgreen}0.023 & \bf\color{rgreen}0.46 & \bf\color{rgreen}0.026 & \bf\color{rgreen}0.333 & \bf\color{rgreen}0.893 & \bf\color{rgreen}0.852 & \bf\color{rgreen}0.042 & \bf\color{rgreen}0.31 & \bf\color{rgreen}0.012 & \bf\color{rgreen}0.407&\multicolumn{2}{c}{\multirow{1}{*}{}}\\
        \cline{1-12}
    \end{tabularx}
    }
    \vspace*{-\figskipcaption px}
    \caption{{\bf Quantitative Results on ModelNet.} Results for bathtubs, chairs, desks, tables and all ten categories combined (ModelNet10). As the ground truth SDFs are merely approximations (\cf \secref{sec:data}), we concentrate on Hamming distance (\Abs; lower is better) and intersection-over-union (\IoU; higher is better). Only for chairs, we report accuracy \Acc and completeness \Compl in voxels (voxel length at $32^3$ voxels). We also indicate the level of supervision \red{(see \tabref{tab:data})}. Again, we report the \DVAE shape prior as reference and color the best weakly-supervised approach using {\bf\color{rgreen}green} and the best fully-supervised approach in {\bf\color{rred}red}.    }
    \label{tab:results-modelnet}
    \vspace*{-\figskipbelow px}
\end{table*}
\begin{figure}[t]
    \vspace*{-\figskipabove px}
    \vspace*{2px}
    \centering
    {\scriptsize
        
    \renewcommand{\cleana}{858}
    \newcommand{\cleanak}{2}
    
    \renewcommand{\cleanb}{330}
    \newcommand{\cleanbk}{5}
    
    \newcommand{\cleanc}{165}
    \newcommand{\cleanck}{3}
    
    \renewcommand{\noisya}{198} 
    \newcommand{\noisyak}{2}
    
    \renewcommand{\noisyb}{759} 
    \newcommand{\noisybk}{5}
    
    \newcommand{\noisyc}{165} 
    \newcommand{\noisyck}{3}
    
    \begin{subfigure}[t]{0.5\textwidth}
        \vspace{0px}\centering
   		\begin{subfigure}[t]{0.15\textwidth}
   			\vspace{0px}\centering
   			$k = 3$\\
   			\includegraphics[width=1.5cm,trim={\cropleft cm \croplower cm \cropright cm \cropupper cm},clip]{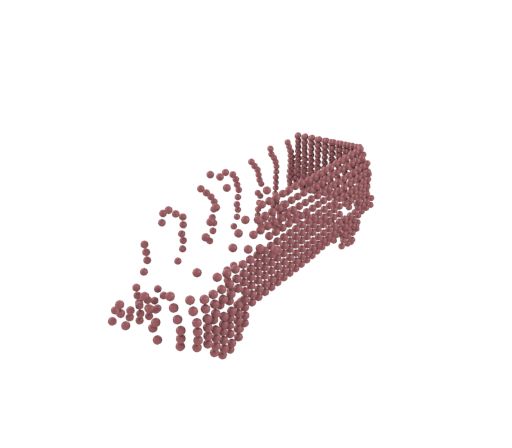}
   		\end{subfigure}
   		\begin{subfigure}[t]{0.15\textwidth}
   			\vspace{0px}\centering
   			\AML\\
   			\includegraphics[width=1.5cm,trim={\cropleft cm \croplower cm \cropright cm \cropupper cm},clip]{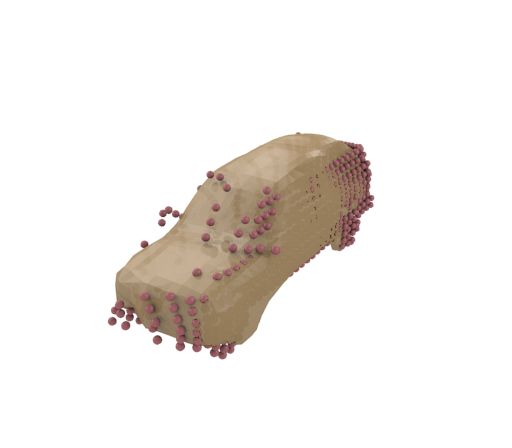}
   		\end{subfigure}
   		\begin{subfigure}[t]{0.15\textwidth}
   			\vspace{0px}\centering
   			GT\\
   			\includegraphics[width=1.5cm,trim={\cropleft cm \croplower cm \cropright cm \cropupper cm},clip]{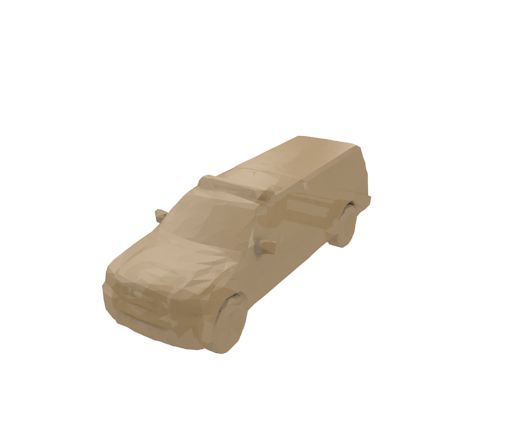}
   		\end{subfigure}
   		\begin{subfigure}[t]{0.15\textwidth}
   			\vspace{0px}\centering
   			$k = 5$\\
   			\includegraphics[width=1.5cm,trim={\cropleft cm \croplower cm \cropright cm \cropupper cm},clip]{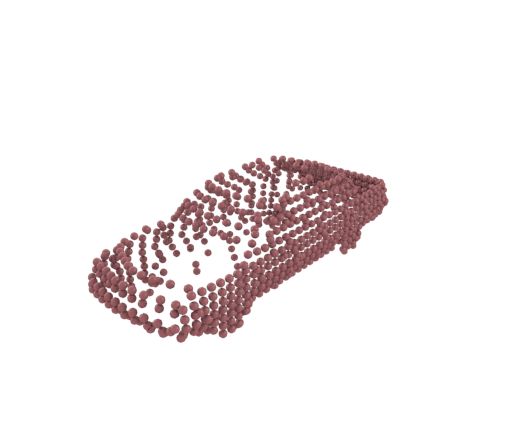}
   		\end{subfigure}
   		\begin{subfigure}[t]{0.15\textwidth}
   			\vspace{0px}\centering
   			\AML\\
   			\includegraphics[width=1.5cm,trim={\cropleft cm \croplower cm \cropright cm \cropupper cm},clip]{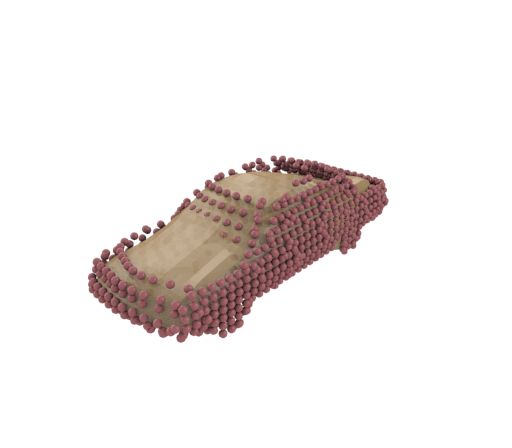}
   		\end{subfigure}
   		\begin{subfigure}[t]{0.15\textwidth}
   			\vspace{0px}\centering
   			GT\\
   			\includegraphics[width=1.5cm,trim={\cropleft cm \croplower cm \cropright cm \cropupper cm},clip]{gdat_shapenet_clean_low_\cleanb_gt_only}
   		\end{subfigure}
   		\\[-4px]
   		\begin{subfigure}[t]{0.15\textwidth}
   			\vspace{0px}\centering
   			\includegraphics[width=1.5cm,trim={\cropleft cm \croplower cm \cropright cm \cropupper cm},clip]{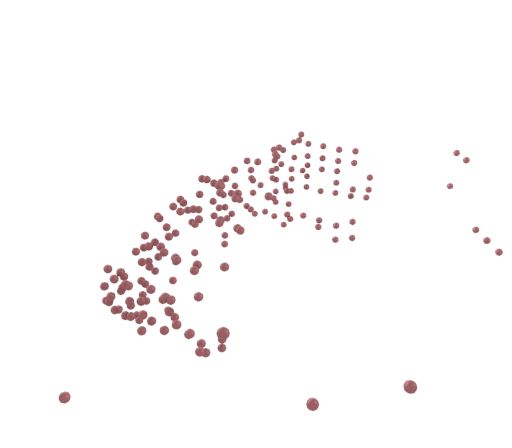}
   		\end{subfigure}
   		\begin{subfigure}[t]{0.15\textwidth}
   			\vspace{0px}\centering
   			\includegraphics[width=1.5cm,trim={\cropleft cm \croplower cm \cropright cm \cropupper cm},clip]{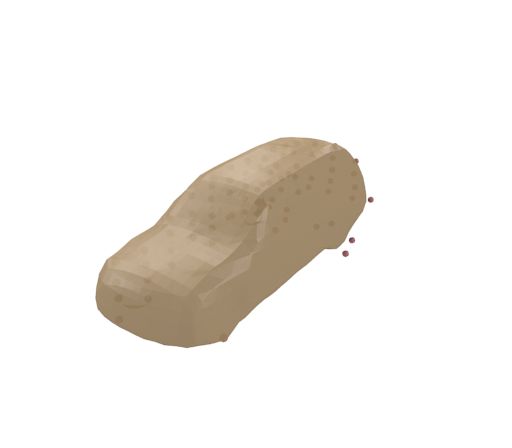}
   		\end{subfigure}
   		\begin{subfigure}[t]{0.15\textwidth}
   			\vspace{0px}\centering
   			\includegraphics[width=1.5cm,trim={\cropleft cm \croplower cm \cropright cm \cropupper cm},clip]{gdat_shapenet_clean_low_\noisyc_gt_only}
   		\end{subfigure}
   		\begin{subfigure}[t]{0.15\textwidth}
   			\vspace{0px}\centering
   			\includegraphics[width=1.5cm,trim={\cropleft cm \croplower cm \cropright cm \cropupper cm},clip]{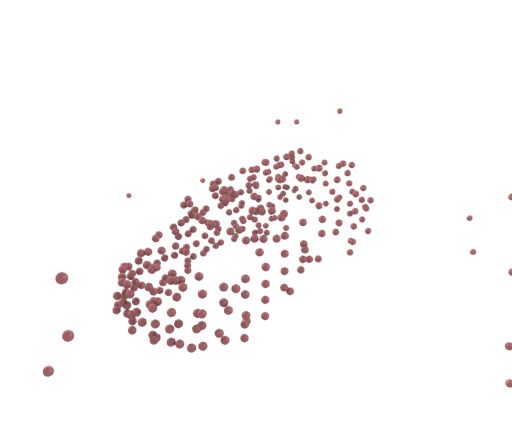}
   		\end{subfigure}
   		\begin{subfigure}[t]{0.15\textwidth}
   			\vspace{0px}\centering
   			\includegraphics[width=1.5cm,trim={\cropleft cm \croplower cm \cropright cm \cropupper cm},clip]{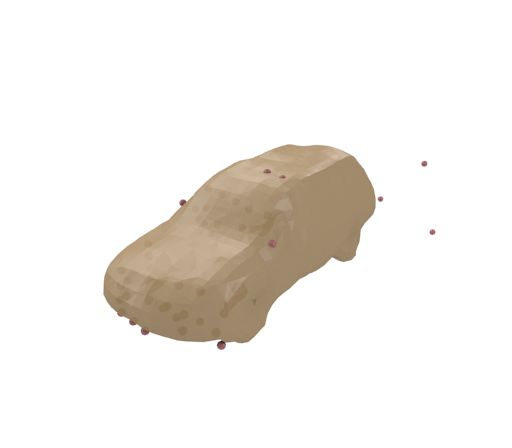}
   		\end{subfigure}
   		\begin{subfigure}[t]{0.15\textwidth}
   			\vspace{0px}\centering
   			\includegraphics[width=1.5cm,trim={\cropleft cm \croplower cm \cropright cm \cropupper cm},clip]{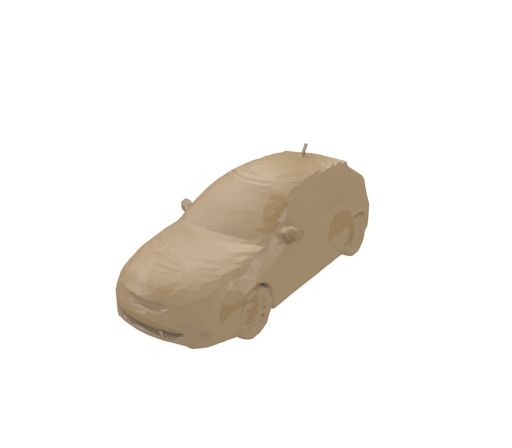}
   		\end{subfigure}
        \subcaption{\clean and -noisy, $k$ Views, Low Resolution ($24\ntimes54\ntimes24$)}
    \end{subfigure}
    \\[4px]
    \renewcommand{\cleana}{297} 
    \renewcommand{\cleanb}{0} 
    \renewcommand{\noisya}{264} %
    \renewcommand{\noisyb}{99} 
    \begin{subfigure}[t]{0.5\textwidth}
        \vspace{0px}\centering
        \begin{subfigure}[t]{0.15\textwidth}
            \vspace{0px}\centering
            \Dai\\
            \includegraphics[width=1.5cm,trim={\cropleft cm \croplower cm \cropright cm \cropupper cm},clip]{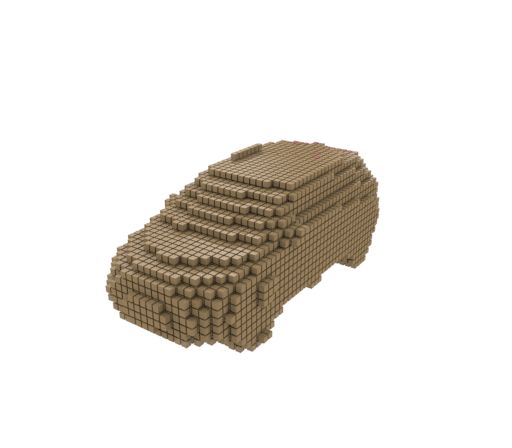}
        \end{subfigure}
        \begin{subfigure}[t]{0.15\textwidth}
        	\vspace{0px}\centering
        	\AML\\
        	\includegraphics[width=1.5cm,trim={\cropleft cm \croplower cm \cropright cm \cropupper cm},clip]{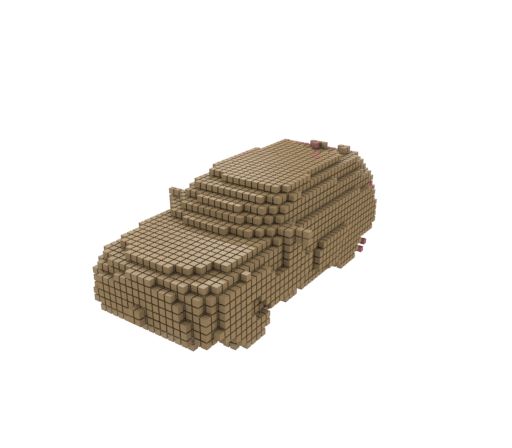}
        \end{subfigure}
        \begin{subfigure}[t]{0.15\textwidth}
            \vspace{0px}\centering
            GT\\
            \includegraphics[width=1.5cm,trim={\cropleft cm \croplower cm \cropright cm \cropupper cm},clip]{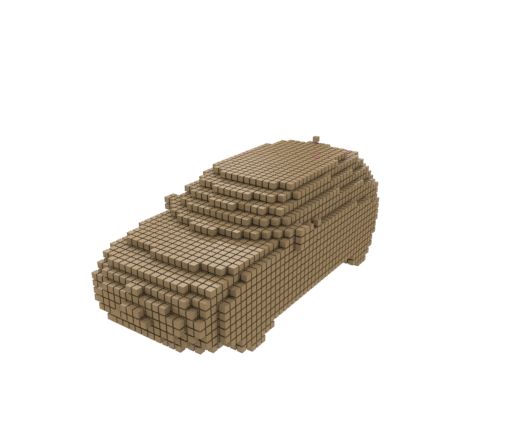}
        \end{subfigure}
        \begin{subfigure}[t]{0.15\textwidth}
            \vspace{0px}\centering
            \Dai\\
            \includegraphics[width=1.5cm,trim={\cropleft cm \croplower cm \cropright cm \cropupper cm},clip]{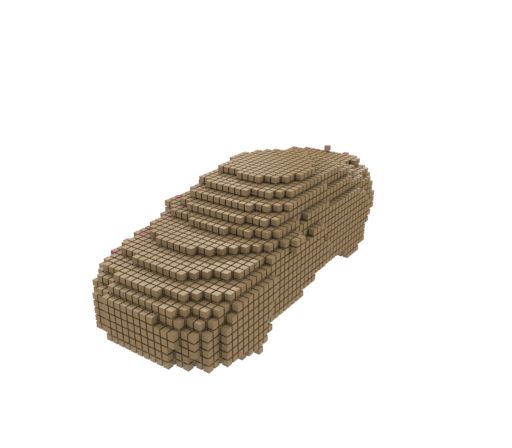}
        \end{subfigure}
        \begin{subfigure}[t]{0.15\textwidth}
            \vspace{0px}\centering
            \AML\\
            \includegraphics[width=1.5cm,trim={\cropleft cm \croplower cm \cropright cm \cropupper cm},clip]{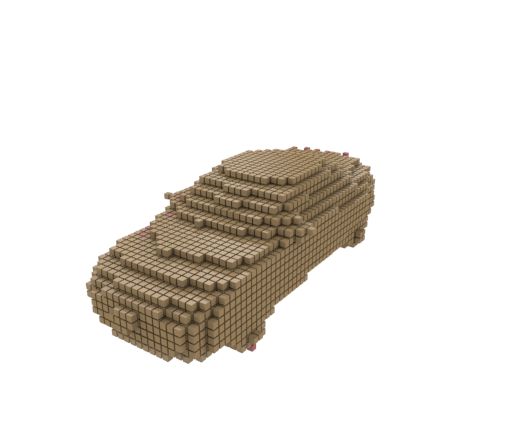}
        \end{subfigure}
        \begin{subfigure}[t]{0.15\textwidth}
            \vspace{0px}\centering
            GT\\
            \includegraphics[width=1.5cm,trim={\cropleft cm \croplower cm \cropright cm \cropupper cm},clip]{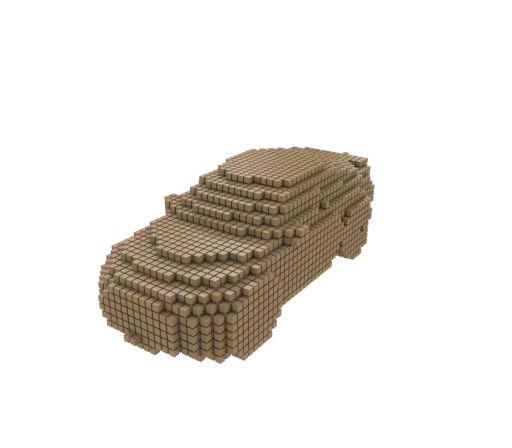}
        \end{subfigure}
        \\[2px]
        \hspace*{-6px}
        \begin{subfigure}[t]{0.15\textwidth}
            \vspace{0px}\centering
            \includegraphics[width=1.5cm,trim={\cropleft cm \croplower cm \cropright cm \cropupper cm},clip]{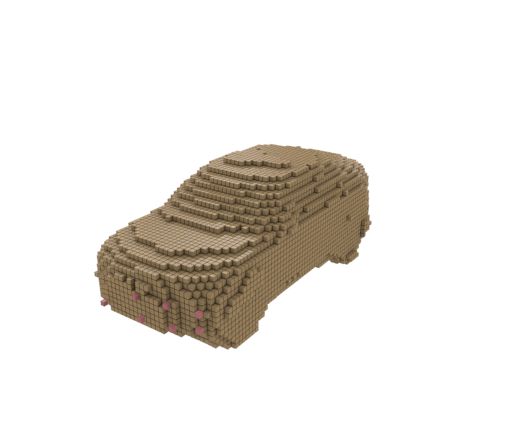}
        \end{subfigure}
        \begin{subfigure}[t]{0.15\textwidth}
        	\vspace{0px}\centering
        	\includegraphics[width=1.5cm,trim={\cropleft cm \croplower cm \cropright cm \cropupper cm},clip]{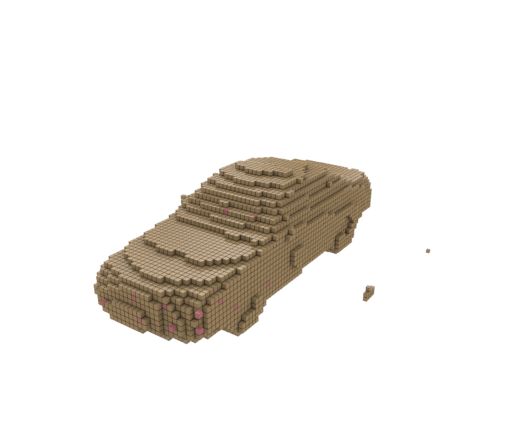}
        \end{subfigure}
        \begin{subfigure}[t]{0.15\textwidth}
            \vspace{0px}\centering
            \includegraphics[width=1.5cm,trim={\cropleft cm \croplower cm \cropright cm \cropupper cm},clip]{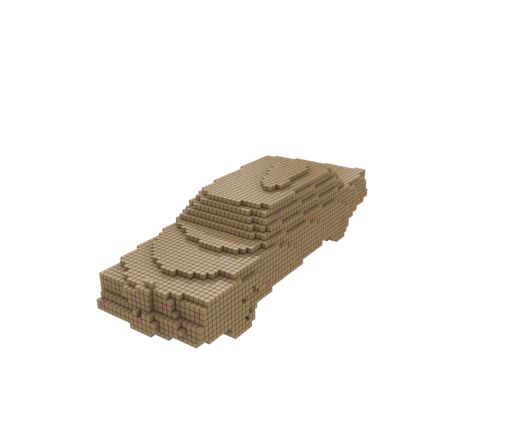}
        \end{subfigure}
        \begin{subfigure}[t]{0.15\textwidth}
            \vspace{0px}\centering
            \includegraphics[width=1.5cm,trim={\cropleft cm \croplower cm \cropright cm \cropupper cm},clip]{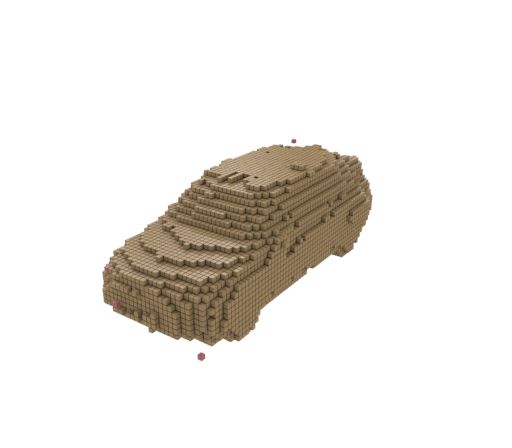}
        \end{subfigure}
        \begin{subfigure}[t]{0.15\textwidth}
            \vspace{0px}\centering
            \includegraphics[width=1.5cm,trim={\cropleft cm \croplower cm \cropright cm \cropupper cm},clip]{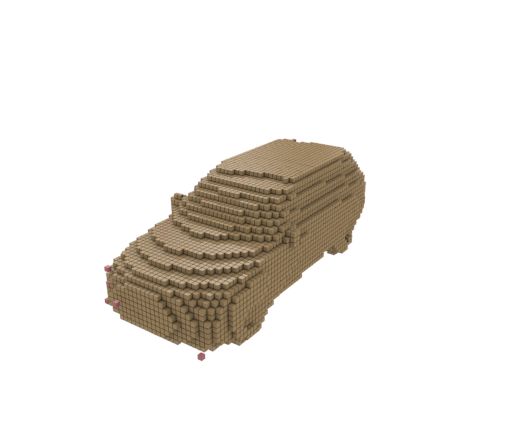}
        \end{subfigure}
        \begin{subfigure}[t]{0.15\textwidth}
            \vspace{0px}\centering
            \includegraphics[width=1.5cm,trim={\cropleft cm \croplower cm \cropright cm \cropupper cm},clip]{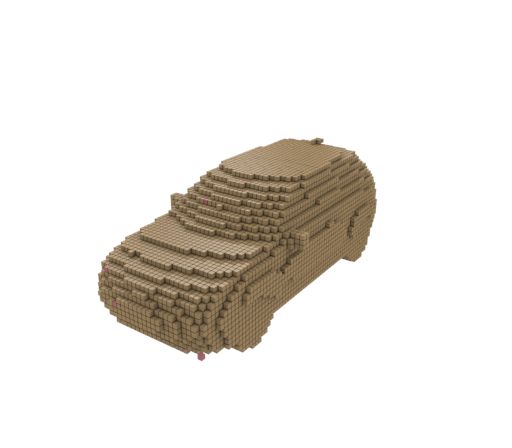}
        \end{subfigure}
        \subcaption{\clean and -noisy, Medium ($32\ntimes72\ntimes32$) and High ($48\ntimes108\ntimes48$) Resolution}
    \end{subfigure}
    \\[4px]
    \begin{subfigure}[t]{0.5\textwidth}
       	\vspace{0px}\centering
    	\begin{subfigure}[t]{0.15\textwidth}
    		\vspace{0px}\centering
    		\Dai\\
    		\includegraphics[width=1.5cm,trim={\cropleft cm \croplower cm \cropright cm \cropupper cm},clip]{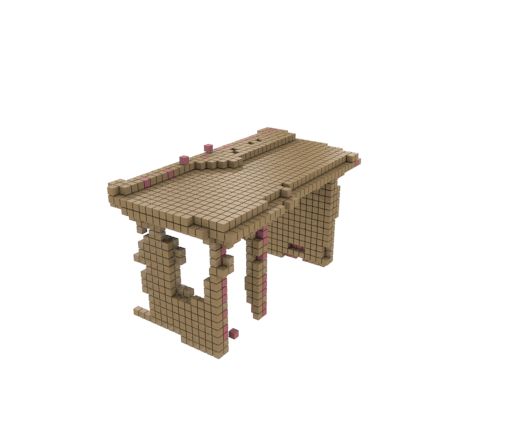}
    	\end{subfigure}
    	\begin{subfigure}[t]{0.15\textwidth}
    		\vspace{0px}\centering
    		\AML\\
    		\includegraphics[width=1.5cm,trim={\cropleft cm \croplower cm \cropright cm \cropupper cm},clip]{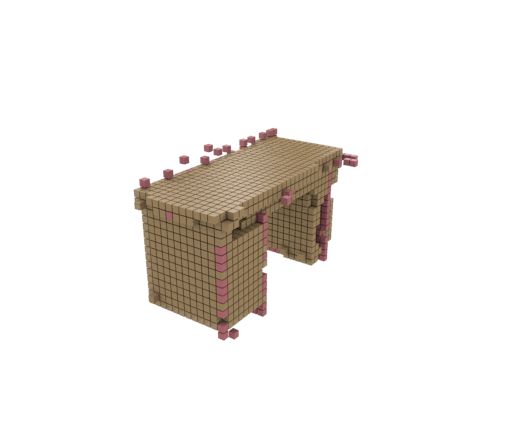}
    	\end{subfigure}
    	\begin{subfigure}[t]{0.15\textwidth}
    		\vspace{0px}\centering
    		GT\\
    		\includegraphics[width=1.5cm,trim={\cropleft cm \croplower cm \cropright cm \cropupper cm},clip]{gdat_modelnet_desk_low_\deska_bin}
    	\end{subfigure}
    	\begin{subfigure}[t]{0.15\textwidth}
    		\vspace{0px}\centering
    		\Dai\\
    		\includegraphics[width=1.5cm,trim={\cropleft cm \croplower cm \cropright cm \cropupper cm},clip]{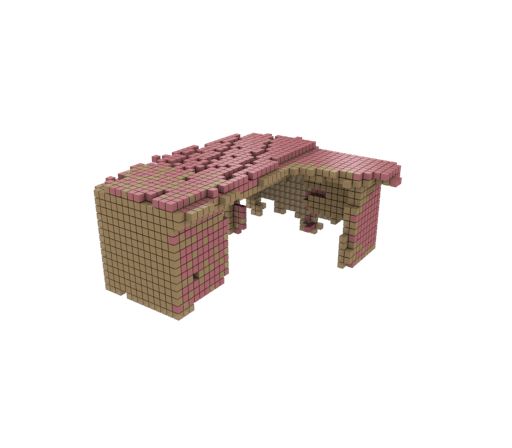}
    	\end{subfigure}
    	\begin{subfigure}[t]{0.15\textwidth}
    		\vspace{0px}\centering
    		\AML\\
    		\includegraphics[width=1.5cm,trim={\cropleft cm \croplower cm \cropright cm \cropupper cm},clip]{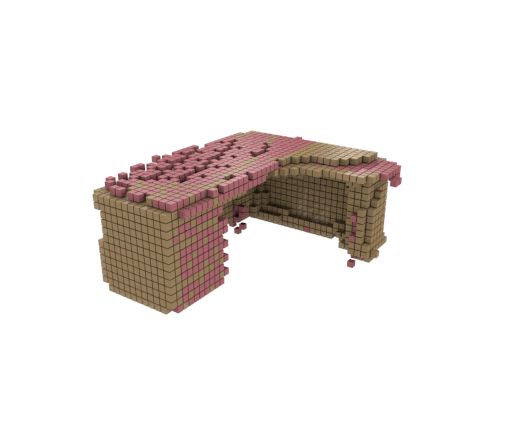}
    	\end{subfigure}
    	\begin{subfigure}[t]{0.15\textwidth}
    		\vspace{0px}\centering
    		GT\\
    		\includegraphics[width=1.5cm,trim={\cropleft cm \croplower cm \cropright cm \cropupper cm},clip]{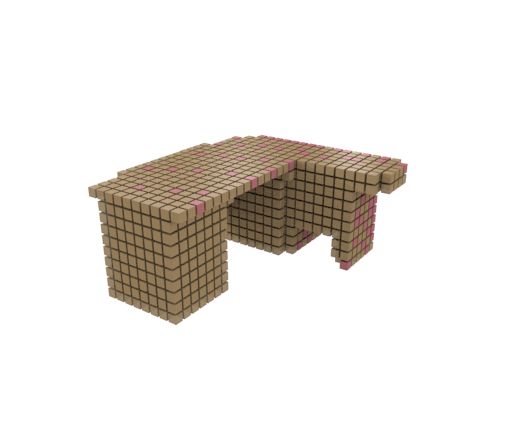}
    	\end{subfigure}
    	\\[-2px]
        \hspace*{-6px}
    	\begin{subfigure}[t]{0.15\textwidth}
    		\vspace{0px}\centering
    		\includegraphics[width=1.5cm,trim={\cropleft cm \croplower cm \cropright cm \cropupper cm},clip]{gexp_clean_chair_high_10_wide_d_sup_3_3_res_\chaira}
    	\end{subfigure}
    	\begin{subfigure}[t]{0.15\textwidth}
    		\vspace{0px}\centering
    		\includegraphics[width=1.5cm,trim={\cropleft cm \croplower cm \cropright cm \cropupper cm},clip]{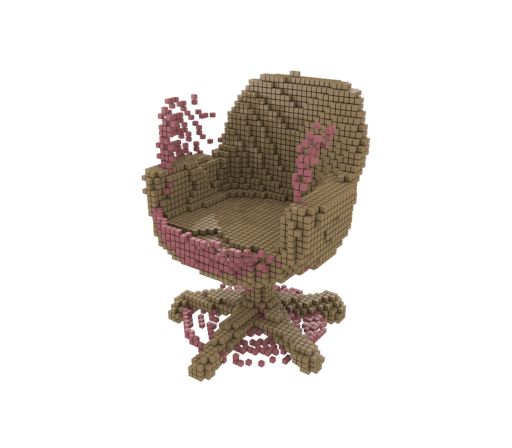}
    	\end{subfigure}
    	\begin{subfigure}[t]{0.15\textwidth}
    		\vspace{0px}\centering
    		\includegraphics[width=1.5cm,trim={\cropleft cm \croplower cm \cropright cm \cropupper cm},clip]{gdat_modelnet_chair_low_\chaira_bin}
    	\end{subfigure}
    	\begin{subfigure}[t]{0.15\textwidth}
    		\vspace{0px}\centering
    		\includegraphics[width=1.5cm,trim={\cropleft cm \croplower cm \cropright cm \cropupper cm},clip]{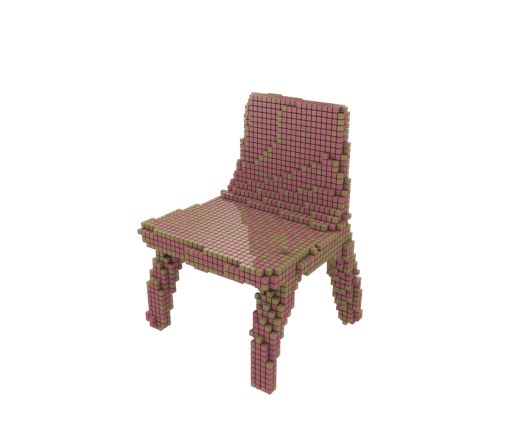}
    	\end{subfigure}
    	\begin{subfigure}[t]{0.15\textwidth}
    		\vspace{0px}\centering
    		\includegraphics[width=1.5cm,trim={\cropleft cm \croplower cm \cropright cm \cropupper cm},clip]{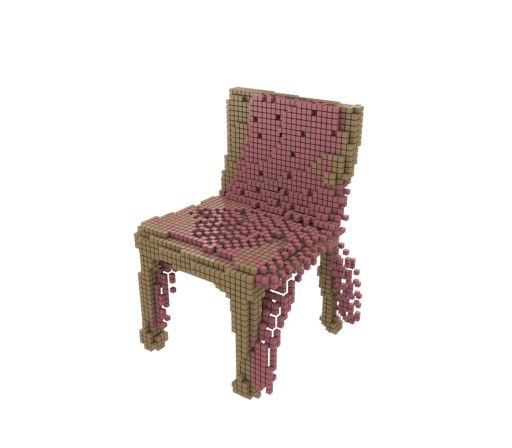}
    	\end{subfigure}
    	\begin{subfigure}[t]{0.15\textwidth}
    		\vspace{0px}\centering
    		\includegraphics[width=1.5cm,trim={\cropleft cm \croplower cm \cropright cm \cropupper cm},clip]{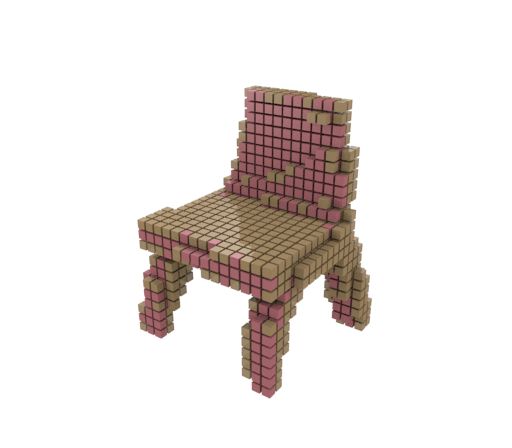}
    	\end{subfigure}
        \subcaption{ModelNet desks and chairs, Medium ($48^3$) and High ($64^3$) Resolution}
   	\end{subfigure}
    }
    \vspace*{-\figskipcaption px}
    \caption{{\bf Multi-View and Higher-Resolution Results on ShapeNet and ModelNet.} While \AML is designed for especially sparse observations, it also performs well in a multi-view setting. Additionally, higher resolutions allow to predict more detailed shapes. Shapes, occupancy grids or meshes, in {\color{rbeige}beige} and observations in {\color{rred}red}.}
    \label{fig:results-synthetic-extra}
    \vspace*{-\figskipbelow px}
\end{figure}
\newcommand{\modelneta}{15318} 
\newcommand{\modelnetb}{2664}
\newcommand{\modelnetc}{9324} 
\newcommand{\modelnetd}{18648} 
\newcommand{\modelnete}{9990}
\newcommand{\modelnetf}{5994}
\newcommand{\modelnetg}{12654}
\newcommand{\modelneth}{11988}
\newcommand{\modelneti}{7326}
\begin{figure}[t!]
    \vspace*{-\figskipabove px}
    \vspace*{2px}
    \centering
    {\scriptsize
    
    \begin{subfigure}[t]{0.07\textwidth}
        \vspace{0px}\centering
        \Dai\\
        \includegraphics[width=1.5cm,trim={\cropleft cm \croplower cm \cropright cm \cropupper cm},clip]{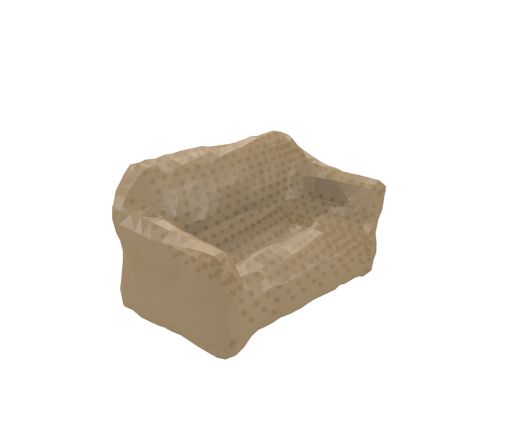}
    \end{subfigure}
    \begin{subfigure}[t]{0.07\textwidth}
        \vspace{0px}\centering
        \AML\\
        \includegraphics[width=1.5cm,trim={\cropleft cm \croplower cm \cropright cm \cropupper cm},clip]{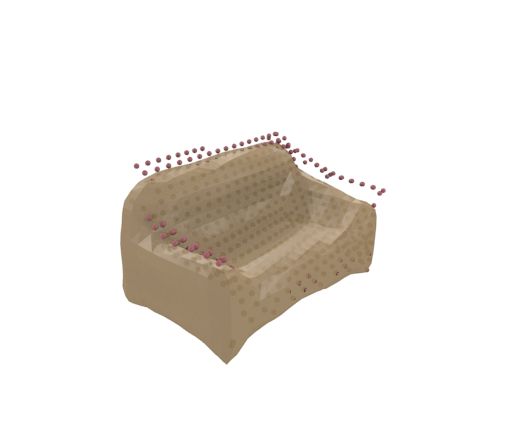}
    \end{subfigure}
    \begin{subfigure}[t]{0.07\textwidth}
        \vspace{0px}\centering
        GT\\
        \includegraphics[width=1.5cm,trim={\cropleft cm \croplower cm \cropright cm \cropupper cm},clip]{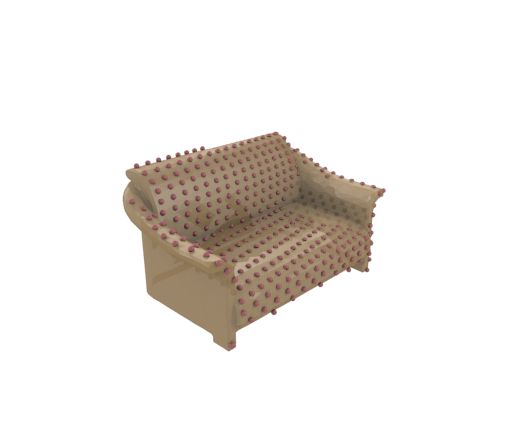}
    \end{subfigure}
    \begin{subfigure}[t]{0.07\textwidth}
        \vspace{0px}\centering
        \Dai\\
        \includegraphics[width=1.5cm,trim={\cropleft cm \croplower cm \cropright cm \cropupper cm},clip]{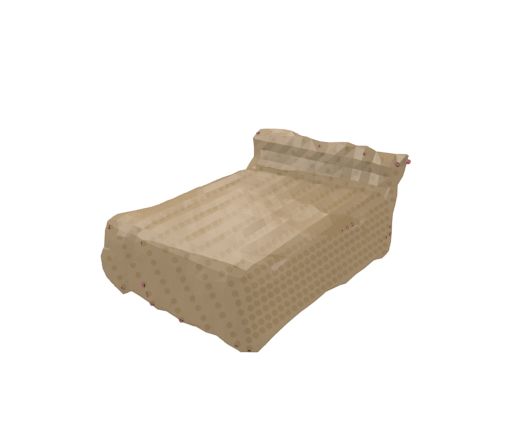}
    \end{subfigure}
    \begin{subfigure}[t]{0.07\textwidth}
        \vspace{0px}\centering
        \AML\\
        \includegraphics[width=1.5cm,trim={\cropleft cm \croplower cm \cropright cm \cropupper cm},clip]{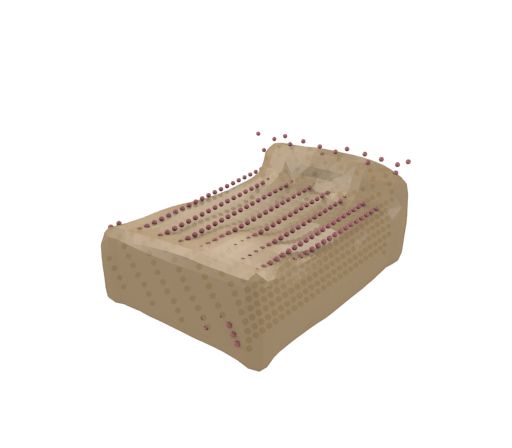}
    \end{subfigure}
    \begin{subfigure}[t]{0.07\textwidth}
        \vspace{0px}\centering
        GT\\
        \includegraphics[width=1.5cm,trim={\cropleft cm \croplower cm \cropright cm \cropupper cm},clip]{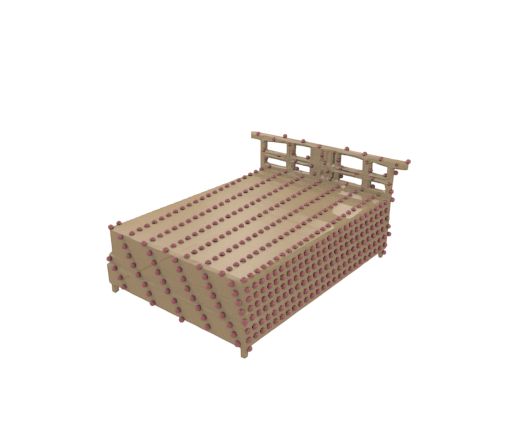}
    \end{subfigure}
    \\[-4px]
    \begin{subfigure}[t]{0.07\textwidth}
        \vspace{0px}\centering
        \includegraphics[width=1.5cm,trim={\cropleft cm \croplower cm \cropright cm \cropupper cm},clip]{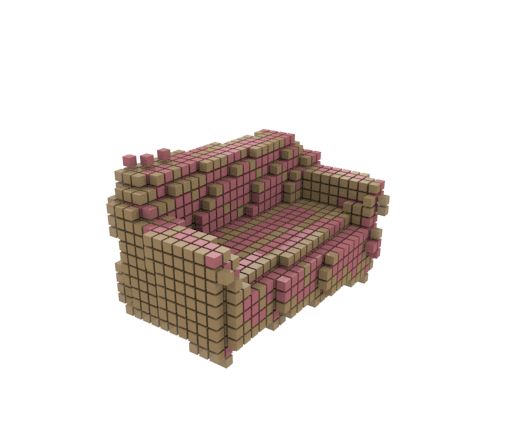}
    \end{subfigure}
    \begin{subfigure}[t]{0.07\textwidth}
        \vspace{0px}\centering
        \includegraphics[width=1.5cm,trim={\cropleft cm \croplower cm \cropright cm \cropupper cm},clip]{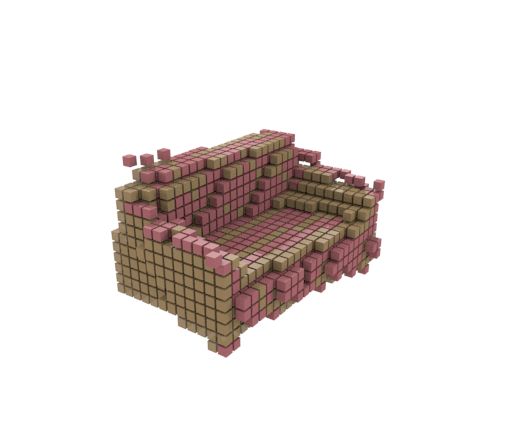}
    \end{subfigure}
    \begin{subfigure}[t]{0.07\textwidth}
        \vspace{0px}\centering
        \includegraphics[width=1.5cm,trim={\cropleft cm \croplower cm \cropright cm \cropupper cm},clip]{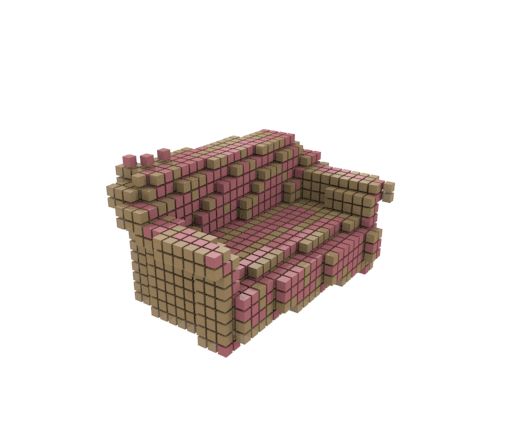}
    \end{subfigure}
    \begin{subfigure}[t]{0.07\textwidth}
        \vspace{0px}\centering
        \includegraphics[width=1.5cm,trim={\cropleft cm \croplower cm \cropright cm \cropupper cm},clip]{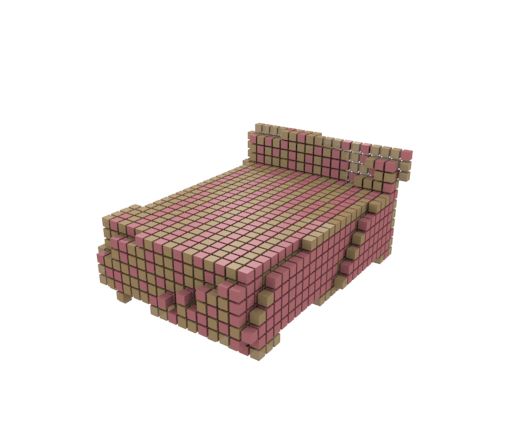}
    \end{subfigure}
    \begin{subfigure}[t]{0.07\textwidth}
        \vspace{0px}\centering
        \includegraphics[width=1.5cm,trim={\cropleft cm \croplower cm \cropright cm \cropupper cm},clip]{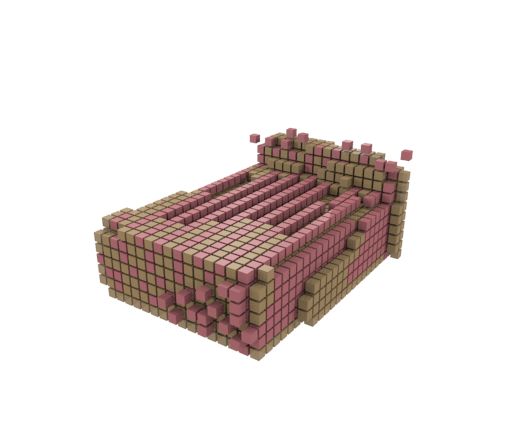}
    \end{subfigure}
    \begin{subfigure}[t]{0.07\textwidth}
        \vspace{0px}\centering
        \includegraphics[width=1.5cm,trim={\cropleft cm \croplower cm \cropright cm \cropupper cm},clip]{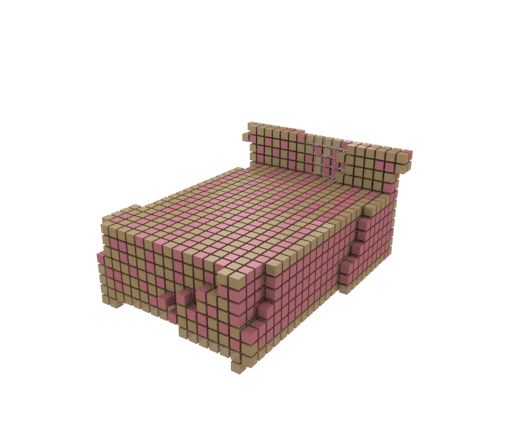}
    \end{subfigure}
    \\[-4px]
    \begin{subfigure}[t]{0.07\textwidth}
        \vspace{0px}\centering
        \includegraphics[width=1.5cm,trim={\cropleft cm \croplower cm \cropright cm \cropupper cm},clip]{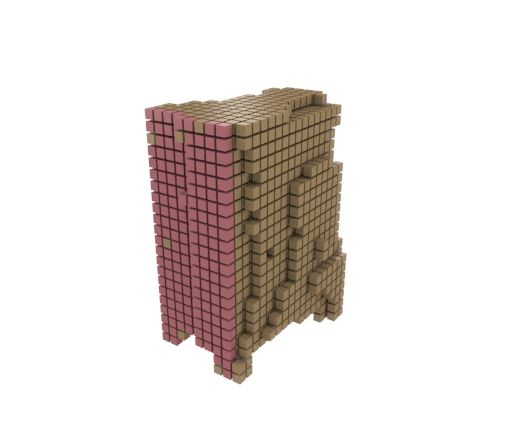}
    \end{subfigure}
    \begin{subfigure}[t]{0.07\textwidth}
        \vspace{0px}\centering
        \includegraphics[width=1.5cm,trim={\cropleft cm \croplower cm \cropright cm \cropupper cm},clip]{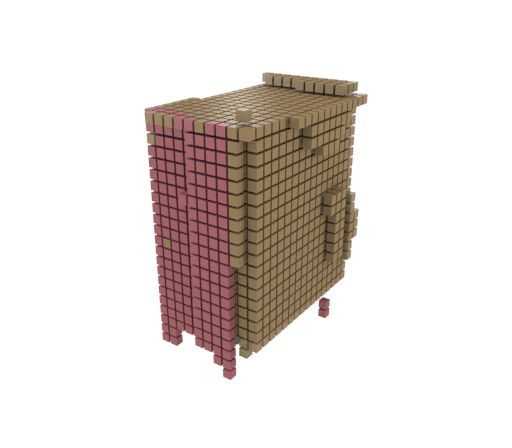}
    \end{subfigure}
    \begin{subfigure}[t]{0.07\textwidth}
        \vspace{0px}\centering
        \includegraphics[width=1.5cm,trim={\cropleft cm \croplower cm \cropright cm \cropupper cm},clip]{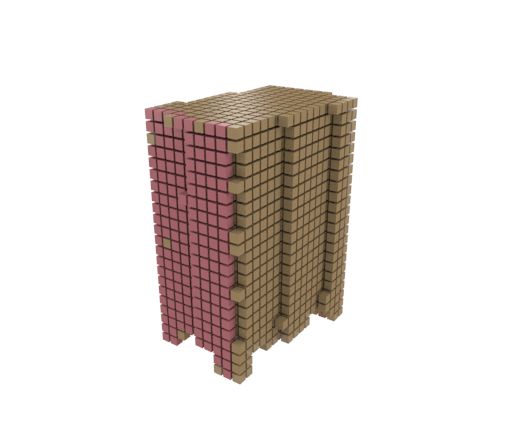}
    \end{subfigure}
    \begin{subfigure}[t]{0.07\textwidth}
        \vspace{0px}\centering
        \includegraphics[width=1.5cm,trim={\cropleft cm \croplower cm \cropright cm \cropupper cm},clip]{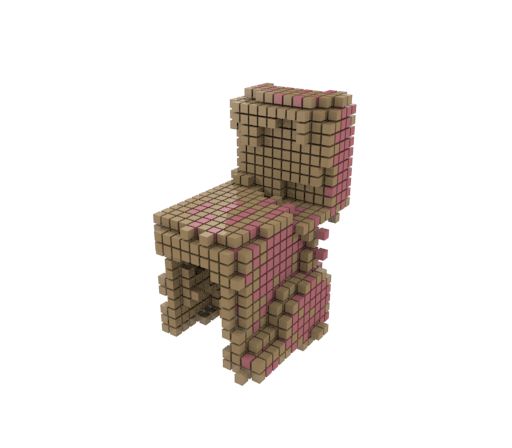}
    \end{subfigure}
    \begin{subfigure}[t]{0.07\textwidth}
        \vspace{0px}\centering
        \includegraphics[width=1.5cm,trim={\cropleft cm \croplower cm \cropright cm \cropupper cm},clip]{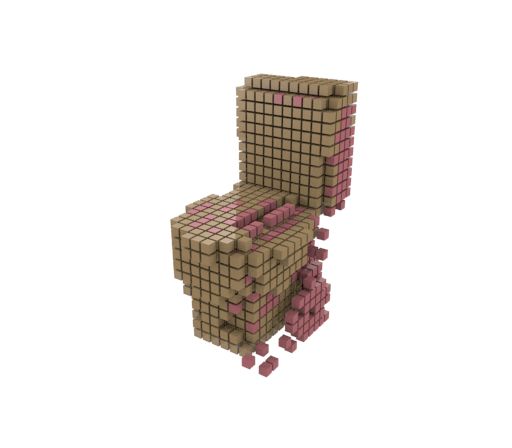}
    \end{subfigure}
    \begin{subfigure}[t]{0.07\textwidth}
        \vspace{0px}\centering
        \includegraphics[width=1.5cm,trim={\cropleft cm \croplower cm \cropright cm \cropupper cm},clip]{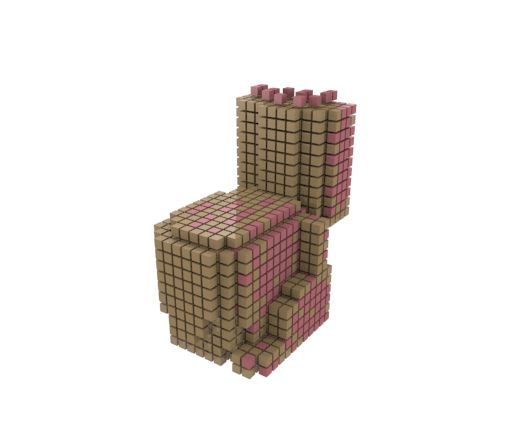}
    \end{subfigure}
    \\[-4px]
    \begin{subfigure}[t]{0.07\textwidth}
        \vspace{0px}\centering
        \includegraphics[width=1.5cm,trim={\cropleft cm \croplower cm \cropright cm \cropupper cm},clip]{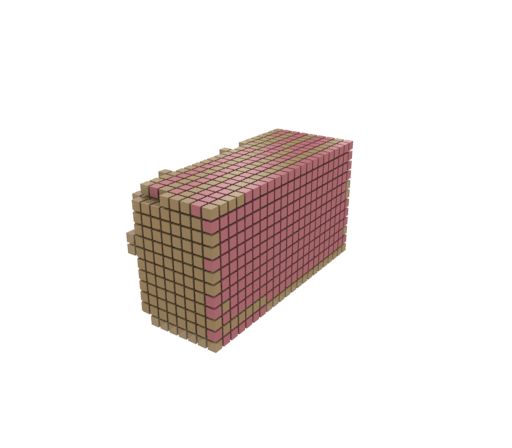}
    \end{subfigure}
    \begin{subfigure}[t]{0.07\textwidth}
        \vspace{0px}\centering
        \includegraphics[width=1.5cm,trim={\cropleft cm \croplower cm \cropright cm \cropupper cm},clip]{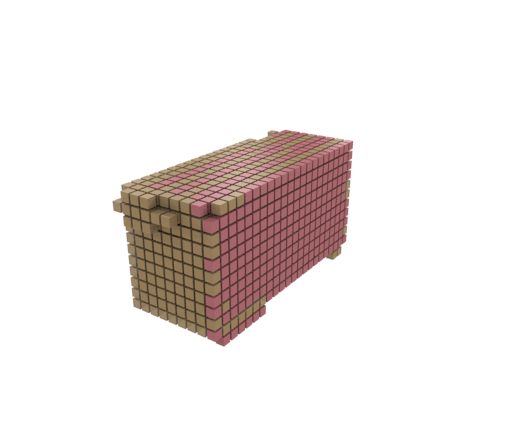}
    \end{subfigure}
    \begin{subfigure}[t]{0.07\textwidth}
        \vspace{0px}\centering
        \includegraphics[width=1.5cm,trim={\cropleft cm \croplower cm \cropright cm \cropupper cm},clip]{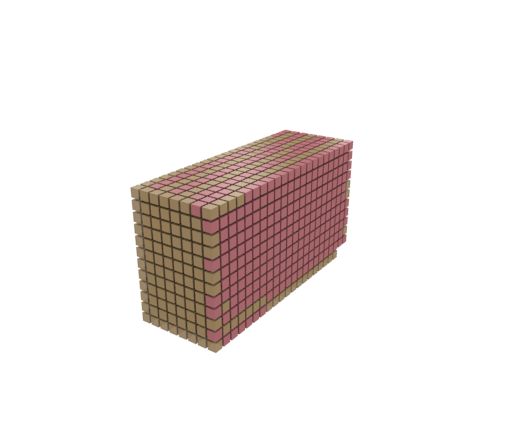}
    \end{subfigure}
    \begin{subfigure}[t]{0.07\textwidth}
        \vspace{0px}\centering
        \includegraphics[width=1.5cm,trim={\cropleft cm \croplower cm \cropright cm \cropupper cm},clip]{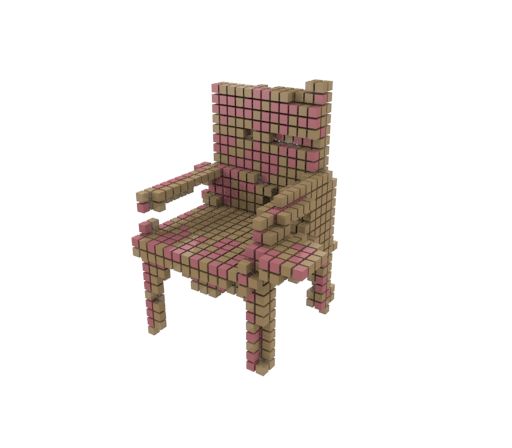}
    \end{subfigure}
    \begin{subfigure}[t]{0.07\textwidth}
        \vspace{0px}\centering
        \includegraphics[width=1.5cm,trim={\cropleft cm \croplower cm \cropright cm \cropupper cm},clip]{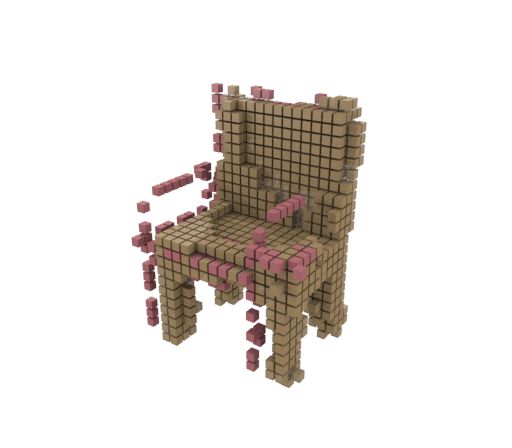}
    \end{subfigure}
    \begin{subfigure}[t]{0.07\textwidth}
        \vspace{0px}\centering
        \includegraphics[width=1.5cm,trim={\cropleft cm \croplower cm \cropright cm \cropupper cm},clip]{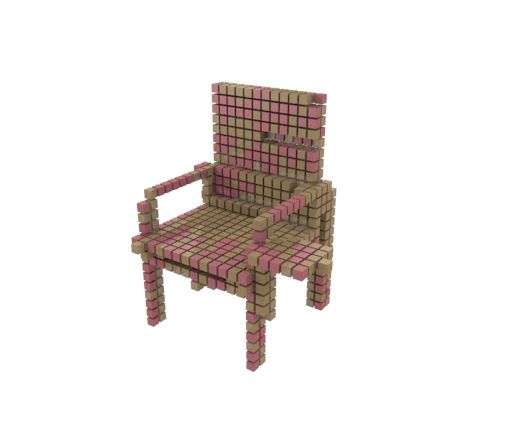}
    \end{subfigure}
    \\[-4px]
    \begin{subfigure}[t]{0.07\textwidth}
        \vspace{0px}\centering
        \includegraphics[width=1.5cm,trim={\cropleft cm \croplower cm \cropright cm \cropupper cm},clip]{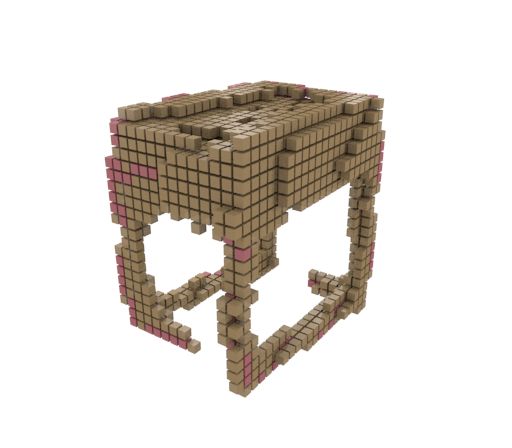}
    \end{subfigure}
    \begin{subfigure}[t]{0.07\textwidth}
        \vspace{0px}\centering
        \includegraphics[width=1.5cm,trim={\cropleft cm \croplower cm \cropright cm \cropupper cm},clip]{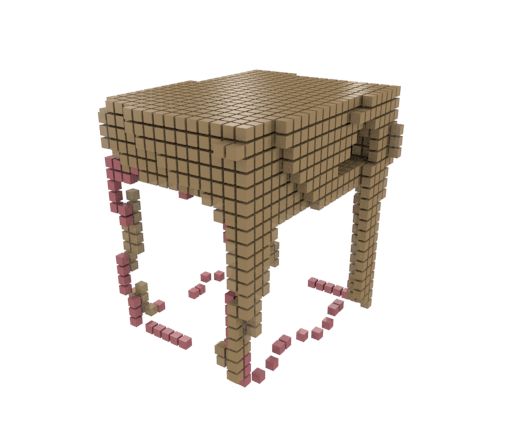}
    \end{subfigure}
    \begin{subfigure}[t]{0.07\textwidth}
        \vspace{0px}\centering
        \includegraphics[width=1.5cm,trim={\cropleft cm \croplower cm \cropright cm \cropupper cm},clip]{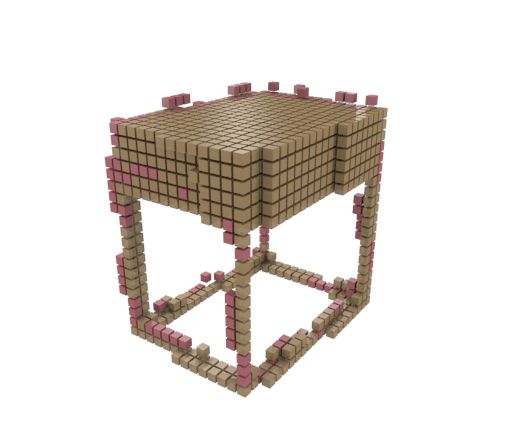}
    \end{subfigure}\begin{subfigure}[t]{0.07\textwidth}
        \vspace{0px}\centering
        \includegraphics[width=1.5cm,trim={\cropleft cm \croplower cm \cropright cm \cropupper cm},clip]{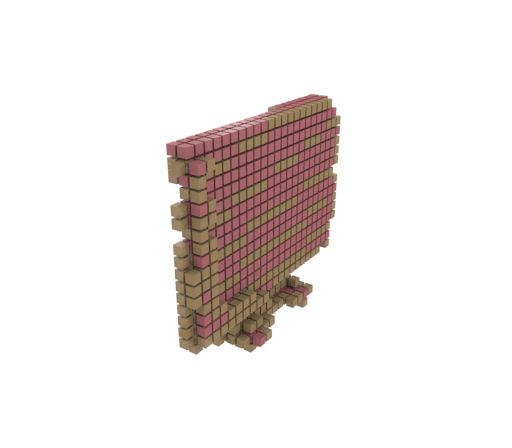}
    \end{subfigure}
    \begin{subfigure}[t]{0.07\textwidth}
        \vspace{0px}\centering
        \includegraphics[width=1.5cm,trim={\cropleft cm \croplower cm \cropright cm \cropupper cm},clip]{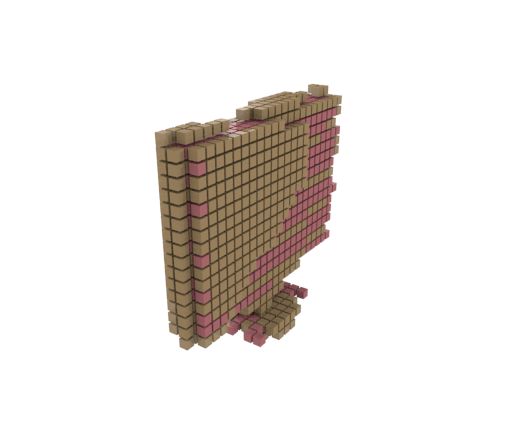}
    \end{subfigure}
    \begin{subfigure}[t]{0.07\textwidth}
        \vspace{0px}\centering
        \includegraphics[width=1.5cm,trim={\cropleft cm \croplower cm \cropright cm \cropupper cm},clip]{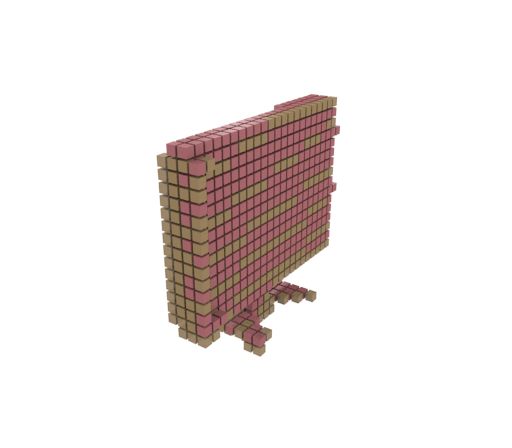}
    \end{subfigure}
    }
    \vspace*{-\figskipcaption px}
    \caption{{\bf Category-Agnostic Results on ModelNet10.} \AML is able to recover detailed shapes of the correct object category even without category supervision (as provided to \Dai). Shapes (occupancy grids and meshes) in {\color{rbeige}beige} and observations in {\color{rred}red} at low resolution ($32^3$ voxels).}
    \label{fig:results-modelnet-extra-a}
    \vspace*{-\figskipbelow px}
\end{figure}
\newcommand{\kittia}{0}
\newcommand{\kittib}{612}
\newcommand{\kittic}{1224}
\newcommand{\kittid}{2754}
\newcommand{\kittie}{3060}
\newcommand{\kittif}{5508} 
\begin{figure}[t]
    \vspace*{-\figskipabove px}
    \vspace*{2px}
	\centering
	{\scriptsize 

    \begin{subfigure}[t]{0.5\textwidth}
        \vspace{0px}\centering
		\begin{subfigure}[t]{0.15\textwidth}
			\vspace{0px}\centering
			Obs\\
			\includegraphics[width=1.5cm,trim={\cropleft cm \croplower cm \cropright cm \cropupper cm},clip]{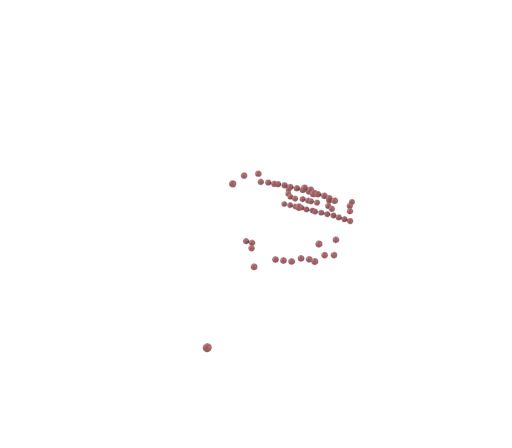}
		\end{subfigure}
		\begin{subfigure}[t]{0.15\textwidth}
			\vspace{0px}\centering
			\Dai\\
			\includegraphics[width=1.5cm,trim={\cropleft cm \croplower cm \cropright cm \cropupper cm},clip]{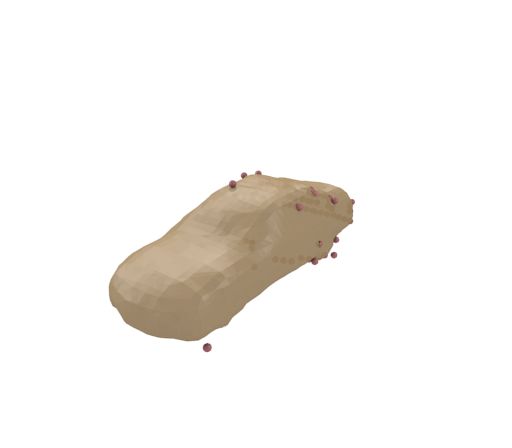}
		\end{subfigure}
		\begin{subfigure}[t]{0.15\textwidth}
			\vspace{0px}\centering
			\Engelmann\\
			\includegraphics[width=1.5cm,trim={\cropleft cm \croplower cm \cropright cm \cropupper cm},clip]{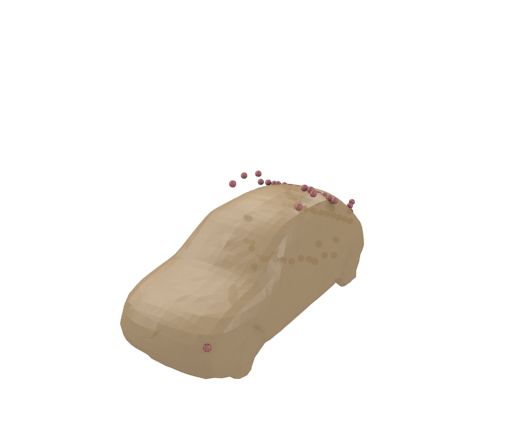}
		\end{subfigure}
		\begin{subfigure}[t]{0.15\textwidth}
			\vspace{0px}\centering
			\AML\\
			\includegraphics[width=1.5cm,trim={\cropleft cm \croplower cm \cropright cm \cropupper cm},clip]{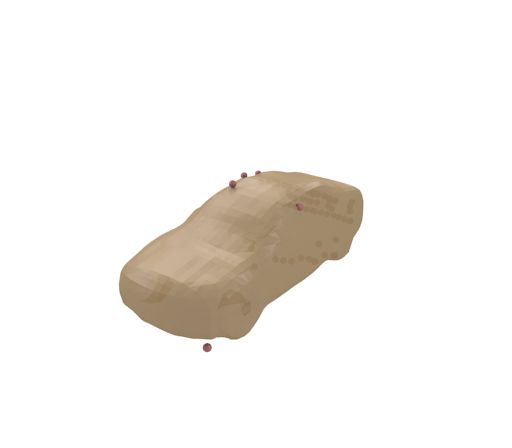}
		\end{subfigure}
		\begin{subfigure}[t]{0.15\textwidth}
			\vspace{0px}\centering
			\AML\\
			\includegraphics[width=1.5cm,trim={\cropleft cm \croplower cm \cropright cm \cropupper cm},clip]{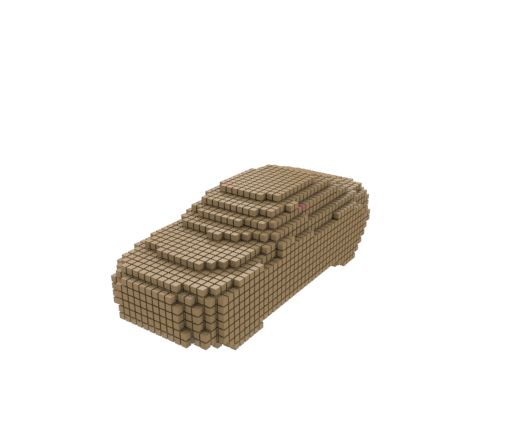}
		\end{subfigure}
		\begin{subfigure}[t]{0.15\textwidth}
			\vspace{0px}\centering
			GT\\
			\includegraphics[width=1.5cm,trim={\cropleft cm \croplower cm \cropright cm \cropupper cm},clip]{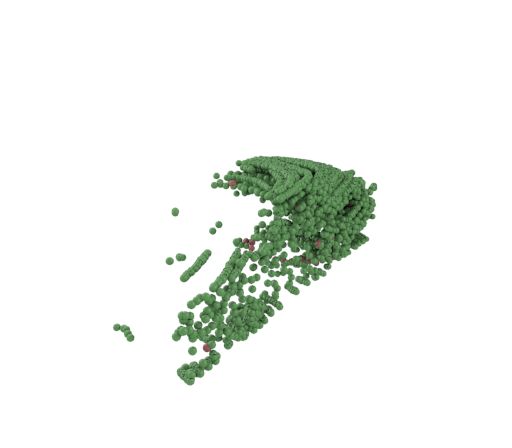}
		\end{subfigure}
		\\[-4px]
		\begin{subfigure}[t]{0.15\textwidth}
			\vspace{0px}\centering
			\includegraphics[width=1.5cm,trim={\cropleft cm \croplower cm \cropright cm \cropupper cm},clip]{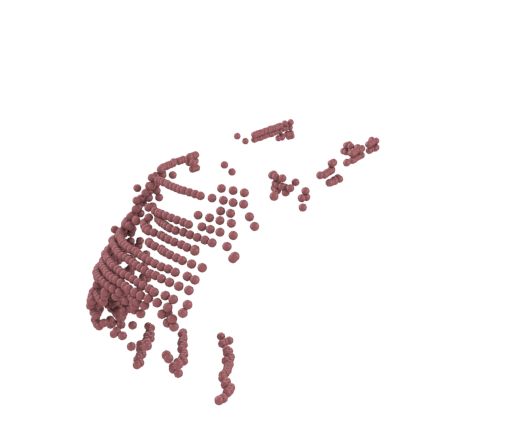}
		\end{subfigure}
		\begin{subfigure}[t]{0.15\textwidth}
			\vspace{0px}\centering
			\includegraphics[width=1.5cm,trim={\cropleft cm \croplower cm \cropright cm \cropupper cm},clip]{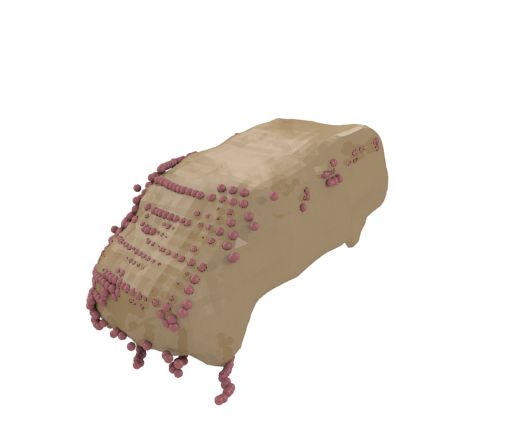}
		\end{subfigure}
		\begin{subfigure}[t]{0.15\textwidth}
			\vspace{0px}\centering
			\includegraphics[width=1.5cm,trim={\cropleft cm \croplower cm \cropright cm \cropupper cm},clip]{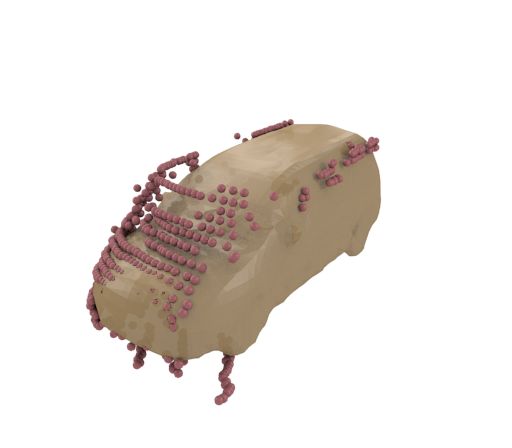}
		\end{subfigure}
		\begin{subfigure}[t]{0.15\textwidth}
			\vspace{0px}\centering
			\includegraphics[width=1.5cm,trim={\cropleft cm \croplower cm \cropright cm \cropupper cm},clip]{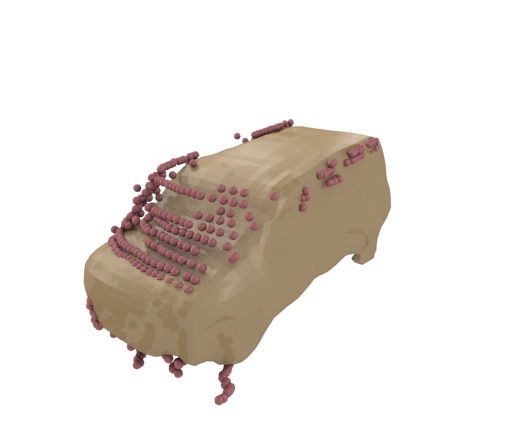}
		\end{subfigure}
		\begin{subfigure}[t]{0.15\textwidth}
			\vspace{0px}\centering
			\includegraphics[width=1.5cm,trim={\cropleft cm \croplower cm \cropright cm \cropupper cm},clip]{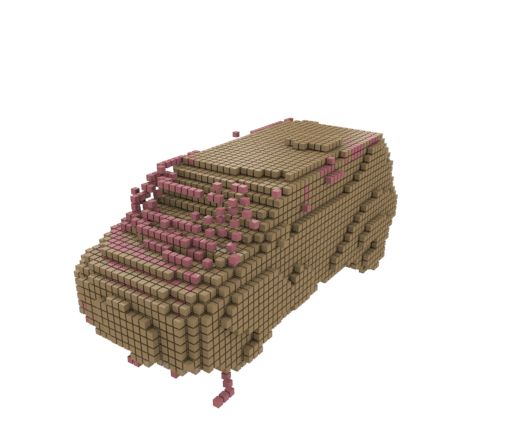}
		\end{subfigure}
		\begin{subfigure}[t]{0.15\textwidth}
			\vspace{0px}\centering
			\includegraphics[width=1.5cm,trim={\cropleft cm \croplower cm \cropright cm \cropupper cm},clip]{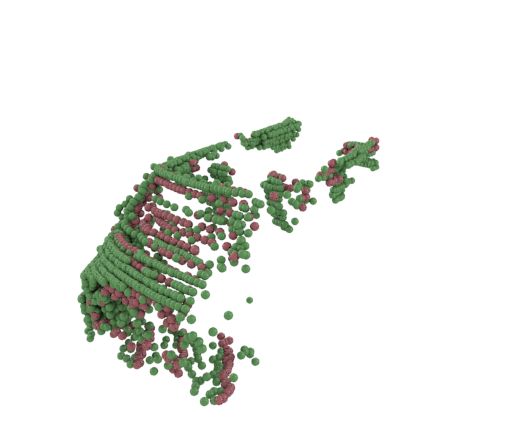}
		\end{subfigure}
		\\[-4px]
		\begin{subfigure}[t]{0.15\textwidth}
			\vspace{0px}\centering
			\includegraphics[width=1.5cm,trim={\cropleft cm \croplower cm \cropright cm \cropupper cm},clip]{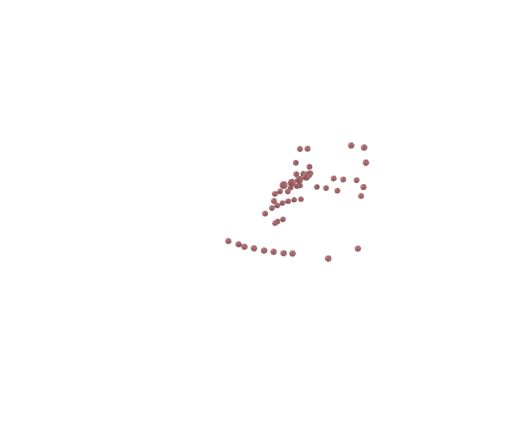}
		\end{subfigure}
		\begin{subfigure}[t]{0.15\textwidth}
			\vspace{0px}\centering
			\includegraphics[width=1.5cm,trim={\cropleft cm \croplower cm \cropright cm \cropupper cm},clip]{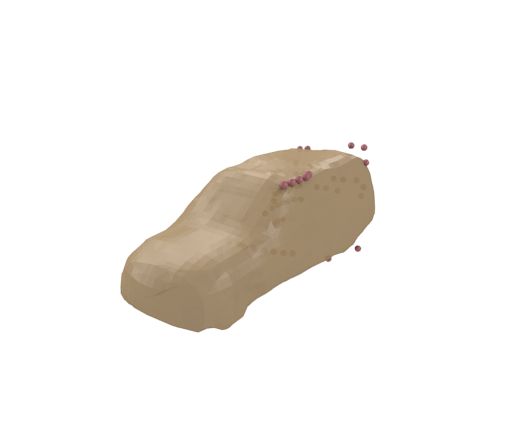}
		\end{subfigure}
		\begin{subfigure}[t]{0.15\textwidth}
			\vspace{0px}\centering
			\includegraphics[width=1.5cm,trim={\cropleft cm \croplower cm \cropright cm \cropupper cm},clip]{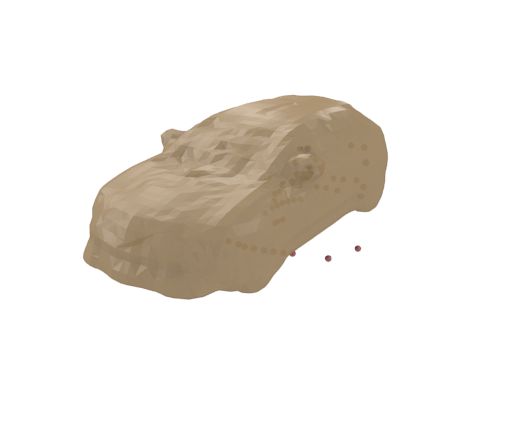}
		\end{subfigure}
		\begin{subfigure}[t]{0.15\textwidth}
			\vspace{0px}\centering
			\includegraphics[width=1.5cm,trim={\cropleft cm \croplower cm \cropright cm \cropupper cm},clip]{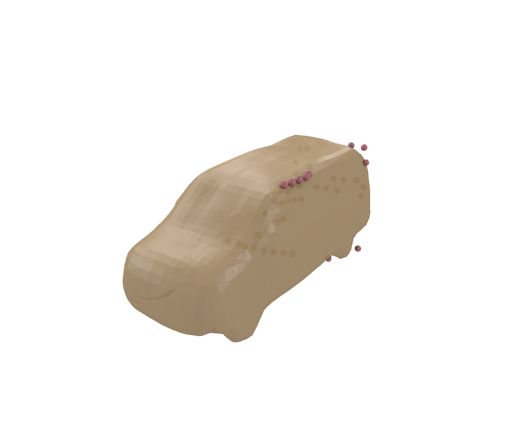}
		\end{subfigure}
		\begin{subfigure}[t]{0.15\textwidth}
			\vspace{0px}\centering
			\includegraphics[width=1.5cm,trim={\cropleft cm \croplower cm \cropright cm \cropupper cm},clip]{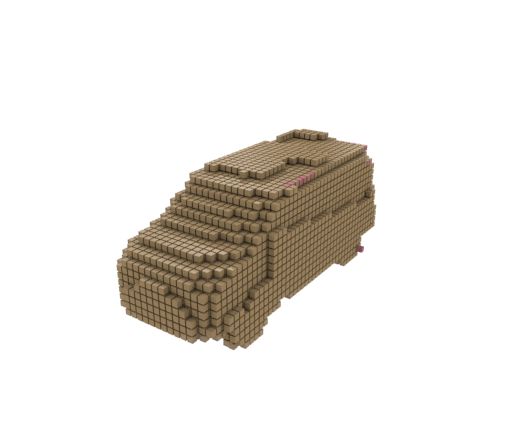}
		\end{subfigure}
		\begin{subfigure}[t]{0.15\textwidth}
			\vspace{0px}\centering
			\includegraphics[width=1.5cm,trim={\cropleft cm \croplower cm \cropright cm \cropupper cm},clip]{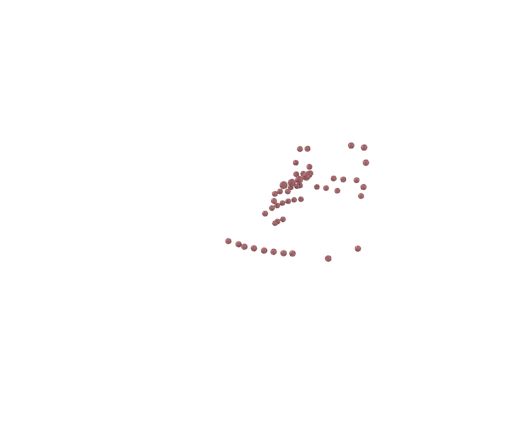}
		\end{subfigure}
		\\[-4px]
		\begin{subfigure}[t]{0.15\textwidth}
			\vspace{0px}\centering
			\includegraphics[width=1.5cm,trim={\cropleft cm \croplower cm \cropright cm \cropupper cm},clip]{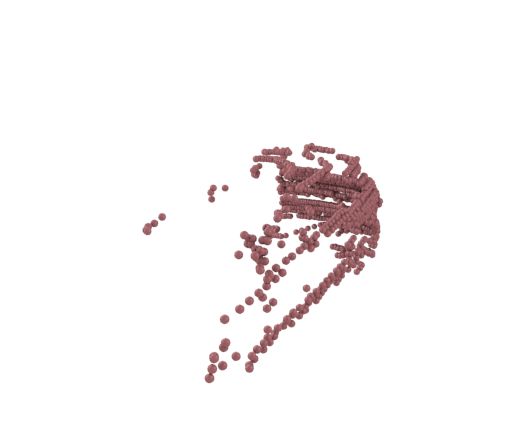}
		\end{subfigure}
		\begin{subfigure}[t]{0.15\textwidth}
			\vspace{0px}\centering
			\includegraphics[width=1.5cm,trim={\cropleft cm \croplower cm \cropright cm \cropupper cm},clip]{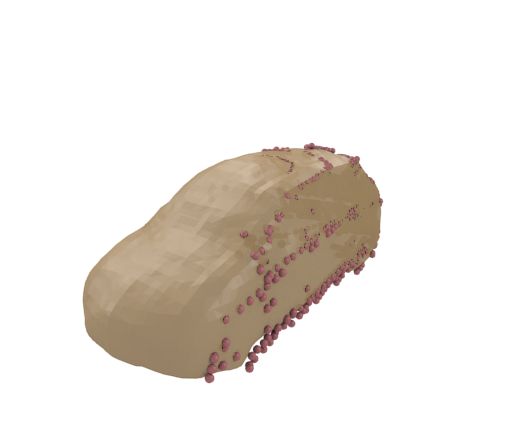}
		\end{subfigure}
		\begin{subfigure}[t]{0.15\textwidth}
			\vspace{0px}\centering
			\includegraphics[width=1.5cm,trim={\cropleft cm \croplower cm \cropright cm \cropupper cm},clip]{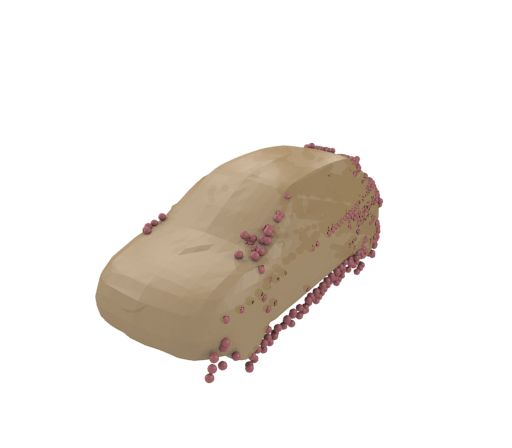}
		\end{subfigure}
		\begin{subfigure}[t]{0.15\textwidth}
			\vspace{0px}\centering
			\includegraphics[width=1.5cm,trim={\cropleft cm \croplower cm \cropright cm \cropupper cm},clip]{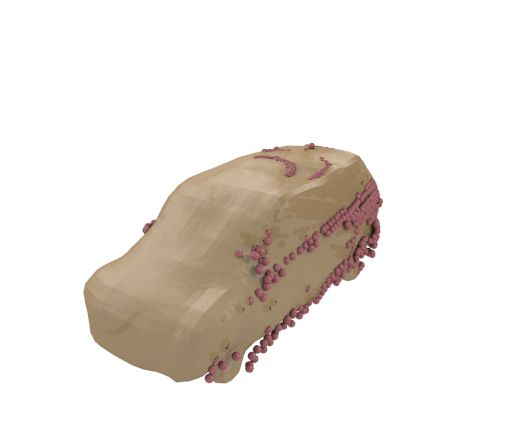}
		\end{subfigure}
		\begin{subfigure}[t]{0.15\textwidth}
			\vspace{0px}\centering
			\includegraphics[width=1.5cm,trim={\cropleft cm \croplower cm \cropright cm \cropupper cm},clip]{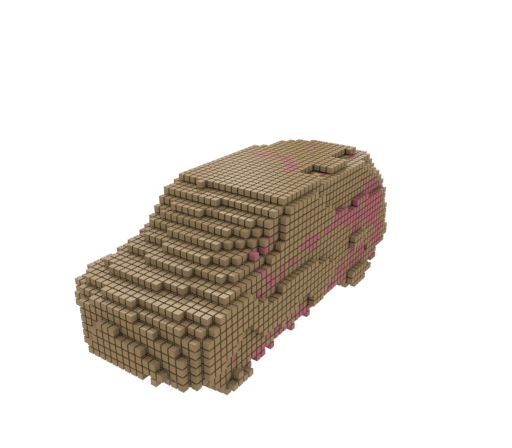}
		\end{subfigure}
		\begin{subfigure}[t]{0.15\textwidth}
			\vspace{0px}\centering
			\includegraphics[width=1.5cm,trim={\cropleft cm \croplower cm \cropright cm \cropupper cm},clip]{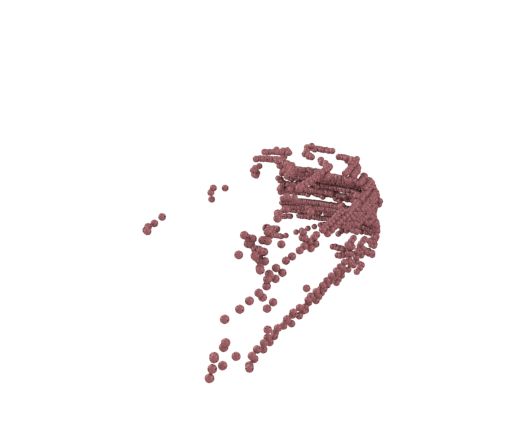}
		\end{subfigure}
		\\ 
		\begin{subfigure}[t]{0.15\textwidth}
			\vspace{0px}\centering
			\includegraphics[width=1.5cm,trim={\cropleft cm \croplower cm \cropright cm \cropupper cm},clip]{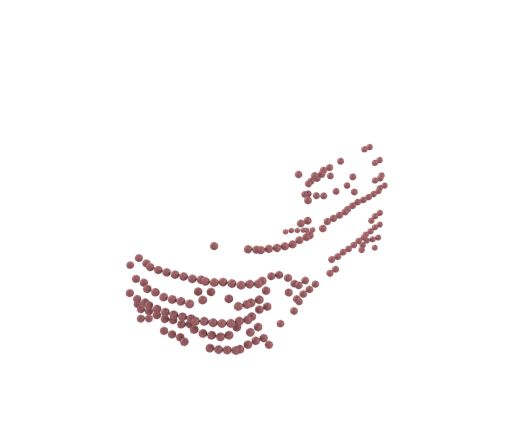}
		\end{subfigure}
		\begin{subfigure}[t]{0.15\textwidth}
			\vspace{0px}\centering
			\includegraphics[width=1.5cm,trim={\cropleft cm \croplower cm \cropright cm \cropupper cm},clip]{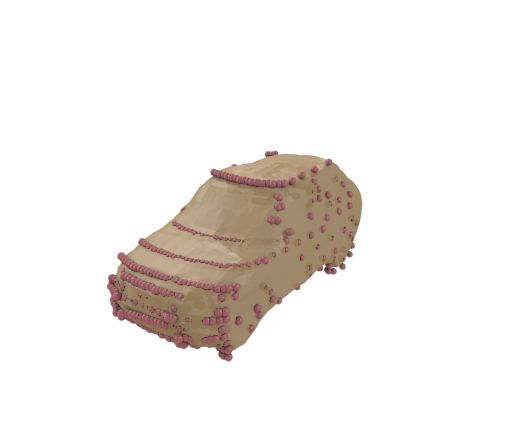}
		\end{subfigure}
		\begin{subfigure}[t]{0.15\textwidth}
			\vspace{0px}\centering
			\includegraphics[width=1.5cm,trim={\cropleft cm \croplower cm \cropright cm \cropupper cm},clip]{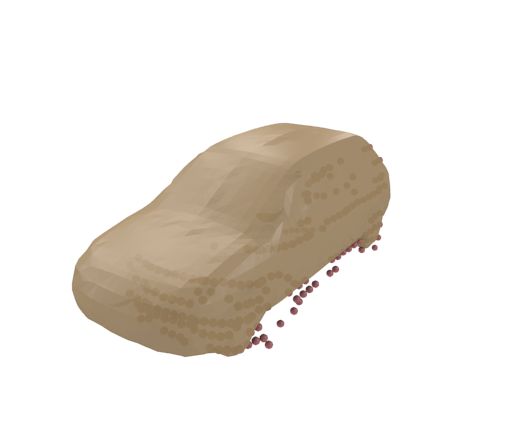}
		\end{subfigure}
		\begin{subfigure}[t]{0.15\textwidth}
			\vspace{0px}\centering
			\includegraphics[width=1.5cm,trim={\cropleft cm \croplower cm \cropright cm \cropupper cm},clip]{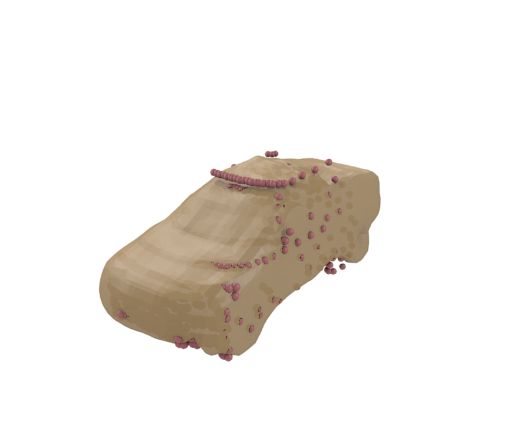}
		\end{subfigure}
		\begin{subfigure}[t]{0.15\textwidth}
			\vspace{0px}\centering
			\includegraphics[width=1.5cm,trim={\cropleft cm \croplower cm \cropright cm \cropupper cm},clip]{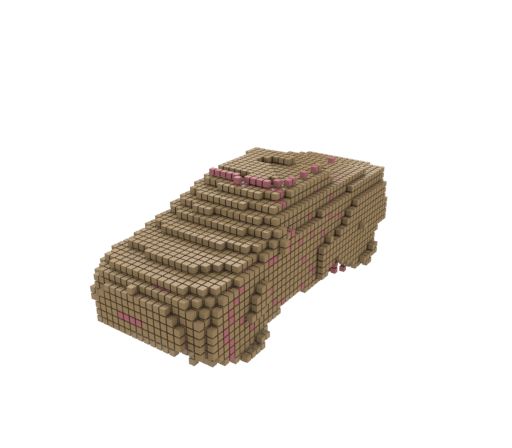}
		\end{subfigure}
		\begin{subfigure}[t]{0.15\textwidth}
			\vspace{0px}\centering
			\includegraphics[width=1.5cm,trim={\cropleft cm \croplower cm \cropright cm \cropupper cm},clip]{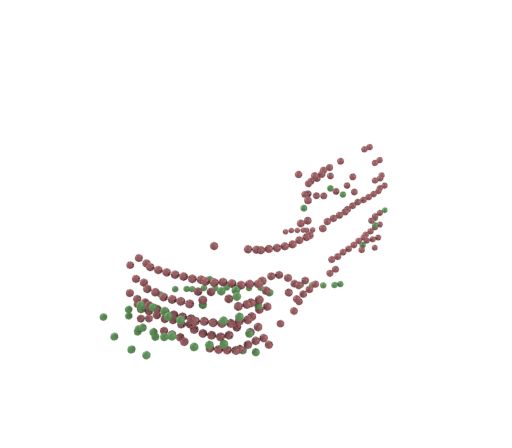}
		\end{subfigure}
        \subcaption{KITTI, Medium Resolution ($32\ntimes72\ntimes32$)}
	\end{subfigure}
	\\[4px]
	\begin{subfigure}[t]{0.5\textwidth}
        \vspace{0px}\centering
		\begin{subfigure}[t]{0.01\textwidth}
			\vspace{0px}\centering
			\rotatebox[]{90}{Obs\hspace*{0.25cm}}
		\end{subfigure}
		\begin{subfigure}[t]{0.15\textwidth}
			\vspace{0px}\centering
			\includegraphics[width=1.55cm,trim={\cropleft cm \croplower cm \cropright cm \cropupper cm},clip]{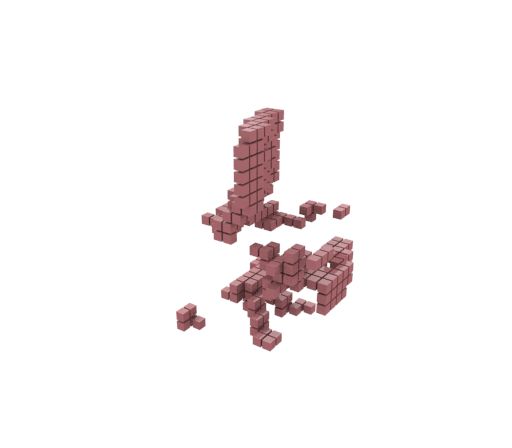}
		\end{subfigure}
		\begin{subfigure}[t]{0.15\textwidth}
			\vspace{0px}\centering
			\includegraphics[width=1.55cm,trim={\cropleft cm \croplower cm \cropright cm \cropupper cm},clip]{gdat_yang_chair_1_bin_points}
		\end{subfigure}
		\begin{subfigure}[t]{0.15\textwidth}
			\vspace{0px}\centering
			\includegraphics[width=1.55cm,trim={\cropleft cm \croplower cm \cropright cm \cropupper cm},clip]{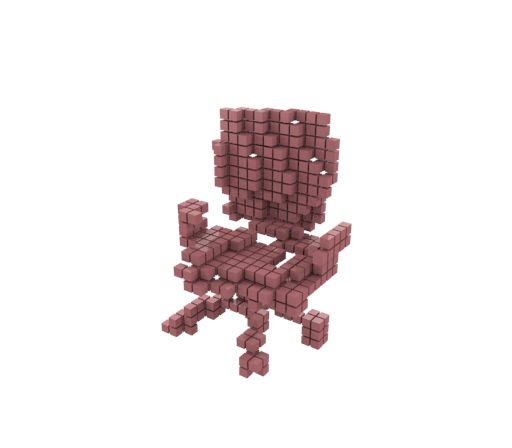}
		\end{subfigure}
		\begin{subfigure}[t]{0.15\textwidth}
			\vspace{0px}\centering
			\includegraphics[width=1.55cm,trim={\cropleft cm \croplower cm \cropright cm \cropupper cm},clip]{gdat_yang_table_5_bin_points}
		\end{subfigure}
		\begin{subfigure}[t]{0.15\textwidth}
			\vspace{0px}\centering
			\includegraphics[width=1.55cm,trim={\cropleft cm \croplower cm \cropright cm \cropupper cm},clip]{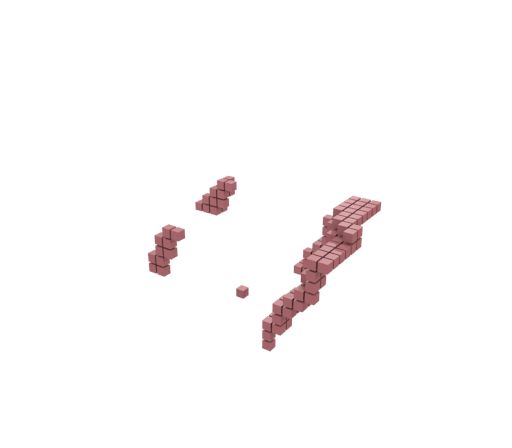}
		\end{subfigure}
		\begin{subfigure}[t]{0.15\textwidth}
			\vspace{0px}\centering
			\includegraphics[width=1.55cm,trim={\cropleft cm \croplower cm \cropright cm \cropupper cm},clip]{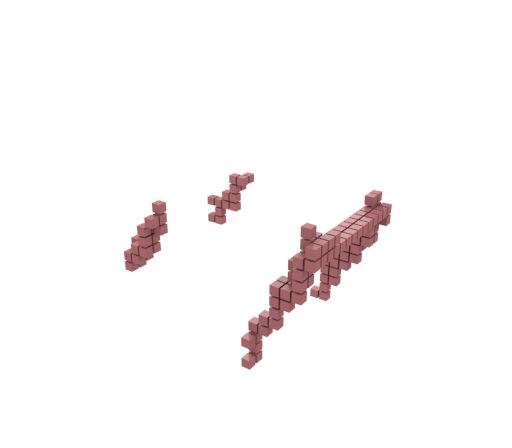}
		\end{subfigure}
		\\
		\begin{subfigure}[t]{0.01\textwidth}
			\vspace{0px}\centering
			\rotatebox[]{90}{\AML\hspace*{0.25cm}}
		\end{subfigure}
		\begin{subfigure}[t]{0.15\textwidth}
			\vspace{0px}\centering
			\includegraphics[width=1.55cm,trim={\cropleft cm \croplower cm \cropright cm \cropupper cm},clip]{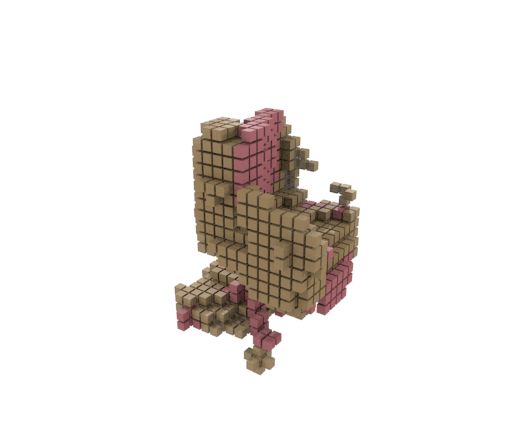}
		\end{subfigure}
		\begin{subfigure}[t]{0.15\textwidth}
			\vspace{0px}\centering
			\includegraphics[width=1.55cm,trim={\cropleft cm \croplower cm \cropright cm \cropupper cm},clip]{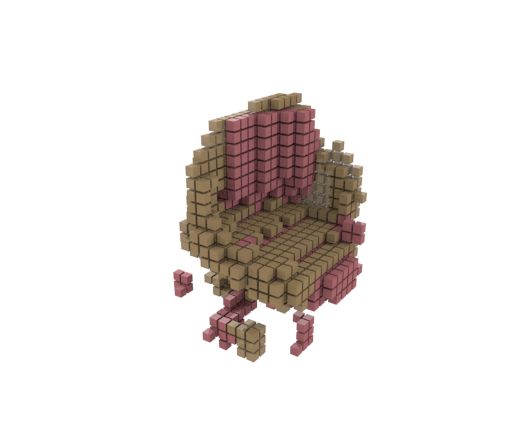}
		\end{subfigure}
		\begin{subfigure}[t]{0.15\textwidth}
			\vspace{0px}\centering
			\includegraphics[width=1.55cm,trim={\cropleft cm \croplower cm \cropright cm \cropupper cm},clip]{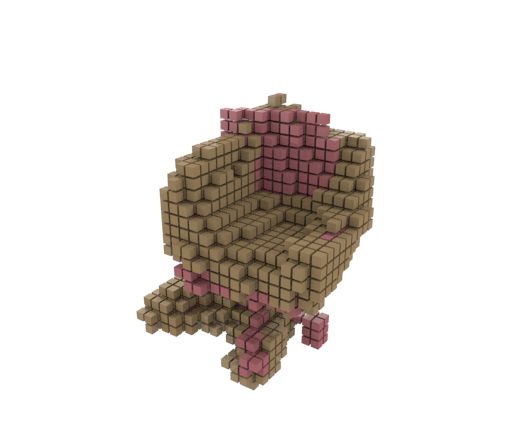}
		\end{subfigure}
		\begin{subfigure}[t]{0.15\textwidth}
			\vspace{0px}\centering
			\includegraphics[width=1.55cm,trim={\cropleft cm \croplower cm \cropright cm \cropupper cm},clip]{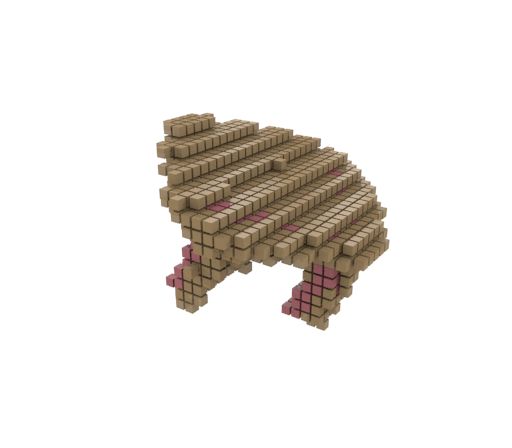}
		\end{subfigure}
		\begin{subfigure}[t]{0.15\textwidth}
			\vspace{0px}\centering
			\includegraphics[width=1.55cm,trim={\cropleft cm \croplower cm \cropright cm \cropupper cm},clip]{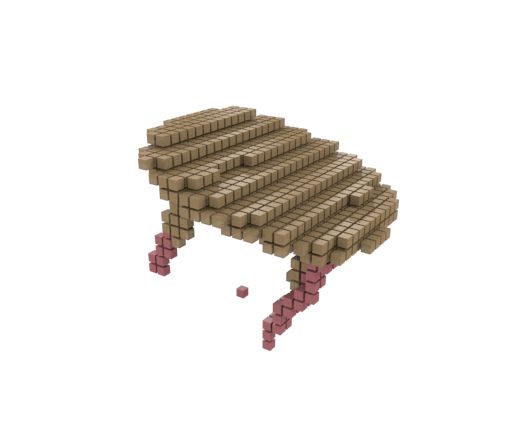}
		\end{subfigure}
		\begin{subfigure}[t]{0.15\textwidth}
			\vspace{0px}\centering
			\includegraphics[width=1.55cm,trim={\cropleft cm \croplower cm \cropright cm \cropupper cm},clip]{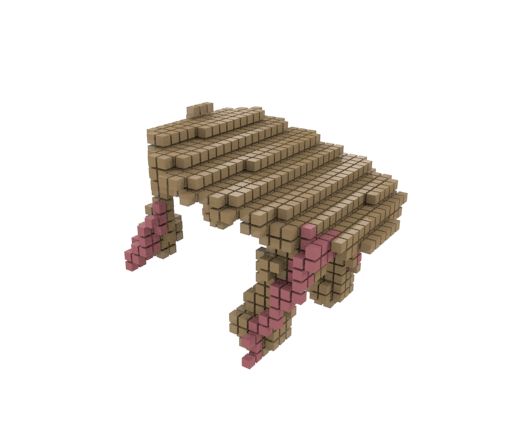}
		\end{subfigure}
        \subcaption{\Kinect, Low Resolution ($32^3$)}
	\end{subfigure}
    }
    \vspace*{-\figskipcaption px}
	\caption{{\bf Qualitative Results on KITTI and \Kinect.} On KITTI, \AML visually outperforms both \Dai and \Engelmann while being faster and requiring less supervision. On \Kinect, \AML demonstrates that it is able to generalize from as few as $30$ training samples. Predicted shapes (occupancy grids or meshes) in {\color{rbeige}beige} and observations in {\color{rred}red}; additionally, partial ground truth in {\color{rgreen}green}.}
	\label{fig:results-real}
    \vspace*{-\figskipbelow px}
\end{figure}
\begin{figure}[t]
    \vspace*{-\figskipabove px}
    \vspace*{2px}
    \centering
    {\scriptsize
    
    \begin{subfigure}[t]{0.5\textwidth}
        \vspace{0px}\centering
        \hspace*{-18px}
	    \begin{subfigure}[t]{0.13\textwidth}
	        \vspace{0px}\centering
	        \AML\\
	        \includegraphics[width=1.5cm,trim={\cropleft cm \croplower cm \cropright cm \cropupper cm},clip]{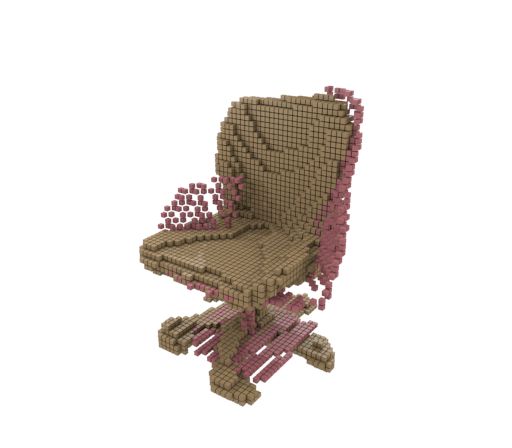}
	    \end{subfigure}
	    \begin{subfigure}[t]{0.13\textwidth}
	        \vspace{0px}\centering
	        GT\\
	        \includegraphics[width=1.5cm,trim={\cropleft cm \croplower cm \cropright cm \cropupper cm},clip]{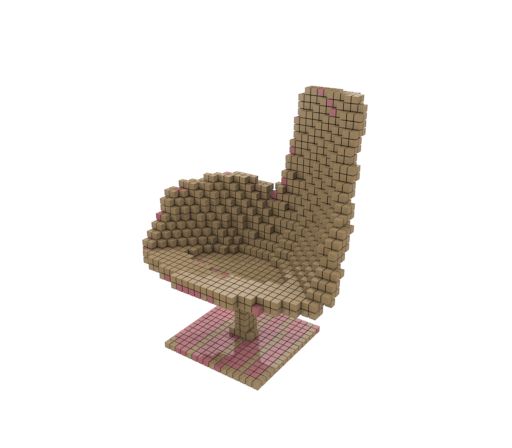}
	    \end{subfigure}
	    \begin{subfigure}[t]{0.13\textwidth}
	        \vspace{0px}\centering
	        \AML\\
	        \includegraphics[width=1.5cm,trim={\cropleft cm \croplower cm \cropright cm \cropupper cm},clip]{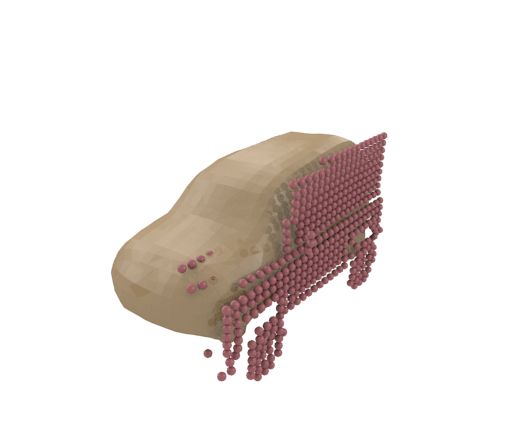}
	    \end{subfigure}
	    \begin{subfigure}[t]{0.13\textwidth}
	        \vspace{0px}\centering
	        GT\\
	        \includegraphics[width=1.5cm,trim={\cropleft cm \croplower cm \cropright cm \cropupper cm},clip]{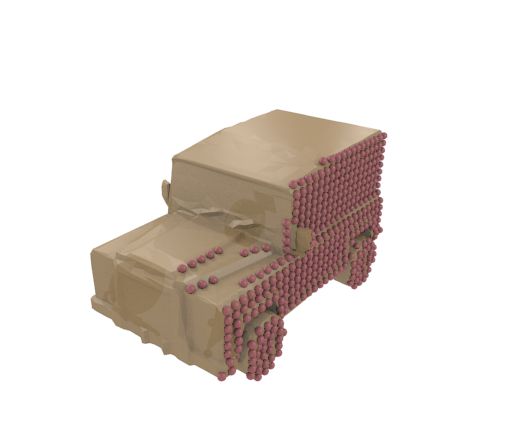}
	    \end{subfigure}
	    \begin{subfigure}[t]{0.13\textwidth}
	        \vspace{0px}\centering
	        \AML\\
	        \includegraphics[width=1.5cm,trim={\cropleft cm \croplower cm \cropright cm \cropupper cm},clip]{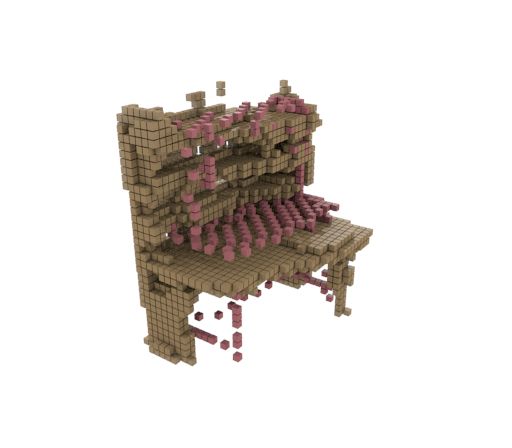}
	    \end{subfigure}
	    \begin{subfigure}[t]{0.13\textwidth}
	        \vspace{0px}\centering
	        \Dai\\
	        \includegraphics[width=1.5cm,trim={\cropleft cm \croplower cm \cropright cm \cropupper cm},clip]{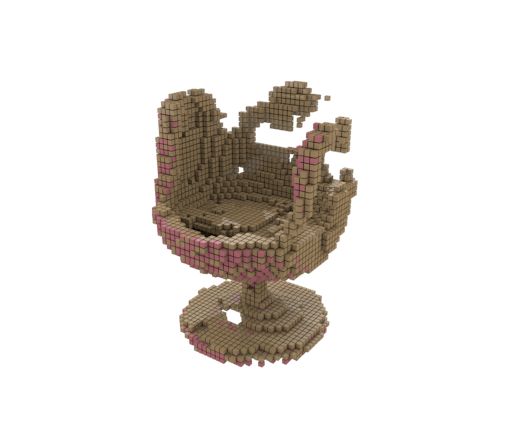}
	    \end{subfigure}
	    \begin{subfigure}[t]{0.13\textwidth}
	        \vspace{0px}\centering
	        \Dai\\
	        \includegraphics[width=1.5cm,trim={\cropleft cm \croplower cm \cropright cm \cropupper cm},clip]{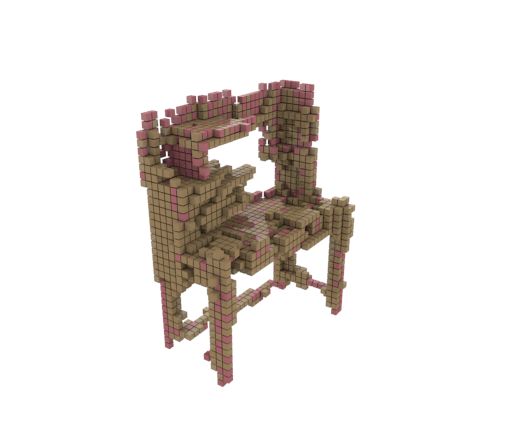}
	    \end{subfigure}
        \subcaption{Difficulties with Exotic Shapes and Fine Structures}
	\end{subfigure}
    \\[4px]
    \begin{subfigure}[t]{0.5\textwidth}
	    \begin{subfigure}[t]{0.15\textwidth}
	        \vspace{0px}\centering
	        \Dai\\
	        \includegraphics[width=1.5cm,trim={\cropleft cm \croplower cm \cropright cm \cropupper cm},clip]{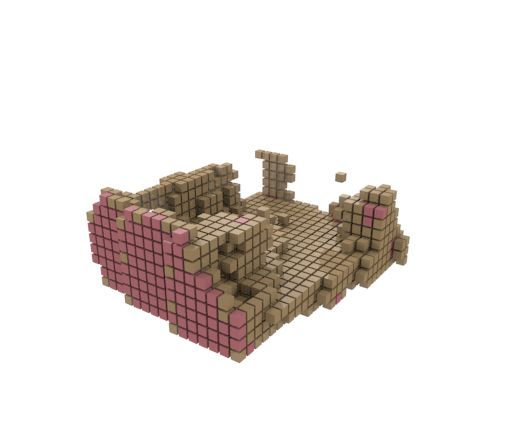}
	    \end{subfigure}
	    \begin{subfigure}[t]{0.15\textwidth}
	        \vspace{0px}\centering
	        \AML\\
	        \includegraphics[width=1.5cm,trim={\cropleft cm \croplower cm \cropright cm \cropupper cm},clip]{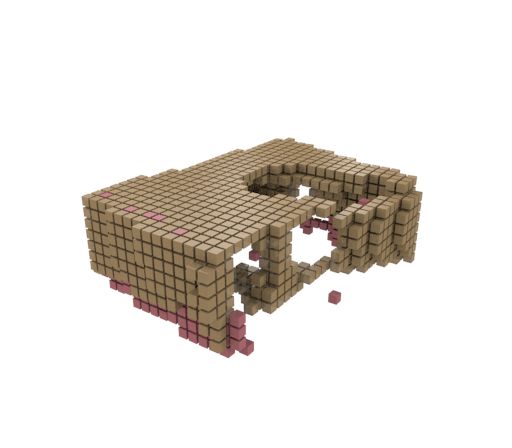}
	    \end{subfigure}
	    \begin{subfigure}[t]{0.15\textwidth}
	        \vspace{0px}\centering
	        GT\\
	        \includegraphics[width=1.5cm,trim={\cropleft cm \croplower cm \cropright cm \cropupper cm},clip]{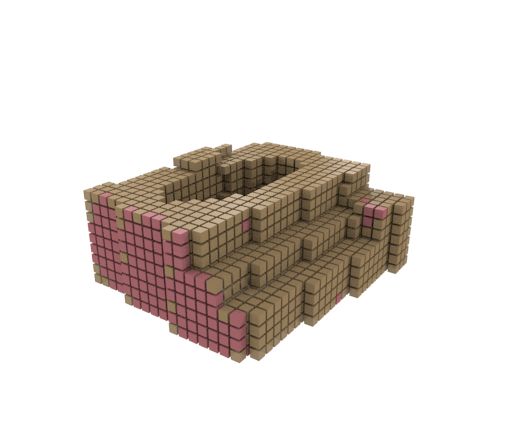}
	    \end{subfigure}
	    \begin{subfigure}[t]{0.15\textwidth}
	        \vspace{0px}\centering
	        \Dai\\
	        \includegraphics[width=1.5cm,trim={\cropleft cm \croplower cm \cropright cm \cropupper cm},clip]{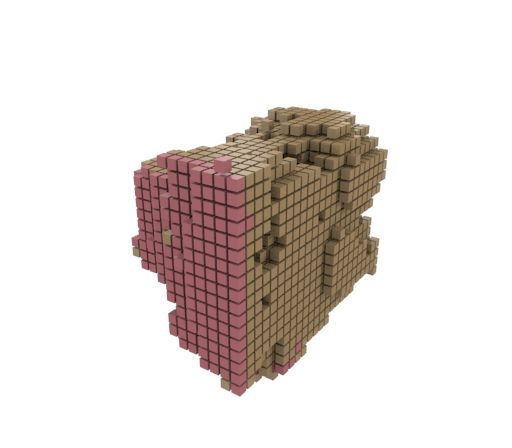}
	    \end{subfigure}
	    \begin{subfigure}[t]{0.15\textwidth}
	        \vspace{0px}\centering
	        \AML\\
	        \includegraphics[width=1.5cm,trim={\cropleft cm \croplower cm \cropright cm \cropupper cm},clip]{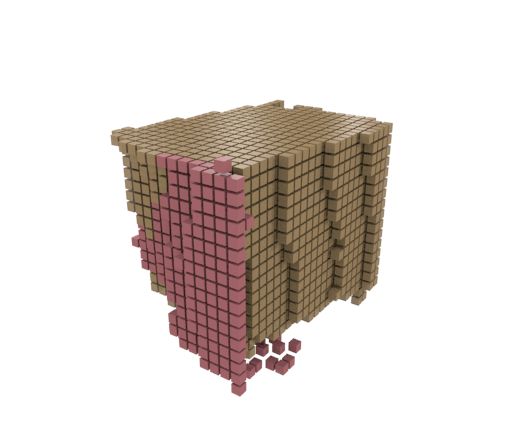}
	    \end{subfigure}
	    \begin{subfigure}[t]{0.15\textwidth}
	        \vspace{0px}\centering
	        GT\\
	        \includegraphics[width=1.5cm,trim={\cropleft cm \croplower cm \cropright cm \cropupper cm},clip]{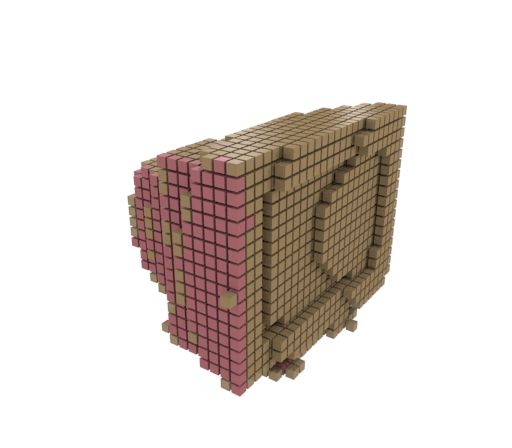}
	    \end{subfigure}
        \subcaption{Difficulties with Multiple Object Categories}
	\end{subfigure}
    }    
    \vspace*{-\figskipcaption px}
    \caption{{\bf Failures Cases.} On the top, we show that \AML has difficulties with exotic shapes, not represented in the latent space; and both \AML and \Dai have difficulties with fine details. The bottom row demonstrates that it is difficult to infer the correct object category from sparse observations, even under full supervision as required by \Dai. Shapes (occupancy grids and mehses) in {\color{rbeige}beige} and observations in {\color{rred}red} from various resolutions.}
    \label{fig:results-failures}
    \vspace*{-\figskipbelow px}
\end{figure}

\subsection{Experimental Evaluation}
\label{sec:experiments}

Quantitative results are summarized in \tabref{tab:results-shapenet} (ShapeNet and KITTI) and \ref{tab:results-modelnet} (ModelNet). Qualitative results for the shape prior are shown in \figref{fig:results-shape-prior} and \ref{fig:results-latent-space}; shape completion results are shown in \figref{fig:results-synthetic} (ShapeNet and ModelNet) and \ref{fig:results-real} (KITTI and \Kinect).

\boldparagraph{Latent Space Dimensionality}
Regarding our \DVAE shape prior, we found the dimensionality $Q$ to be of crucial importance as it defines the trade-off between reconstruction accuracy and random sample quality (\ie, the quality of the generative model). A higher-dimension-al latent space usually results in higher-quality reconstructions but also imposes the difficulty of randomly generating meaningful shapes. Across all datasets, we found $Q = 10$ to be suitable -- which is significantly smaller compared to related work: $35$ in \citep{Liu2017ARXIV}, $6912$ in \citep{Sharma2016ARXIV}, $200$ for \citep{Wu2016NIPS,Smith2017ARXIV} or $64$ in \citep{Girdhar2016ECCV}. Still, we are able to obtain visually appealing results. Finally, in \figref{fig:results-shape-prior} we show qualitative results, illustrating good reconstruction performance and reasonable random samples across resolutions.

\figref{fig:results-latent-space} shows a t-SNE \citep{Maaten2008JMLR} visualization as well as a projection of the $Q = 10$ dimensional latent space, color coding the $10$ object categories of ModelNet10. The \DVAE clusters the object categories within the support region of the unit Gaussian. In the t-SNE visualization, we additionally see ambiguities arising in ModelNet10, \eg, night stands and dressers often look indistinguishable while monitors are very dissimilar to all other categories. Overall, these findings support our decision to use a \DVAE with $Q=10$ as shape prior.

\boldparagraph{Ablation Study}
In \tabref{tab:results-shapenet}, we show quantitative results of our model on \clean and \noisy. First, we report the reconstruction quality of the \DVAE shape prior as reference. Then, we consider the \DVAE shape prior (\BL), and its mean prediction (\M) as simple baselines. The poor performance of both illustrates the difficulty of the benchmark. For \AML, we also consider its deterministic variant, \dAML (see \secref{sec:method}). Quantitatively, there is essentially no difference; however, \figref{fig:results-shape-prior} demonstrates that \AML is able to predict more detailed shapes. We also found that using both occupancy and SDFs is necessary to obtain good performance -- as is using both point observations and free space.

Considering \figref{fig:results-latent-space}, we additionally demonstrate that the embedding learned by \AML, \ie, the embedding of incomplete observations within the latent shape space, is able to associate observations with corresponding shapes even under weak supervision. In particular, we show a t-SNE visualization and a projection of the latent space for \AML trained on \clean. We color-code $10$ randomly chosen ground truth shapes, resulting in $100$ observations ($10$ views per shape). \AML is usually able to embed observations near the corresponding ground truth shapes, without explicit supervision (\eg, for violet, pink, blue or teal, the observations -- points -- are close to the corresponding ground truth shapes -- ``x''). Additionally, \AML also matches the unit Gaussian prior distribution reasonably well.

\boldparagraph{Comparison to Baselines on Synthetic Data}
For ShapeNet, \tabref{tab:results-shapenet} demonstrates that \AML outperforms data-driven approaches such as \Engelmann, \ICP and \ML and is able to compete with fully-supervised approaches, \Dai and \Sup, while using only $8\%$ or less supervision. We also note that~\AML outperforms \ML, illustrating that amortized inference is beneficial. Furthermore, \Dai outperforms \Sup, illustrating the advantage of propagating low-level information (through skip connections) without bottleneck. Most importantly, the performance gap between \AML and \Dai is rather small considering the difference in supervision (more than $92\%$) and on \noisy, the drop in performance for \Dai and \Sup is larger than for \AML suggesting that \AML handles noise and sparsity more robustly. \figref{fig:results-synthetic} shows that these conclusions also apply visually where \AML performs en par with \Dai.

For ModelNet, in \tabref{tab:results-modelnet}, we mostly focus on occupancy grids (as the derived SDFs are approximate, \cf \secref{sec:data}) and show that chairs, desks or tables are more difficult. However, \AML is still able to predict high-quality shapes, outperforming data-driven approaches. Additionally, in comparison to ShapeNet, the gap between \AML and fully-supervised approaches (\Dai and \Sup) is surprisingly small -- not reflecting the difference in supervision. This means that even under full supervision, these object categories are difficult to complete. In terms of accuracy (\Acc) and completeness (\Compl), \eg, for chairs, \AML outperforms \ICP and \ML; \Dai and \Sup, on the other hand, outperform \AML. Still, considering \figref{fig:results-synthetic}, \AML predicts visually appealing meshes although the reference shape SDFs on ModelNet are merely approximate. Qualitatively, \AML also outperforms its data-driven rivals; only \Dai predicts shapes slightly closer to the ground truth.

\boldparagraph{Multiple Views and Higher Resolutions}
In \tabref{tab:results-shapenet}, we consider multiple, $k \in \{2,3,5\}$, randomly fused observations (from the $10$ views per shape). Generally, additional observations are beneficial (also \cf \figref{fig:results-synthetic-extra}); however, fully-supervised approaches such as \Dai benefit more significantly than \AML. Intuitively, especially on \noisy, $k = 5$ noisy observations seem to impose contradictory constraints that cannot be resolved under weak supervision. We also show that higher resolution allows both \AML and \Dai to predict more detailed shapes, see \figref{fig:results-synthetic-extra}; for \AML this is significant as, \eg, on \noisy, the level of supervision reduces to less than $1\%$. \red{Also note that \AML is able to handle the slightly asymmetric desks in \figref{fig:results-synthetic-extra} due to the strong shape prior which itself includes symmetric and less symmetric shapes.}

\boldparagraph{Multiple Object Categories}
We also investigate the category-agnostic case, considering all ten ModelNet10 object categories; here, we train a single \DVAE shape prior (as well as a single model for \Dai and \Sup) across all ten object categories. As can be seen in \tabref{tab:results-modelnet}, the gap between \AML and fully-supervised approaches, \Dai and \Sup, further shrinks; even fully-supervised methods have difficulties distinguishing object categories based on sparse observations. \figref{fig:results-synthetic-extra} shows that \AML is able to not only predict reasonable shapes, but also identify the correct object category. In contrast to \Dai, which predicts slightly more detailed shapes, this is significant as \AML does not have access to object category information during training.

\boldparagraph{Comparison on Real Data}
On KITTI, considering \figref{fig:results-real}, we illustrate that \AML consistently predicts detailed shapes regardless of the noise and sparsity in the inputs. Our qualitative results suggest that \AML is able to predict more detailed shapes compared to \Dai and \Engelmann; additionally, \Engelmann is distracted by sparse and noisy observations. Quantitatively, instead, \Dai and \Sup outperform \AML. However, this is mainly due to two factors: first, the ground truth collected on KITTI does rarely cover the full car; and second, we put significant effort into faithfully modeling KITTI's noise statistics in \noisy, allowing \Dai and \Sup to generalize very well. The latter effort, especially, can be avoided by using our weakly-supervised approach, \AML.

On \Kinect, also considering \figref{fig:results-real}, only $30$ observations are available for training. It can be seen that \AML predicts reasonable shapes for tables. We find it interesting that \AML is able to generalize from only $30$ training examples. In this sense, \AML functions similar to \ML, in that the objective is trained to overfit to few samples. This, however, cannot work in all cases, as demonstrated by the chairs where \AML tries to predict a suitable chair, but does not fit the observations as well. Another problem witnessed on \Kinect, is that the shape prior training samples need to be aligned to the observations (with respect to the viewing angles). For the chairs, we were not able to guess the viewing trajectory correctly (\cf \citep{Yang2018ARXIVb}).

\boldparagraph{Failure Cases}
\AML and \Dai often face similar problems, as illustrated in \figref{fig:results-failures}, suggesting that these problems are inherent to the used shape representations or the learning approach independent of the level of supervision. For example, both \AML and \Dai have problems with fine, thin structures that are hard to reconstruct properly at any resolution. Furthermore, identifying the correct object category on ModelNet10 from sparse observations is difficult for both \AML and \Sup. Finally, \AML additionally has difficulties with exotic objects that are not well represented in the latent shape space as, \eg, designed chairs.

\boldparagraph{Runtime}
{At low resolution, \AML as well as the fully-supervised approaches \Dai and \Sup, are particular fast, requiring up to $2ms$ on a NVIDIA\texttrademark\xspace GeForce\textregistered\xspace GTX TITAN using Torch \citep{Collobert2011NIPSWORK}. Data-driven approaches (\eg, \Engelmann, \ICP and \ML), on the other hand, take considerably longer. \Engelmann, for instance requires $168ms$ on average for completing the shape of a sparse LIDAR observation from KITTI using an Intel\textregistered\xspace Xeon\textregistered\xspace E5-2690 @2.6Ghz and the multi-threaded Ceres solver \citep{AgarwalCeres}. \ICP and \ML take longest, requiring up to $38s$ and $75s$ (not taking into account the point sampling process for the shapes), respectively. Except for \Engelmann and \ICP, all approaches scale with the used resolution and the employed architecture.}

\section{Conclusion}
\label{sec:conclusion}

In this paper, we presented a novel, weakly-supervised learning-based approach to 3D shape completion from sparse and noisy point cloud observations. We used a (denoising) variational auto-encoder \citep{Im2017AAAI,Kingma2014ICLR} to learn a latent space of shapes for one or multiple object categories using synthetic data from ShapeNet \citep{Chang2015ARXIV} or ModelNet \citep{Wu2015CVPR}. Based on the learned generative model, \ie, decoder, we formulated 3D shape completion as a maximum likelihood problem. In a second step, we then fixed the learned generative model and trained a new recognition model, \ie encoder, to amortize, \ie \emph{learn}, the maximum likelihood problem. Thus, our {\bf Amortized Maximum Likelihood (\AML)} approach to 3D shape completion can be trained in a weakly-supervised fashion. Compared to related data-driven approaches, \eg, \citep{Rock2015CVPR,Haene2014CVPR,Li2015CGF,Engelmann2016GCPR,Engelmann2017WACV,Nan2012TG,Bao2013CVPR,Dame2013CVPR,Ngyuen2016CVPR}, our approach offers fast inference at test time; in contrast to other learning-based approaches, \eg, \citep{Riegler2017THREEDV,Smith2017ARXIV,Dai2017CVPRa,Sharma2016ARXIV,Fan2017CVPR,Rezende2016ARXIV,Yang2018ARXIVb,Wang2017ICCV,Varley2017IROS,Han2017ICCV}, we do not require full supervision during training. Both characteristics render our approach useful for robotic scenarios where full supervision is often not available such as in autonomous driving, \eg, on KITTI \citep{Geiger2012CVPR}, or indoor robotics, \eg, on \Kinect \citep{Yang2018ARXIVb}.

On two newly created synthetic shape completion benchmarks, derived from ShapeNet's cars and ModelNet10, as well as on real data from KITTI and, we demonstrated that \AML outperforms related data-driven approaches \citep{Engelmann2016GCPR,Gupta2015CVPR} while being significantly faster. We further showed that \AML is able to compete with fully-supervised approaches \citep{Dai2017CVPRa}, both quantitatively and qualitatively, while using only $3-10\%$ supervision or less. In contrast to \citep{Rock2015CVPR,Haene2014CVPR,Li2015CGF,Engelmann2016GCPR,Engelmann2017WACV,Nan2012TG,Bao2013CVPR,Dame2013CVPR}, we additionally showed that \AML is able to generalize across object categories without category supervision during training. On \Kinect, we also demonstrated that our \AML approach is able to generalize from very few training examples. In contrast to \citep{Girdhar2016ECCV,Liu2017ARXIV,Sharma2016ARXIV,Wu2015CVPR,Dai2017CVPRa,Firman2016CVPR,Han2017ICCV,Fan2017CVPR}, we considered resolutions up to $48 \ntimes 108 \ntimes 48$ and $64^3$ voxels as well as significantly sparser observations. Overall, our experiments demonstrate two key advantages of the proposed approach: significantly reduced runtime and increased performance compared to data-driven approaches showing that amortizing inference is highly effective.

In future work, we would like to address several aspects of our \AML approach. First, the shape prior is essential for weakly-supervised shape completion, as also noted by \cite{Gwak2017ARXIV}. However, training expressive generative models in 3D is still difficult. Second, larger resolutions imply significantly longer training times; alternative shape representations and data structures such as point clouds \citep{Qi2017CVPR,Qi2017NIPS,Fan2017CVPR} or octrees \citep{Riegler2017CVPR,Riegler2017THREEDV,Haene2017ARXIV} might be beneficial. Finally, jointly tackling pose estimation and shape completion seems promising~\citep{Engelmann2016GCPR}.

{\small
    \bibliographystyle{spbasic}
    \bibliography{bibliography_long,bibliography,bibliography_custom}
}

\end{document}